\RequirePackage{fix-cm}
\documentclass[10pt,twocolumn,letterpaper]{article}
\usepackage[pagenumbers]{cvpr} 

\usepackage[utf8]{inputenc}
\usepackage[T1]{fontenc}
\usepackage{amsfonts}        
\usepackage{nicefrac}        
\usepackage{microtype}       
\usepackage{multirow}        
\usepackage{adjustbox}       
\usepackage{float}           
\usepackage{textcomp}        
\usepackage{pifont}          
\usepackage{colortbl}        
\usepackage{tabularray}      
\usepackage{fontawesome5}    
\usepackage{listings}        
\usepackage{etoc}            

\usepackage{booktabs}
\usepackage{tabularx}
\usepackage{array}

\usepackage{tikz}
\usetikzlibrary{arrows.meta, positioning, calc}

\usepackage[most]{tcolorbox}
\tcbuselibrary{listings,breakable}
\tcbset{enhanced jigsaw, breakable}

\usepackage[ruled,vlined,linesnumbered]{algorithm2e}

\usepackage{amsmath,amsfonts,bm}

\usepackage[most]{tcolorbox}  
\usepackage{tabularray}       
\UseTblrLibrary{booktabs}     
\usepackage{xcolor}           
\usepackage{graphicx}         
\usepackage{microtype}        

\usepackage{caption}
\usepackage{subcaption}  

\usepackage{amsmath}
\usepackage{amsfonts}
\usepackage{amssymb} 
\usepackage[T1]{fontenc}

\definecolor{GalleryGold}{RGB}{186,146,58} 
\definecolor{Canvas}{RGB}{250,250,248}

\definecolor{GalleryGold}{HTML}{D4AF37}
\definecolor{Canvas}{HTML}{1A1A1B}
\definecolor{TextSilver}{HTML}{F0F0F0}

\newcommand{\newterm}[1]{{\bf #1}}

\def\eqref#1{equation~\ref{#1}}

\def\1{\bm{1}}

\DeclareMathAlphabet{\mathsfit}{\encodingdefault}{\sfdefault}{m}{sl}
\SetMathAlphabet{\mathsfit}{bold}{\encodingdefault}{\sfdefault}{bx}{n}

\newcommand{\Can}{\mathcal{C}}          
\newcommand{\Non}{\mathcal{N}}          
\newcommand{\M}{M}                      
\newcommand{\Mp}{M'}                    
\newcommand{\Fit}{\mathrm{Fit}}
\newcommand{\LP}{\mathrm{LP}}
\newcommand{\PC}{\mathrm{PC}}
\newcommand{\CWR}{\mathrm{CWR}}
\newcommand{\ind}{\mathbf{1}}           
\renewcommand{\newterm}[1]{\textbf{#1}} 

\definecolor{qcoral}{RGB}{200, 80, 80}      
\definecolor{qpurple}{RGB}{100, 120, 200}   
\definecolor{qamber}{RGB}{80, 160, 100}     
\definecolor{qteal}{RGB}{150, 90, 180}      

\definecolor{sky}{HTML}{6EC3F4}
\definecolor{violet}{HTML}{A875FF}
\definecolor{mint}{HTML}{2ED7C3}
\definecolor{ink}{HTML}{1F2A44}
\definecolor{paper}{HTML}{F8FAFE}

\definecolor{SemColor}{RGB}{108,123,158}
\definecolor{AesColor}{RGB}{186,108,83}
\definecolor{FidColor}{RGB}{122,146,118}
\definecolor{BGBar}{RGB}{238,236,232}

\definecolor{StepOneBg}{HTML}{E3F2FD}
\definecolor{StepTwoBg}{HTML}{E8F5E9}
\definecolor{StepThreeBg}{HTML}{FFF3E0}
\definecolor{StepText}{HTML}{1A1A1A}
\definecolor{insightblue}{RGB}{0,105,148}
\definecolor{insightback}{RGB}{240,248,255}

\definecolor{GalleryGold}{HTML}{AF9164}
\definecolor{Canvas}{HTML}{FDFBF7}
\definecolor{Charcoal}{HTML}{333333}
\definecolor{PaintBlue}{HTML}{A3C1AD}

\newcommand{\artimg}[2]{%
  \multirow{#1}{=}{%
    \centering
    \adjustbox{%
      max width=2.8cm,%
      max height=\dimexpr #1\baselineskip\relax,%
      keepaspectratio%
    }{\includegraphics{#2}}%
  }%
}

\newcommand{\StepHeader}[2]{%
  \par\noindent
  \colorbox{#1}{%
    \parbox{\dimexpr\hsize-2\fboxsep\relax}{%
      \textbf{\color{StepText}#2}}}%
  \par\vspace{3pt}%
}

\newtcolorbox{GlassCard}[1][]{
  breakable, enhanced,
  colback=paper,
  colframe=violet!70!ink,
  boxrule=0.8pt, arc=3mm, outer arc=3mm,
  left=4mm, right=4mm, top=3mm, bottom=3mm,
  borderline={0.6pt}{-0.5mm}{white},
  overlay={
    \begin{scope}
      \path[fill opacity=0.22, draw=none]
        ([xshift=-6mm,yshift=4mm]frame.north west) [rounded corners=8mm]
        -- ([xshift=6mm,yshift=4mm]frame.north east)
        -- ([xshift=6mm,yshift=-6mm]frame.south east)
        -- ([xshift=-6mm,yshift=-6mm]frame.south west) -- cycle;
    \end{scope}},
  title={Challenger Prompt Construction},
  fonttitle=\bfseries\sffamily\large,
  coltitle=white,
  colbacktitle=sky!90!violet,
  boxed title style={
    frame code={
      \path[draw=none, fill=sky!85!violet]
        ([yshift=-0.8mm,xshift=-0.8mm]frame.north west)
        rectangle ([yshift=3.8mm,xshift=0.8mm]frame.north east);},
    interior code={
      \path[shade, left color=sky!90, right color=violet!85]
        ([yshift=-0.8mm,xshift=-0.8mm]frame.north west)
        rectangle ([yshift=3.8mm,xshift=0.8mm]frame.north east);},
  },
  borderline west={2mm}{0pt}{mint!80!sky},
  drop fuzzy shadow=ink!25!white,
  #1
}

\usepackage[most]{tcolorbox}

\definecolor{auditBlue}{HTML}{577DA8}
\definecolor{auditTeal}{HTML}{26847C}
\definecolor{auditViolet}{HTML}{7968B4}
\definecolor{auditAmber}{HTML}{BB8A3B}
\definecolor{auditRose}{HTML}{B46E8C}
\definecolor{auditGreen}{HTML}{4A905A}
\definecolor{auditCoral}{HTML}{BB7059}

\newtcolorbox{auditstep}[2]{
    enhanced,
    breakable,
    colback=#2!4,
    colframe=#2!35,
    colbacktitle=#2!12,
    coltitle=black,
    fonttitle=\bfseries,
    title={#1},
    borderline west={2pt}{0pt}{#2},
    boxrule=0.35pt,
    arc=1pt,
    left=5pt,
    right=5pt,
    top=4pt,
    bottom=4pt,
    before skip=4pt,
    after skip=4pt
}

\providecommand{\Fit}{\mathrm{Fit}}
\providecommand{\DLR}{\mathrm{DLR}}

\definecolor{valgreen}{RGB}{214,240,214}
\definecolor{valred}{RGB}{248,214,214}
\newcommand{\valgood}[1]{\cellcolor{valgreen}#1}
\newcommand{\valbad}[1]{\cellcolor{valred}#1}

\definecolor{cvprblue}{rgb}{0.21,0.49,0.74}
\usepackage[pagebackref,breaklinks,colorlinks,allcolors=cvprblue]{hyperref}

\def\confName{CVPR}
\def\confYear{2027}

\title{The Silent Brush: Evaluating Artistic Style Leakage in AI Art Generation}

\author{Ninad Joshi \quad Ashutosh Ranjan \quad Vivek Srivastava \quad Shirish Karande\\
TCS Research, India\\
{\tt\small \{joshi.ninad, ashutosh.ranjan2, srivastava.vivek2, shirish.karande\}@tcs.com}
}

\begin{document}
\maketitle

\begin{abstract}
Text-to-image models are trained on web-scale data that include copyrighted and stylistically distinctive artworks, raising concerns about ownership, attribution, and unintended reuse. These models can learn stylistic patterns and reproduce them without any reference in the prompt, a phenomenon we term \textit{\textbf{The Silent Brush}}. The risk is concrete for artists who personalize these models on their own artworks for co-creation and exploration. They personalize the base model rather than replacing it, so artworks strongly encoded in the base model can resurface in generations meant to carry the artist's style. Existing evaluations focus on near-duplicate retrieval or membership inference and miss this unrequested resurfacing. We introduce \textit{\textbf{ArtArena}}, a model-agnostic protocol for auditing style leakage, built on four guiding principles: whether models imitate artworks, how they blend imitated artworks, whether these artworks exhibit proactive blending, and whether its strength can be steered. We validate ArtArena on SD v1.5, SDXL and SANA-1.5 through controlled interventions whose effects are known a priori, testing whether it recovers them. Furthermore, we demonstrate ArtArena as a practical audit for artists. We first identify the canonical artworks with the highest leakage potential for the base model. We then determine which of the canonical artworks are most likely to leak into generations meant to reflect the artist's style after personalization. We present a case study with two contemporary artists, Trent Anthony Francis and Gwenn Seemel. Each audit across the three models is summarized in an artist-facing leakage card.
\end{abstract}

\vspace{-5.5mm}

\section{Introduction}
\label{sec:intro}
\begin{figure}[t]
  \centering
  \includegraphics[width=\linewidth]{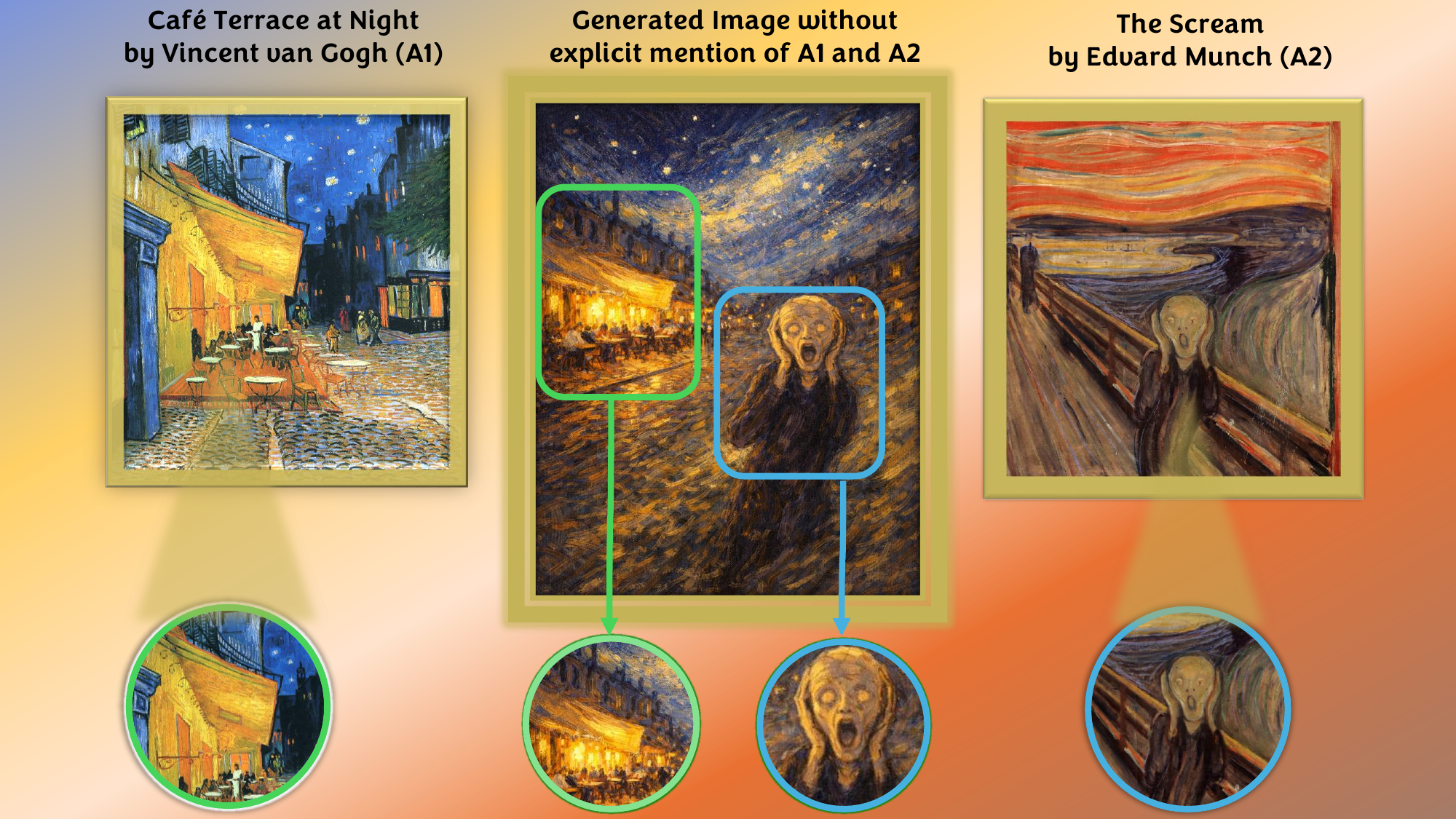}
  \caption{%
    \textit{\textbf{The Silent Brush}:}
    An image generated by a text-to-image model that, without explicitly
    prompting \emph{Caf\'e Terrace at Night} or \emph{The Scream},
    visibly incorporates elements from both artworks.
    \textbf{Prompt used for image generation:}
    ``A solitary, androgynous figure stands rigid in the foreground,
    clutching its head in a silent scream as golden caf\'e light spills
    onto the street behind.
    The glowing awning and receding tables stretch into the night,
    while the sky above churns in agitated waves of blue and yellow.
    Loose, directional brushstrokes bind figure, architecture, and sky
    into a single nocturnal field of anxiety, light, and distorted
    perception.''}
  \label{fig:teaser}
  \vspace{-0.45cm}
\end{figure}

Every artwork carries a distinctive stylistic fingerprint shaped by choices in color, composition, and technique \citep{jafarpour2009stylistic,dangeti2024style}, and generative models can learn and reproduce these patterns \citep{chen2025stylesentinel,wan2024style}. As these models enter creative workflows, their outputs may appear novel while retaining influences from existing artistic traditions. This raises a central question: \textbf{\textit{how can we evaluate stylistic influence within generative models?}}

Influence is integral to art: creativity research describes innovation as the recombination of existing ideas \citep{boden2004creative,koestler1964act}, art theory views artworks as intertextual \citep{barthes1977image,kristeva1980word}, and cognitive psychology shows how stylistic cues activate learned schemas that shape similarity judgments \citep{rosch1975cognitive,barsalou1985ideals}. Technological reproducibility, however, raises questions of originality, authorship, and consent \citep{benjamin1935work}. Generative models internalize style as statistical regularities such as texture correlations and color distributions \citep{everaert2023diffusion,arias2025color}, capture global covariance structures that support style-consistent generation under weak or ambiguous prompts \citep{li2024understanding,gu2023memorization}, and may hold stylistic priors that act as attractors during sampling \citep{pham2025memorization} and influence generations without explicit style conditioning \citep{deng2024z}. Detailed related work is discussed in Section~\ref{sec:genai_models_datasets} of the Appendix.

Yet how stylistic influences interact, compete, and blend during generation remains poorly understood. Existing evaluations emphasize prompt fidelity and perceptual quality rather than the relative influence of learned artworks, and evidence of memorization and leakage shows that learned patterns resurface under targeted prompting \citep{shokri2017membership,carlini2021extracting} without explaining why certain styles prevail over others. 
\begin{figure*}[t]
  \centering
  \IfFileExists{LeakageCards_Trent/SDXL_Trent_csd_LeakageCard_v4_Type2_Dumbbell.pdf}{%
    \includegraphics[width=0.85\textwidth]{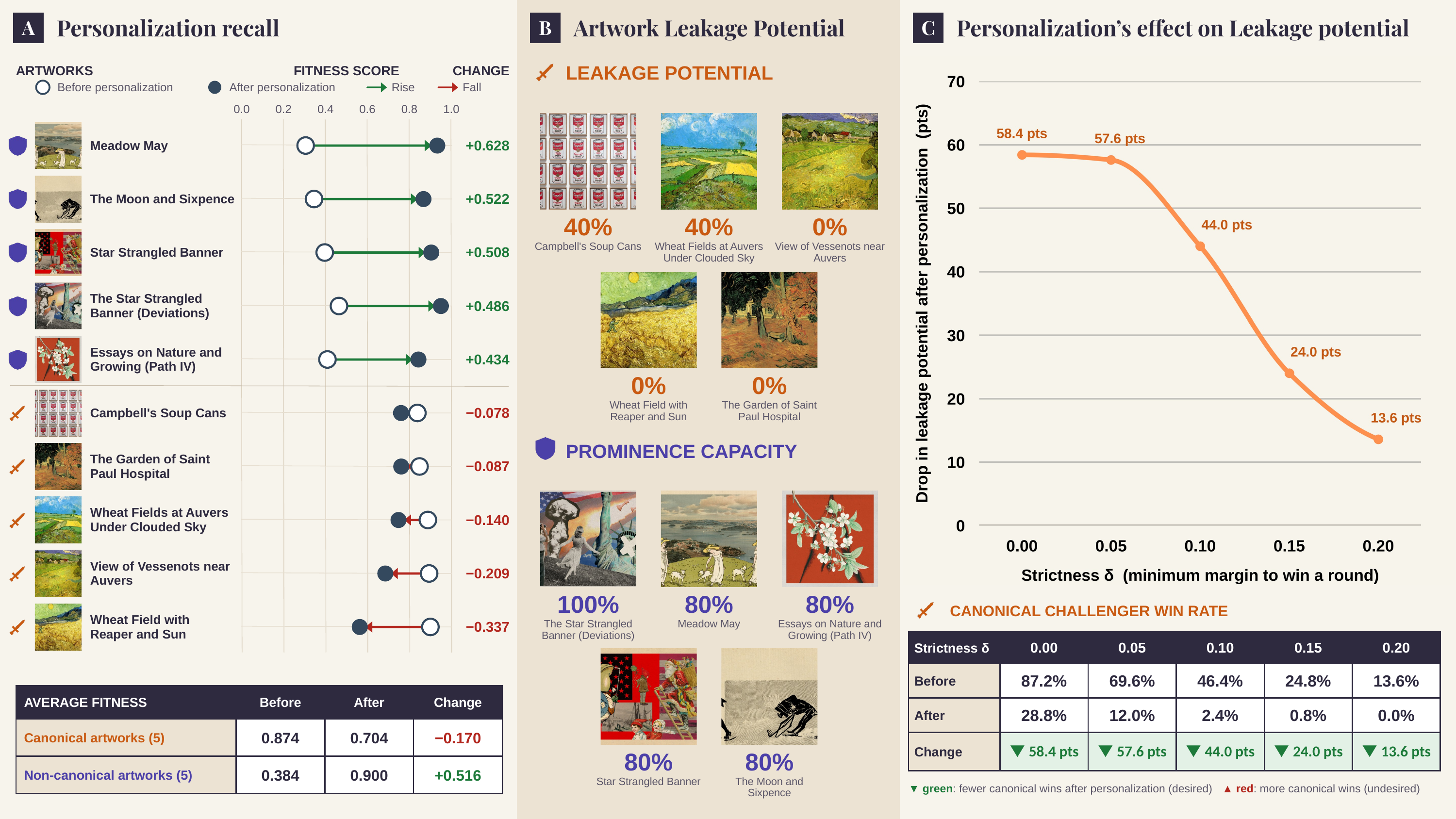}}{%
    \fbox{\parbox[c][6cm][c]{0.95\textwidth}{\centering
      \textit{Placeholder: SDXL leakage card (Trent Anthony Francis, CSD)}}}}
  \caption{Leakage card for SDXL personalized using non-canonical artworks by Trent Anthony Francis, evaluated for fidelity using CSD.}
  \label{fig:LC_SDXL_Trent_CSD}
\end{figure*}

We call the unrequested resurfacing of learned stylistic characteristics the \textbf{Silent Brush} (Figure~\ref{fig:teaser}). It matters most to visual artists, who use text-to-image models for exploration, production, and communication \citep{ko2023largescale}. Artists may prompt pre-trained models \citep{chang2023promptartists}, adapt them to their own artworks through \newterm{personalization} \citep{abuzuraiq2024personalizing,rajcic2024diffractive}, or train models only on their own work \citep{elgammal2023folly}. Since training from scratch demands substantial compute \citep{abuzuraiq2024personalizing,chen2024pixartalpha} and a single artist's collection may be too small for satisfactory quality \citep{pamies2026sketchit}, personalization methods such as Specialist Diffusion \citep{lu2023specialist}, StyleDrop \citep{sohn2023styledrop}, and StyleBoost \citep{park2023styleboost} offer a practical alternative.

Artists nevertheless report concerns with both kinds of model. Pre-trained models may produce outputs that are less novel in content and style \citep{zhou2024generative}, and nearly two-thirds of over 500 surveyed artists worried about resemblance to existing artworks \citep{elgammal2023folly}. With personalized models, illustrators recognized fragments of their style without its emergent qualities \citep{porquet2025copying}, and practitioners reported stylistic priors persisting from the base model \citep{kozak2026flux}. These concerns are consistent with models learning artworks unevenly: diffusion models disproportionately memorize duplicated or atypical images \citep{carlini2023extracting}, reproduce some artistic styles more faithfully than others \citep{somepalli2024measuring}, and generate recurring visual patterns under ambiguous prompts \citep{simonen2026default}. In personalized models, this unevenness becomes \newterm{leakage}: characteristics of canonical artworks learned by the base model persist and resurface in generations meant to reflect the artist's own style. 


Existing methods do not measure this leakage. Replication analyses find near-copies of training images \citep{somepalli2023diffusion}, and style similarity measures score resemblance to individual reference artworks \citep{somepalli2024measuring}; neither asks whether a canonical artwork prevails over the style the artist intends. Protective approaches prevent unauthorized style imitation \citep{shan2023glaze,vanle2023antidreambooth} but do not audit the artist's own model, personalization evaluations focus on the intended concept and may miss changes to others \citep{laria2026driftscope}, and artist-centered workflows rely on visual inspection \citep{grimes2026blkassist}. Crucially, \textbf{stylistic leakage is relational}: it occurs when a canonical artwork prevails over the intended style within the same generation, so it cannot be assessed one artwork at a time. Artists therefore need a systematic way to distinguish general stylistic resemblance from artwork-specific influence: to identify which canonical artworks interfere with their own, and to what extent.

To address this gap, we introduce \textit{\textbf{ArtArena}}, a model-agnostic auditing protocol built on four guiding principles (Section~\ref{sec:artarena}) and operationalized in three stages. \textit{Entry Trials} assess imitation under explicit prompting, \textit{Motif Duels} pit an explicitly named artwork against another represented only by its motifs, and the \textit{Influence Ledger} ranks artworks by their relative stylistic dominance across these interactions.

We demonstrate ArtArena in a case study (Section~\ref{sec:personalization}) with SD v1.5, SDXL, and SANA-1.5 \citep{rombach2022high,sdxl_base_1_0,xie2025sana15} and two contemporary artists, Trent Anthony Francis and Gwenn Seemel. Our contributions are:

\begin{itemize}[leftmargin=*]
    \item \textbf{Problem.} We characterize the \textit{Silent Brush} and the leakage it causes in personalized text-to-image models.

    \item \textbf{Method.} We introduce \textit{ArtArena}, a model-agnostic and metric-specific auditing protocol, and validate its sensitivity through controlled interventions that strengthen or weaken individual artworks' stylistic influence (Section~\ref{sec:Val_AA}).

    \item \textbf{Application.} We show how ArtArena audits personalized models and summarize each audit in an artist-facing \newterm{leakage card}.
\end{itemize}

\newtcolorbox{insightbox}[1]{
   enhanced,
   colback=insightback,
   colframe=insightblue,
   coltitle=white,
   fonttitle=\bfseries,
   sharp corners,
   boxrule=0pt,
   leftrule=4pt,
   attach title to upper,
   title={\color{orange}\faLightbulb\ }, 
   separator sign none,
   description font=\mdseries,
   arc=2mm,
   breakable,
   top=2mm,
   bottom=2mm,
   left=3mm,
   right=3mm,
}

\section{\textit{ArtArena}: An Auditing Protocol}
\label{sec:artarena}

\subsection{Guiding Principles}
\label{sec:guiding_principles}

Motivated by these concerns, we identify four fundamental questions artists may raise about stylistic influence and establish them as the guiding principles of our stylistic leakage audit protocol. 

\noindent\textbf{Q1: Which artworks can a model imitate?}
An artist may first seek to understand whether a generative model can reproduce recognizable characteristics of existing artworks. When prompted with an artwork's identity, a model may generate images resembling its content, composition, or style. However, artistic style encompasses multiple visual characteristics, including texture and color, which may not be reproduced with equal fidelity \citep{gatys2015neural,somepalli2024measuring}. An auditing method should therefore assess the extent to which a model can imitate individual artworks under explicit prompting, providing a foundation for examining their interactions during generation.

\noindent\textbf{Q2: How does a model blend characteristics of different artworks?}
The ability to imitate individual artworks does not necessarily explain how their characteristics interact during generation. Since a model learns stylistic representations from numerous artworks, its outputs may reflect the combined influence of these learned characteristics. An artist may therefore question whether the characteristics of a requested artwork remain recognizable in the presence of other artistic influences, or whether such influences interfere with its intended appearance. Previous research has examined the distinction between content and style \citep{gatys2015neural,frenkel2024blora}, the recombination of local features from training examples in convolutional diffusion models \citep{kamb2025analytic}, and the role of text conditioning in training-image replication \citep{somepalli2023understanding}.


\noindent\textbf{Q3: Do certain artworks exhibit proactive blending?}
Beyond individual interactions, an artist may wish to identify artworks whose characteristics consistently prevail over others during blending, including when they are not explicitly requested. We refer to an artwork's tendency to assert its stylistic characteristics over those of other artworks during generation as \textbf{proactive blending}. An auditing method should therefore assess how consistent and how strong this tendency is across interactions with different artworks.

\noindent\textbf{Q4: Can the strength of proactive blending be steered?}
Artists personalizing or adapting pre-trained models may seek to understand whether the influence of previously learned artworks can be strengthened or weakened, particularly when such influences interfere with their intended style. An auditing method should therefore be capable of detecting changes in relative stylistic influence following model modifications. Furthermore, its sensitivity should be evaluated through controlled interventions with known intended effects, consistent with established practices for evaluating explanation and attribution methods \citep{adebayo2018sanity,wang2023evaluating,wang2024unlearning}.

\begin{figure}[t]
\centering
\resizebox{\linewidth}{!}{
\includegraphics[]{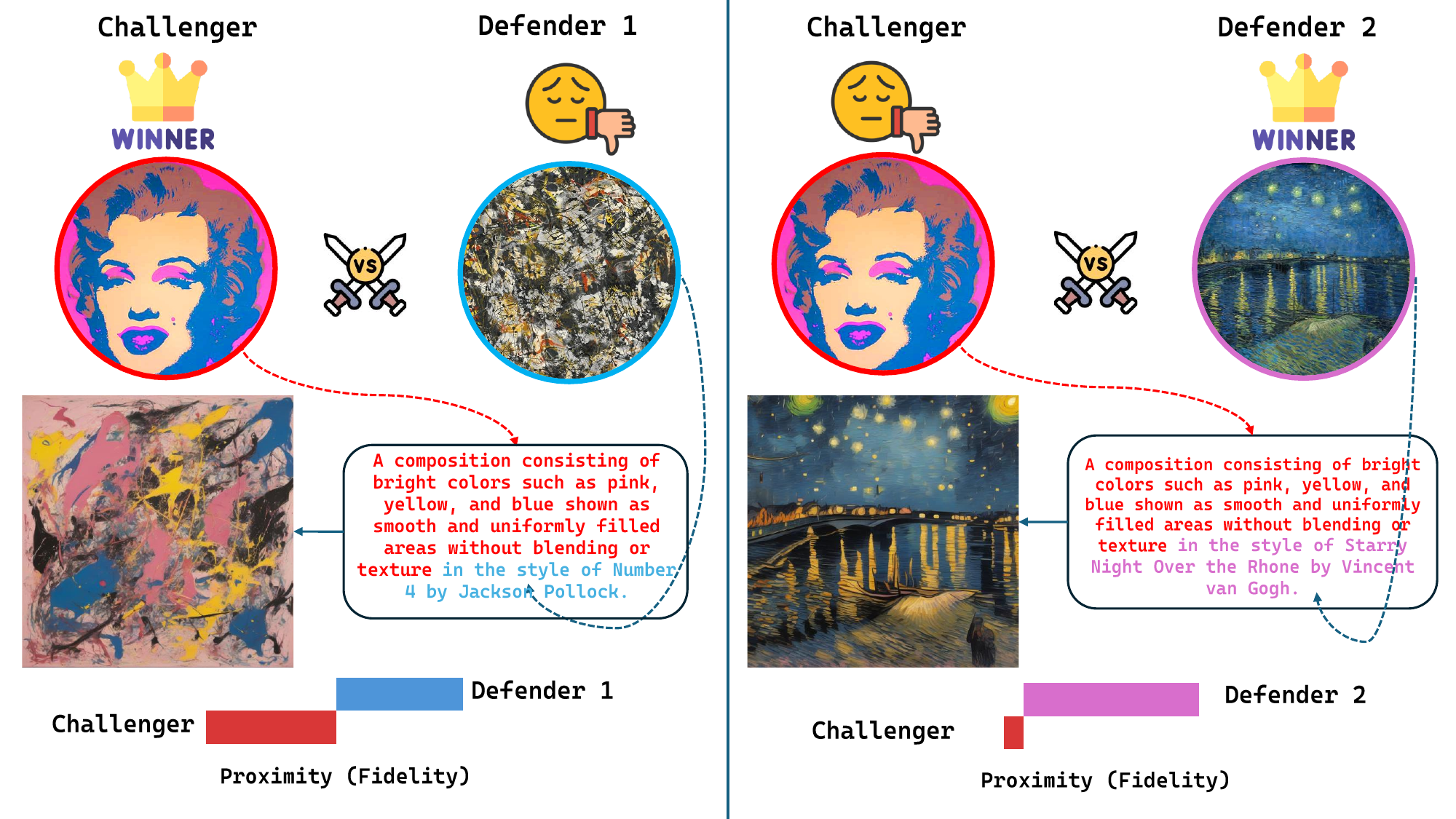}
}
\caption{Illustration of a \textit{Motif Duel} for SDXL, evaluated using fidelity-based proximity (CSD). The results demonstrate how combining the challenger's motifs with different defenders produces stylistically distinct outputs.}
\label{fig: battle_sdxl_fid}
\vspace{-5mm}
\end{figure}


\subsection{The ArtArena Protocol}
\label{sec:protocol}
\paragraph{\textbf{Notations summary.}}
{\textit{ArtArena}} is a protocol that makes stylistic influence in text-to-image systems measurable through controlled interactions between artworks. We present the pseudocode of \textit{ArtArena} in Algorithm~\ref{algo: art_arena} in the Appendix. The input \texttt{Artworks} is a list of records \{\texttt{title}, \texttt{artist}, \texttt{reference\_image}\}. The text‑to‑image generator is \texttt{Model} \(M\). The comparator \texttt{Proximity} is any measure that compares generated images to a \texttt{reference\_image}; we denote its evaluation generically by \(\operatorname{prox}\text{(}\cdot,\cdot\text{)}\). Global parameters are \(K\) (samples per prompt), \(R\) (rounds per match), \textit{fitness\_threshold} \((\tau_f)\) for entry, and \textit{margin\_delta} \((\delta)\) for round outcomes. The algorithm outputs \texttt{FitSet} (artworks passing the fitness test), \texttt{Matches} (pairwise results), and \texttt{Leaderboard} (a ranked map of stylistic influence). Motifs for a challenger \(c\) are obtained by \texttt{ExtractMotifs}(\(c\)) as an ordered list of phrases; the defender template for artwork \(d\) is the string:
\texttt{DefenderTemplate}($d$) = \texttt{``}$\langle\mathrm{title}$($d$)$\rangle \texttt{ in the style of }$ $\langle\mathrm{artist}$($d$)$\rangle\texttt{''.}$

\noindent{\textbf{Step 1: Entry Trials (Fitness Test).}}
This step establishes which artworks the model can imitate before any interaction between artworks is studied, assessing whether \(M\) can faithfully imitate an artwork’s distinctive style when attribution is explicit (refer to \cref{fig: imitation_example} in the Appendix). For each artwork \(w \in \texttt{Artworks}\), \texttt{Prompt}($w$) = \texttt{DefenderTemplate}($w$).
Generate \(K\) samples \( \texttt{Images} = \{x_1,\dots,x_K\}\) via \(M\text{(\texttt{Prompt}($w$))}\). Compute

{\small
\begin{equation}
\label{eq:fitness}
\mathrm{Fit}(w) = \frac{1}{K} \sum_{k=1}^{K}
\operatorname{prox}\!\big(x_k, \texttt{reference\_image}(w)\big)
\end{equation}}
If \(\mathrm{Fit}\text{($w$)} \ge \tau_f\), the artwork is added to \texttt{FitSet}. A high proximity score under explicit prompting is consistent with the model drawing on learned stylistic traces rather than generating from diffuse influence, a behavior consistent with memory and priming effects \citep{tulving1985many,stevens2008implicit}. High fitness shows that a model can reproduce an artwork when explicitly requested, making it a plausible candidate for unrequested reappearance. However, fitness alone does not establish that the artwork will prevail in other generations; this is examined through Motif Duels and the Influence Ledger. The artworks comprising the FitSet are presented in the Appendix (Tabs.~\ref{tab:SD_CLIP_fitness}--\ref{tab:SANA_CSD_fitness}).

\noindent{\textbf{Step 2: Motif Duels (Battleground).}}
This step examines how imitated artworks blend within the generations and whether one artwork is favored over another. In this stage, artworks admitted to \texttt{FitSet} compete in round-robin interactions, where each artwork faces every other as challenger and defender. The defender’s identity is anchored by its title and artist, while the challenger contributes motifs drawn from its own stylistic vocabulary. For each match, we sample combinations of motifs from the challenger and blend them with the defender’s template to compose prompts (see \cref{fig: battle_sdxl_fid}). Across \(R\) rounds, the model generates \(K\) images per prompt, and proximity scores are computed against both reference artworks:
{\small
\begin{equation}
\begin{aligned}
s_c(r) &= \frac{1}{K} \sum_{k=1}^{K}
\operatorname{prox}\!\big(x_{r,k}, \texttt{ref}(c)\big), \\
s_d(r) &= \frac{1}{K} \sum_{k=1}^{K}
\operatorname{prox}\!\big(x_{r,k}, \texttt{ref}(d)\big)
\end{aligned}
\end{equation}}
A round is awarded to the challenger if \(s_c(r) - s_d(r) > \delta\), and a round is awarded to the defender if \(s_d(r) - s_c(r) > \delta\). If neither side exceeds the other by more than \(\delta\), the round is a draw and counts for neither side. We write $W_\delta(c,d)$ for the number of rounds that challenger $c$ wins against defender $d$ at margin $\delta$. The artwork that wins more rounds claims the match and is credited with a match win in the \texttt{Leaderboard}. If both artworks win the same number of rounds, the match is a draw, and neither artwork is credited.

\noindent{\textbf{Step 3: Influence Ledger.}}
The Influence Ledger compares every artwork against all others to reveal which ones blend proactively, identifying artworks that exhibit greater stylistic dominance. After all pairwise matches are complete, the results are aggregated into a \texttt{Leaderboard} by ranking artworks according to their cumulative match wins, computed as the number of matches won across all roles in the round-robin interactions. Ties are broken by the number of matches won as challenger. 





\begin{insightbox}

Consistent challenger wins suggest that an artwork's characteristics can prevail without being named, making it a potential ``silent brush'' under generalized prompts. Conversely, consistent defender wins indicate that its characteristics withstand competing motifs when requested, suggesting a strong stylistic representation that may also resurface unprompted. 
\end{insightbox}

\newtcolorbox{resultbox}[1]{
   enhanced,
   colback=insightback,
   colframe=insightblue,
   coltitle=white,
   fonttitle=\bfseries,
   sharp corners,
   boxrule=0pt,
   leftrule=4pt,
   attach title to upper,
   title={}, 
   separator sign none,
   description font=\mdseries,
   arc=2mm,
   breakable,
   top=2mm,
   bottom=2mm,
   left=3mm,
   right=3mm,
}

\section{Case Study}
\label{sec:personalization}




\subsection{Motivation}
\label{sec:cs_motivation}

This case study investigates whether stylistic influences associated with canonical artworks persist when a text-to-image model is personalized on an artist's own artworks. Personalization may improve the model's ability to reproduce the artist's style without necessarily reducing the influence of canonical artworks favored by the base model; we therefore examine this behavior before and after personalization for two contemporary artists.

Following the Guiding Principles (Section~\ref{sec:guiding_principles}), we first identify canonical artworks exhibiting proactive blending in the base model (Section~\ref{sec:cs_phases}). We then assess whether personalization improves the imitation of the artist's artworks, whether these canonical artworks continue to dominate, and how this influence changes across margins (Section~\ref{sec:leakage_card}). Together, these analyses audit, artwork by artwork, whether previously learned stylistic influences interfere with the artist's intended style and persist after personalization.

\subsection{Audit Workflow}
\label{sec:cs_phases}

We structure the case study into two complementary stages. Throughout, we denote the base model by $\M$, the artist's collection of \newterm{non-canonical artworks} by $\Non$, and the personalized model obtained by adapting $\M$ on $\Non$ by $\Mp$. The \newterm{canonical artworks} $\Can$ are the well-known artworks most likely to leak into the artist's generations, shortlisted in the first stage.

\begin{auditstep}{Canonical Influence Profiling}{auditGreen}
The artist selects a collection of well-known artworks they consider potential leakage candidates, and a proximity measure of their choice that remains fixed throughout the audit. Applying the full ArtArena protocol (Section~\ref{sec:protocol}) to $\M$ produces an Influence Ledger, whose highest-ranked artworks form $\Can$.
\end{auditstep}

\begin{auditstep}{Personalization Leakage Assessment}{auditRose}
The artist personalizes $\M$ on their own artworks with a technique of their choice, assigning each artwork a \newterm{personalization prompt}. ArtArena is applied to both the models with the strategy: the canonical artworks act only as challengers and the artist's artworks only as defenders. A leakage card summarizes how canonical influence changes after personalization (Section~\ref{sec:leakage_card}).
\end{auditstep}

\subsection{Experimental setup}
\label{sec:exp}


\textbf{Artworks and generative models}: For Canonical Influence Profiling, we select the top twenty most popular artists from WikiArt \citep{wikiart_popular} (BSD-3 license), collecting 8{,}737 artworks likely represented in large-scale web corpora such as LAION \citep{schuhmann2022laion}, commonly used to train text-to-image models. These form the canonical collection. For Personalization Leakage Assessment, we study two contemporary artists, Trent Anthony Francis and Gwenn Seemel, neither of whom uses AI-generated imagery in their work. Both publicly support the reuse of their work, with Francis offering royalty-free usage under specified conditions\footnote{\url{https://www.trentanthonyfrancis.com/licensing}} and Seemel declaring their artwork to be in the public domain\footnote{\url{https://gwennseemel.com/uncopyright/}}. Five artworks from each artist form their respective non-canonical sets $\Non$: \emph{Essays on Nature and Growing (Path IV)}, \emph{Meadow May}, \emph{Star Strangled Banner}, \emph{The Moon and Sixpence}, and \emph{The Star Strangled Banner (Deviations)} by Francis; and \emph{American Bison}, \emph{Ladybugs on Daisies}, \emph{Sally with Wolves}, \emph{Siblings}, and \emph{Springtime in Lambertville} by Seemel. We evaluate three widely adopted text-to-image models: two U-Net-based diffusion models, \texttt{stable-diffusion-xl-base-1.0} (SDXL; CreativeML Open RAIL+M License) \citep{sdxl_base_1_0} and Stable Diffusion v1.5 (SD v1.5; CreativeML Open RAIL-M License) \citep{rombach2022high}, and one DiT-based diffusion model, SANA-1.5 (Apache 2.0 License) \citep{xie2025sana15}.

\textbf{Proximity metrics}: We measure proximity with three complementary metrics: CLIP cosine similarity \citep{radford2021learning,hessel2021clipscore} for high-level semantics and composition (Semantics), Learned Perceptual Image Patch Similarity (LPIPS) \citep{zhang2018lpips} for perceptual closeness in structure and texture (Aesthetics), and Contrastive Style Descriptors (CSD) \citep{somepalli2024measuring} for stylistic fidelity such as color, texture, and stroke patterns (Fidelity). For LPIPS, lower values indicate closer imitation, so all comparisons are reversed. Since each metric captures only part of style and has its own limitations, discussed in Sections~\ref{sec:ethical} and~\ref{sec:PM} of the Appendix along with dataset considerations, we use all three together for a balanced assessment of stylistic similarity.


\textbf{Motif extraction and blending}: We construct challenger templates via a two-stage pipeline. First, motif extraction identifies the content elements of each artwork (objects, structures, and spatial components) by retrieving commonly referenced names and descriptions from art-historical sources (\eg, museum and curatorial references), so that the motifs reflect established descriptive conventions. Second, motif blending enumerates all non-empty combinations of an artwork's $m$ motifs and converts each combination into a coherent, style-neutral scene description. From this pool, we sample $R$ challenger templates, one per round, and keep them fixed per artwork throughout the tournament to ensure consistent and comparable matchups. All stages are executed using GPT-4o; examples of extracted motifs and of the prompts obtained by blending them, the prompt templates for motif extraction and blending, the source list for motif descriptions, and additional details are provided in Sections~\ref{sec: motif_extraction} and~\ref{subsec:Artwork_Extracted_Motifs} of the Appendix.


\textbf{Canonical Influence Profiling configuration}: We run ArtArena (refer to Algorithm~\ref{algo: art_arena} in the Appendix for the pseudocode) once per model and proximity metric, using that metric in every stage of the run. Entry Trials generate $K = 5$ images per artwork, and the fitness threshold $\tau_f$ keeps the twenty artworks with the highest fitness under each metric, which limits the computational cost of pairwise comparisons and excludes noisy weak imitations while keeping enough artworks to capture diverse interactions in Motif Duels. Motif Duels use $R = 5$ rounds per match and $K = 1$ image per round. We build the Influence Ledgers at margin $\delta = 0$ (\cref{tab:fidelity_sdxl}; for others, refer to \cref{tab:sd15_ledgers,tab:Sana_clip_tr} in the Appendix) and vary $\delta$ to analyze their sensitivity (refer to Figs.~\ref{fig:SD_delta}--\ref{fig:SANA_delta} in the Appendix). The five highest-ranked artworks of each ledger form the canonical artworks $\Can$ for the corresponding model and metric.


\textbf{Personalization Leakage Assessment configuration}: For each artist and model, we train one DreamBooth LoRA adapter \citep{ruiz2023dreambooth,hu2022lora} covering all five of the artist's artworks, binding each artwork to its own identifier \texttt{<id>}, a string with little prior meaning under the model's text encoder, through the personalization prompt \texttt{painting in the style of <id>}. We retain as $\Mp$ the checkpoint with the highest average fitness on the artist's artworks across the three proximity metrics. Entry Trials then generate $K = 3$ images per artwork, computing the fitness of the artist's artworks with the personalization prompt. In Motif Duels, the defender template \texttt{in the style painting by <id>} anchors each prompt on the artist's artwork; each match has $R = 5$ rounds with $K = 1$ image per round under random seeds, giving $|\Can|\times|\Non| = 25$ matches per model and artist, evaluated at margins $\delta \in \{0, 0.05, 0.10, 0.15, 0.20\}$. The identifiers, training hyperparameters, and full configuration of both audit stages are summarized in the Appendix (\cref{tab:identifiers,tab:setup}).

\begin{table}[t]
\centering
\scriptsize
\setlength{\tabcolsep}{3pt}
\renewcommand{\arraystretch}{0.9}
\setlength{\aboverulesep}{0.25ex}
\setlength{\belowrulesep}{0.25ex}

\caption{Influence Ledgers from Motif Duels for \textbf{SDXL}, for \textbf{Semantics}, \textbf{Aesthetics}, and \textbf{Fidelity} (top to bottom). Ranks are assigned by total match wins, with challenger wins used as a tie-breaker.}
\label{tab:fidelity_sdxl}

\begin{tabularx}{\columnwidth}{@{}r>{\raggedright\arraybackslash}Xcc@{}}
\toprule
\textbf{Rank} & \textbf{Artwork (base model)} & \textbf{Chall.} & \textbf{Def.} \\
\midrule
\multicolumn{4}{@{}c}{\textit{Semantics (CLIP)}} \\ \hline
\addlinespace[1pt]
1  & Vincent van Gogh, \textit{Vincent's Bedroom in Arles} & 19 & 12 \\
2  & Claude Monet, \textit{Water Lilies} & 17 & 13 \\
3  & Vincent van Gogh, \textit{Wheat Field with Reaper and Sun} & 16 & 12 \\
16 & Andy Warhol, \textit{Marilyn Monroe} & 2 & 10 \\
17 & Michelangelo, \textit{David} & 4 & 7 \\
18 & Andy Warhol, \textit{After Marilyn Pink} & 2 & 9 \\
19 & Vincent van Gogh, \textit{Self Portrait with Felt Hat} & 1 & 10 \\
20 & Caravaggio, \textit{Incredulity of Saint Thomas} & 0 & 5 \\
\midrule
\multicolumn{4}{@{}c}{\textit{Aesthetics (LPIPS)}} \\ \hline
\addlinespace[1pt]
1  & Georgia O'Keeffe, \textit{Sky Above Clouds IV} & 17 & 18 \\
2  & Jackson Pollock, \textit{Black \& White (Number 20)} & 15 & 17 \\
3  & Georgia O'Keeffe, \textit{Slightly Open Clam Shell} & 17 & 14 \\
16 & Vincent van Gogh, \textit{Mulberry Tree} & 1 & 10 \\
17 & Jackson Pollock, \textit{Eyes in the Heat} & 1 & 9 \\
18 & Rembrandt, \textit{Three female heads with one sleeping} & 2 & 7 \\
19 & Michelangelo, \textit{Bust of Brutus} & 0 & 7 \\
20 & Michelangelo, \textit{Design for Julius II tomb (first version)} & 0 & 4 \\
\midrule
\multicolumn{4}{@{}c}{\textit{Fidelity (CSD)}} \\ \hline
\addlinespace[1pt]
1  & Vincent van Gogh, \textit{Wheat Fields at Auvers Under Clouded Sky} & 14 & 18 \\
2  & Vincent van Gogh, \textit{Wheat Field with Reaper and Sun} & 11 & 14 \\
3  & Vincent van Gogh, \textit{View of Vessenots near Auvers} & 9 & 14 \\
16 & Andy Warhol, \textit{Marilyn Monroe} & 3 & 12 \\
17 & Andy Warhol, \textit{Three Marilyns} & 1 & 14 \\
18 & Vincent van Gogh, \textit{Self Portrait with Palette} & 2 & 12 \\
19 & Andy Warhol, \textit{After Marilyn Pink} & 1 & 13 \\
20 & Jackson Pollock, \textit{Number 4} & 1 & 12 \\
\bottomrule
\end{tabularx}
\end{table}

\subsection{Leakage Card}
\label{sec:leakage_card}

We introduce the \newterm{leakage card} as a structured summary of the audit findings, enabling artists to interpret the extent of stylistic leakage in their personalized models. It consolidates the observations into three complementary blocks (\cref{fig:LC_SDXL_Trent_CSD}), providing an artwork-level assessment of stylistic interference and its changes following personalization.

\noindent \textbf{Block A: Personalization recall.}
After personalization, an artist expects $\Mp$ to imitate their artworks more faithfully than $\M$. Block~A checks this through the change in fitness (\cref{eq:fitness})
\begin{equation}
\label{eq:delta_fit}
\Delta\Fit(w) = \Fit_{\Mp}(w) - \Fit_{\M}(w), \qquad w \in \Can \cup \Non,
\end{equation}
together with the set means $\overline{\Fit}(S) = |S|^{-1}\sum_{w \in S}\Fit(w)$ for $S \in \{\Can,\Non\}$ on both models. A positive change on $\Non$ confirms that personalization has taken hold, while the change on $\Can$ shows whether the canonical artworks lost strength in the process.

\noindent \textbf{Block B: Artwork leakage potential.}
Stronger recall of the artist's artworks does not imply that canonical styles no longer blend into them. Block~B therefore resolves the targeted Motif Duels on $\Mp$ to individual artworks. At margin $\delta$, we define the \newterm{leakage potential} of a canonical artwork and the \newterm{prominence capacity} of a non-canonical artwork as
\begin{align}
  \LP_\delta(c) &= \frac{100}{|\Non|}\sum_{d \in \Non}\ind[\,c \text{ wins against } d \text{ at } \delta\,],\\
  \PC_\delta(d) &= \frac{100}{|\Can|}\sum_{c \in \Can}\ind[\,d \text{ wins against } c \text{ at } \delta\,],
\end{align}
expressed as percentages of matches won. Block~B reports both quantities at $\delta = 0$, written $\LP(c)$ and $\PC(d)$. $\LP(c)$ identifies the canonical artworks that remain the strongest leakage candidates. $\PC(d)$ identifies the artist's artworks that best retain their stylistic identity, which are the safest to generate around.


\noindent \textbf{Block C: Personalization's effect on leakage potential.}
After personalization, an artist also expects canonical styles to prevail less often in generations centered on their work. Block~C checks this across all canonical artworks collectively by comparing $\M$ and $\Mp$, counting rounds rather than matches, since a match outcome hides how close its rounds were. Using the round counts $W_\delta(c,d)$ of Motif Duels (Section~\ref{sec:protocol}), we define the \newterm{Challenger Win Rate} as the percentage of all rounds won by the challengers,
\begin{equation}
  \CWR(\delta) = \frac{100}{|\Can|\,|\Non|\,R}\sum_{c \in \Can}\sum_{d \in \Non} W_\delta(c,d).
\end{equation}
Drawn rounds count as played but not as won, so the denominator is fixed and $\CWR(\delta)$ cannot increase as $\delta$ grows; we refer to $\delta$ as the \newterm{strictness} of a round. Block~C reports $\CWR_{\M}(\delta)$ and $\CWR_{\Mp}(\delta)$ across strictness levels, together with the \newterm{drop in leakage potential}
\begin{equation}
  \Delta\CWR(\delta) = \CWR_{\M}(\delta) - \CWR_{\Mp}(\delta),
\end{equation}
in percentage points, where a positive drop indicates that personalization has weakened canonical blending. If $\CWR_{\Mp}(\delta)$ falls faster than $\CWR_{\M}(\delta)$ as $\delta$ grows, the canonical round wins that remain after personalization are narrow and do not survive a stricter margin.

\subsection{Insights}
\label{sec:cs_insights}
We present the case study findings from two phases: \textbf{(A) Canonical Influence Profiling} and \textbf{ (B) Personalization Leakage Assessment}, analyzing the four guiding research questions across base-model behavior and personalization.


\paragraph{\underline{(A) Canonical Influence Profiling.}}

\textbf{Artwork imitation varies across models and proximity metrics (Q1).} \textit{Entry Trials} reveal artwork-level variation in imitation across models and proximity metrics (refer to \cref{fig: imitation_box_plot} in the Appendix). SD v1.5 achieves stronger imitation across all three proximity dimensions, while SDXL remains competitive in Semantics, marginally outperforming SD v1.5 among its top-10 artworks. The composition of \texttt{FitSets} also varies across metrics: Monet's artworks are prominent in SD v1.5 Semantics, Pollock's in Aesthetics, and van Gogh's in Fidelity. In contrast, SDXL favors van Gogh in Semantics and Fidelity, whereas SANA-1.5 features Warhol in Semantics and Fidelity and Rembrandt in Aesthetics. 

Notably, van Gogh's artworks recur across multiple models and dimensions, although their relative rankings vary. These differences may arise from the distinct visual characteristics captured by each metric and variations in learned model representations. 

\textit{Overall, imitation is selective rather than uniform, suggesting that canonical artwork candidates must be identified separately for each model and proximity dimension.} 


\textbf{Blending selectively preserves competing artwork characteristics (Q2).}
The influence of challenger motifs depends on the requested defender, with different visual characteristics prevailing across interactions (\cref{fig: battle_sdxl_fid}; refer to Figs.~\ref{fig: battle}--\ref{fig: battle_sana_aes} in the Appendix). Under SD v1.5 Semantics, Monet's \textit{Sunset on the Seine at Lavacourt, Winter Effect} retains its recognizable landscape composition when paired with other Monet artworks, while incorporating their distinctive brushwork, colors, and reflections. However, when paired with Pollock's \textit{Number 3} and \textit{Number 48}, the generated images continue to resemble Monet's landscapes, with little evidence of the defenders' characteristic drip-painting aesthetics.

\noindent 
Conversely, defenders with distinctive visual motifs may overshadow challenger characteristics, reinforcing the influence of explicit artwork conditioning (refer to \cref{fig: battle_sd_sem,fig: battle_sdxl_sem,fig: battle_sana_fid} in the Appendix).

\noindent \textit{These observations suggest that, models may retain a challenger's
recognizable characteristics while leaving those of the intended defender largely absent.}


\textbf{Proactive Influence Extends Beyond Artwork Imitation Fitness (Q3).} 

The \textit{Influence Ledgers} reveal a notable disconnect between imitation fitness and proactive blending. Under SDXL, the leading artworks in Semantics and Fidelity rank only 16th and 15th in \textit{Entry Trials}, respectively, while SANA-1.5's Fidelity leader ranks 17th. Thus, artworks that are not imitated most faithfully can nevertheless prevail over explicitly identified defenders.

\noindent Furthermore, each model--metric combination exhibits preferences for particular artists: van Gogh's artworks occupy all top-three Fidelity positions under SD v1.5 and SDXL, whereas Pollock and Warhol account for seven of the nine top-three positions across proximity dimensions under SANA-1.5 (\cref{tab:fidelity_sdxl}; refer to \cref{tab:sd15_ledgers,tab:Sana_clip_tr} in the Appendix).

\noindent Notably, challenger wins exceed defender wins among leading artworks in several settings, providing evidence consistent with the \textit{Silent Brush} phenomenon. However, challenger victories generally become less widespread at stricter comparison margins $\delta$, except under SANA-1.5 Aesthetics. 

\noindent \textit{These findings show that potential leakage candidates cannot be identified from imitation fitness alone; their relative stylistic dominance must be assessed through \textit{Motif Duels}}.

\paragraph{\underline{ (B) Personalization Leakage Assessment.}}
We next discuss our observations around the personalized model for the artist. We present the insights as follows: 

\textbf{Personalization improves artist imitation, but canonical preferences persist (Q1).}  Block~A of \cref{fig:LC_SDXL_Trent_CSD} shows that personalization improves imitation fitness for all artworks by Trent Anthony Francis and Gwenn Seemel, while reducing the average fitness of shortlisted canonical artworks (refer to Figs.~\ref{fig:LC_SD_Trent_Clip}--\ref{fig:LC_SDXL_Gwenn_CSD} in the Appendix for other settings). 

However, the relative balance differs across models. Under SDXL, both artists' collections surpass the canonical collection in average fitness across all three proximity dimensions. In contrast, SANA-1.5 retains a canonical advantage in most settings despite improved artist imitation, while SD v1.5 exhibits a similar persistence in Fidelity. 

\noindent These findings suggest that learning an artist's artworks does not necessarily diminish previously learned canonical preferences.


\textbf{Canonical challengers can prevail
despite successful personalization (Q2--Q3).}

Block~B reveals that SDXL's greater responsiveness to personalization, observed in \textit{Entry Trials}, extends to \textit{Motif Duels}, where the artists' artworks generally exhibit greater prominence (\cref{fig:LC_SDXL_Trent_CSD}). Conversely, for SD v1.5 and SANA-1.5, canonical challengers continue to dominate. This persistence is particularly pronounced in Aesthetics, where all of Francis's artworks lose every defender match under both SD v1.5 and SANA-1.5, despite improved imitation fitness (refer to \cref{fig:LC_SD_Trent_LPIPS,fig:LC_SANA_Trent_LPIPS} in the Appendix).

The disconnect between imitation and proactive blending observed in Canonical Influence Profiling also persists after personalization: under SD v1.5 Semantics, \textit{Madonna Litta} remains the strongest canonical leakage candidate for both artists despite experiencing the largest reduction in canonical fitness. Furthermore, resistance to canonical influence varies across individual artworks. Seemel's \textit{Siblings}, with its distinctive inverted-house motif, consistently retains some prominence, echoing our earlier observation that distinctive defender motifs can withstand competing influences (Q2). In contrast, Francis's \textit{The Star Strangled Banner (Deviations)} frequently yields to canonical challengers (refer to Figs.~\ref{fig:LC_SD_Trent_Clip}--\ref{fig:LC_SANA_Trent_LPIPS} in the Appendix). \textit{Thus, artists must assess not only which canonical influences persist after personalization, but also which of their own artworks remain susceptible to them.}



\textbf{Canonical challenger wins decline relatively more
under SD v1.5 and SDXL (Q4).} 
Block~C further corroborates SDXL's responsiveness to personalization (\cref{fig:LC_SDXL_Trent_CSD}), with both SDXL and SD v1.5 exhibiting substantial reductions in canonical Challenger Win Rate, particularly in Fidelity. Conversely, SANA-1.5 retains considerable canonical influence in Semantics and Fidelity, even at stricter margins, and exhibits increased challenger victories at certain margins in Aesthetics (refer to Figs.~\ref{fig:LC_SANA_Trent_Clip}--\ref{fig:LC_SANA_Gwenn_LPIPS} in the Appendix). Notably, Seemel experiences greater reductions than Francis, suggesting that personalization effectiveness also depends on the artist's collection. \textit{Thus, artists must assess both the frequency and strength of residual canonical influence after personalization.}

\section{Validating ArtArena} 
\label{sec:Val_AA}
We evaluate whether ArtArena captures the expected changes in imitation and relative stylistic dominance under controlled strengthening and weakening interventions. We select the bottom-1 and top-1 artworks from the Influence Ledgers obtained during Canonical Influence Profiling as targets for strengthening and weakening, respectively. Starting from the same base model $\M$, we construct two modified models through the following interventions:

\begin{itemize}[leftmargin=*, itemsep=2pt]
    \item \textbf{Strengthening}: We fine-tune $\M$ on the bottom-1 artwork using DreamBooth \citep{ruiz2023dreambooth}.

    \item \textbf{Weakening}: We unlearn the top-1 artwork from $\M$ using SurgUn \citep{ranjan2026forgetting}, which applies targeted weight-space updates to remove specific visual concepts from text-conditioned diffusion models.
\end{itemize}
Both interventions use LoRA adapters \citep{hu2022lora} of rank $64$ and the training prompt \texttt{<title> in the style of <artist>}, matching the artwork's defender template in Motif Duels. 
\vspace{-4mm}
\paragraph{Validation measurements.} To assess the effects of these interventions, we repeat Entry Trials and Motif Duels on each modified model $\tilde{\M}$ over the original twenty-artwork \texttt{FitSet} $\mathcal{F}$, retaining the Canonical Influence Profiling configuration. We then compare each intervention target $w$ against its standing in the base model $\M$ along two complementary dimensions: (i) fitness shifts, measured through changes in fitness and imitation rank, and (ii) relative stylistic dominance, assessed through changes in defender loss rate.

\begin{table}[t]
\centering
\scriptsize
\setlength{\tabcolsep}{2.5pt}
\renewcommand{\arraystretch}{1.0}
\caption{\textbf{Controlled validation of ArtArena under SDXL for Fidelity.}
Top: Changes in imitation rank and fitness following strengthening of the bottom-ranked artwork and weakening of the top-ranked artwork from the Influence Ledger.
Bottom: Defender Loss Rate (DLR) before and after each intervention, along with the corresponding changes across comparison margins $\delta$.}
\label{tab:val_sdxl_csd_imit}
\begin{tabularx}{\columnwidth}{@{}>{\raggedright\arraybackslash}X c c c c c c@{}}
\toprule
\multirow{2}{*}{\textbf{Target artwork}} & \multicolumn{3}{c}{\textbf{Imitation rank}} & \multicolumn{3}{c}{\textbf{Fitness score}} \\
\cmidrule(lr){2-4}\cmidrule(l){5-7}
 & Base & New & Shift & Base & New & Change \\
\midrule
Jackson Pollock, \textit{Number 4}\newline(Fine-tuning, bottom-1) & 18 & 13 & \valgood{+5} & 0.7899 & 0.8300 & \valgood{+0.0402} \\
\addlinespace[2pt]
Vincent van Gogh, \textit{Wheat Fields at Auvers Under Clouded Sky}\newline(Unlearning, top-1) & 15 & 20 & \valgood{$-$5} & 0.7976 & 0.7316 & \valgood{$-$0.0660} \\
\bottomrule
\end{tabularx}

\vspace{0.3em}
\label{tab:val_sdxl_csd_def}
\setlength{\tabcolsep}{4pt}
\begin{tabularx}{\columnwidth}{@{}>{\raggedright\arraybackslash}X c c c c c@{}}
\toprule
\multirow{2}{*}{\textbf{Target artwork / Model}} & \multicolumn{5}{c}{\textbf{Defender loss rate (\%) at margin $\delta$}} \\
\cmidrule(l){2-6}
 & 0.00 & 0.05 & 0.10 & 0.15 & 0.20 \\
\midrule
\multicolumn{6}{@{}c}{Jackson Pollock, \textit{Number 4} (bottom-1)} \\ \hline
\quad Base & 47.4 & 36.8 & 29.5 & 20.0 & 13.7 \\
\quad Fine-tuned & 80.0 & 26.3 & 11.6 & 10.5 & 9.5 \\
\quad Change & \valbad{+32.6} & \valgood{$-$10.5} & \valgood{$-$17.9} & \valgood{$-$9.5} & \valgood{$-$4.2} \\
\midrule
\multicolumn{6}{@{}c}{Vincent van Gogh, \textit{Wheat Fields at Auvers Under Clouded Sky} (top-1)} \\ \hline
\quad Base & 16.8 & 6.3 & 2.1 & 0.0 & 0.0 \\
\quad Unlearned & 82.1 & 68.4 & 66.3 & 62.1 & 58.9 \\
\quad Change & \valgood{+65.3} & \valgood{+62.1} & \valgood{+64.2} & \valgood{+62.1} & \valgood{+58.9} \\
\bottomrule
\end{tabularx}
\end{table}
\textbf{Fitness shift}: We measure the change in the target artwork's fitness, $\Delta\Fit(w)$, as in \cref{eq:delta_fit} with $\tilde{\M}$ in place of $\Mp$, and the change in its imitation rank:

{\small
\begin{equation}
\label{eq:val_imitation}
\Delta\rho(w) = \rho_{\M}(w) - \rho_{\tilde{\M}}(w),
\end{equation}
}

where $\rho_{\M}(w)$ denotes the rank of artwork $w$ among $\mathcal{F}$ by fitness under $\M$, with rank~1 being the best imitated; a positive $\Delta\rho(w)$ thus means that $w$ moved up the ranking.

\textbf{Defender loss rate}: Since both interventions use the artwork's defender template, we additionally examine whether its ability to resist challengers changes. Using the round counts $W_\delta(c,w)$ of Motif Duels (Section~\ref{sec:protocol}), we define the \newterm{defender loss rate} ($\DLR$) of $w$ as the percentage of rounds in which challengers defeat it while it serves as the defender:

{\small
\begin{equation}
\label{eq:val_dlr}
\begin{aligned}
\DLR_{\M}(w,\delta)
&= \frac{100}{(|\mathcal{F}|-1)R}
\sum_{c\in\mathcal{F}\setminus\{w\}}
W_\delta(c,w),\\
\Delta\DLR(w,\delta)
&= \DLR_{\tilde{\M}}(w,\delta)
 - \DLR_{\M}(w,\delta).
\end{aligned}
\end{equation}
}

Each artwork defends against $|\mathcal{F}|-1=19$ challengers over $R=5$ rounds, giving $95$ rounds in total. We evaluate $\Delta\DLR(w,\delta)$ in percentage points at margins $\delta\in\{0,0.05,0.10,0.15,0.20\}$. Strengthening is expected to increase fitness, improve rank, and reduce defender losses ($\Delta\Fit(w)>0$, $\Delta\rho(w)>0$, $\Delta\DLR(w,\delta)<0$), whereas weakening is expected to do the opposite.
\vspace{-5mm}
\paragraph{Observations.} Fine-tuning raises the target artwork's fitness, and unlearning lowers it. In nearly all cases, its imitation rank improves or declines accordingly. Importantly, these interventions also affect the targeted artworks' ability to resist challengers. Fine-tuning generally reduces defender losses across margins, whereas unlearning generally increases them (\cref{tab:val_sdxl_csd_imit}; refer to Tabs.~\ref{tab:val_sd_clip_imit}--\ref{tab:val_sana_csd_def} in the Appendix). For instance, under SDXL Fidelity, unlearning \textit{Wheat Fields at Auvers Under Clouded Sky} substantially increases its DLR across all margins, while fine-tuning \textit{Number 4} reduces its defender losses at stricter margins (\cref{tab:val_sdxl_csd_imit}). \textit{These findings show that ArtArena captures the effects of targeted model modifications on an artwork's ability to resist competing artistic influences, beyond changes in imitation fitness alone.}

An exception arises under SD v1.5 Fidelity, where unlearning Vincent van Gogh's \textit{Olive Grove} reduces its imitation fitness and rank as expected, but also lowers its DLR at smaller comparison margins (refer to \cref{tab:val_sd_csd_imit,tab:val_sd_csd_def} in the Appendix).A possible explanation lies in the stylistic similarities among van Gogh's artworks, several of which SD v1.5 imitates with high fidelity (Section~\ref{sec:cs_insights}). Although unlearning weakens the model's ability to reproduce \textit{Olive Grove}, some stylistic characteristics may persist. These include color composition and texture, which are also shared with other van Gogh artworks not targeted for unlearning and may therefore remain represented in the model. Visual inspection suggests that these shared characteristics remain perceptible in the blended generations of Motif Duels, where they may still be captured by the CSD proximity metric, potentially explaining the unexpected reduction in DLR. \textit{This exception highlights the importance of interpreting measured stylistic influence in relation to the capabilities and limitations of the chosen proximity metric.}

\section{Conclusion}
The \textit{Silent Brush} phenomenon, where text-to-image models resurface learned artistic influences unprompted, is challenging to audit. Our proposed ArtArena, validated by controlled interventions, audits it through artwork interactions and reveals leakage that varies across models and aspects of style. Our observations show that this leakage persists even after personalization. Hence, artists should audit their personalized models before use, and future evaluations of generative models should examine not only what they reproduce but also how learned artistic influences propagate within them.

{
    \small
    \bibliographystyle{ieeenat_fullname}
    \bibliography{iclr2026_conference}
}

\clearpage
\appendix

\appendix

\renewcommand{\thesubsection}{A.\arabic{subsection}}
\renewcommand{\thesubsubsection}{A.\arabic{subsection}.\arabic{subsubsection}}

\section*{Appendix}
\addcontentsline{toc}{section}{Appendix}

\etocsettocstyle{\subsection*{Contents}}{}
\etocsetnexttocdepth{subsection}
\localtableofcontents

\subsection{Ethical Considerations and Limitations}
\label{sec:ethical}

Our work aims to make stylistic influence in generative models measurable, but studying this phenomenon introduces important ethical and methodological challenges. Artistic style is historically and culturally situated, and any computational representation necessarily simplifies complex artistic practices into measurable signals. Consequently, the influence scores produced by ArtArena should not be interpreted as objective measures of originality, artistic value, or ownership.

ArtArena relies on proxy similarity metrics such as CLIP similarity, LPIPS, and CSD, each of which captures only partial aspects of style. CLIP-based measures may conflate semantics with style, LPIPS can be sensitive to structural perturbations, and CSD inherits biases from its training data. Although we use multiple complementary metrics, the resulting measurements remain approximations of stylistic influence rather than direct evidence of memorization or copyright infringement.

The motif extraction pipeline also introduces limitations. Motifs are derived using GPT-4o and external art-historical descriptions, which may reflect curatorial bias, omissions, or uneven documentation. Converting motifs into prompts further introduces prompt-design effects that may influence generations independently of the model’s latent representations.

Our experiments focus on highly popular WikiArt artists, motivated by their likely presence in large-scale web datasets. While practical, this limits coverage of underrepresented artistic traditions, regions, and mediums. In addition, ArtArena studies stylistic influence primarily through pairwise interactions, whereas real-world generations may involve many interacting stylistic sources simultaneously.

The framework measures relative stylistic dominance but does not establish causal attribution at the parameter or dataset level. Higher leaderboard rankings indicate stronger reappearance under our protocol, but do not reveal whether this behavior arises from memorization, distributed generalization, dataset frequency, or architectural biases. Similarly, because several contemporary models do not disclose training data, we cannot directly verify the presence of specific artworks in their training corpora.

Beyond technical limitations, there are broader ethical concerns regarding how such evaluations may be interpreted or used. Influence scores could be misused to rank styles competitively, justify appropriation, or reinforce cultural hierarchies already present in web-scale datasets. Moreover, analysis of stylistic persistence may indirectly expose properties of underlying training datasets. To mitigate these risks, ArtArena is intended as a diagnostic framework for transparency, auditing, and governance rather than enforcement or adjudication. Its goal is to support informed discussion around attribution, consent, dataset curation, and cultural equity while avoiding prescriptive judgments about artistic legitimacy or ownership.

Finally, our work does not make legal claims regarding copyright infringement, fair use, or authorship. ArtArena identifies measurable patterns of stylistic persistence and interaction, but translating these observations into legal or normative conclusions depends on broader cultural, regulatory, and jurisdiction-specific considerations outside the scope of this paper.

\subsection{Related Work}
\label{sec:genai_models_datasets}
\paragraph{Imitation and Memorization}
A first line of work evaluates whether diffusion models can reproduce training images with high fidelity. Retrieval-based audits generate multiple samples per prompt and identify memorized instances through pixel-level or embedding-based similarity \citep{somepalli2023diffusion,carlini2021extracting}. Complementary analyses connect memorization to optimization regimes, overparameterization, and data duplication in generative models \citep{zhang2017understanding,feldman2020does,gu2023memorization}. Together, these approaches establish that diffusion models can memorize and reproduce training content at the instance level.

However, these evaluations rely on a one-to-one retrieval paradigm, where each generated image is matched to a single reference. While effective for detecting near-duplicate reproduction, this setup cannot capture how multiple training samples jointly contribute to a single generation. In particular, retrieval-based audits cannot measure the relative influence of different sources within the same output, nor can they detect stylistic patterns that persist without producing an explicit duplicate.

\paragraph{Style Representation and Generative Control}
A second line of work evaluates stylistic imitation rather than exact duplication. Methods based on CLIP classification or learned style embeddings quantify whether generated images align with a target artistic style, including approaches that use CLIP-based similarity or classification \citep{radford2021learning} and learned style descriptors such as Contrastive Style Descriptors (CSD) \citep{somepalli2024measuring}. Defensive approaches such as Glaze aim to disrupt such alignment by perturbing feature representations \citep{shan2023glaze}. These approaches provide useful tools for measuring stylistic similarity at the level of individual images.

However, style similarity is typically computed as a single-image embedding comparison, which conflates style with shared semantic content or layout. Prior work on neural style transfer and diffusion-based conditioning shows that reliable stylization requires separating content structure from style statistics \citep{gatys2015neural,portilla2000parametric,rombach2022high,saharia2022photorealistic}. Where this style and content separation is an important requirement while evaluating style, alone these methods cannot provide a complete study to evaluate whether styles reappear across prompts without explicit attribution or how stylistic signals interact when multiple sources are present.

\begin{algorithm}[t]
\caption{\textit{Art Arena}}
\label{algo: art_arena}
\footnotesize
\SetKwInOut{Input}{Input}\SetKwInOut{Output}{Output\!:}
\Input{%
  \textbf{Artworks:} list of \{\texttt{title}, \texttt{artist}, \texttt{reference\_image}\}.\\
  \textbf{Model $M$:} text-to-image generator.\\
  \textbf{Proximity:} calibrated comparator for generated vs. reference images.\\
  \textbf{Params:} $K$ (generations), $R$ (rounds), $\tau_f$ (fitness threshold), $\delta$ (margin).
}
\Output{FitSet, Matches, Leaderboard}

\StepHeader{StepOneBg}{Step 1: Early Trials (Fitness Test)}
\ForEach{$w$ in Artworks}{
  Prompt $\leftarrow$ ``\texttt{<title(w)> in the style of <artist(w)>}''\;
  Images $\leftarrow$ Generate $K$ samples via $M$(Prompt)\;
  Fit $\leftarrow$ Average(Proximity(Images, reference\_image($w$)))\;
  \If{Fit $\ge$ $\tau_f$}{FitSet.add($w$)}
}

\StepHeader{StepTwoBg}{Step 2: Motif Duels (Battleground)}

\ForEach{$w$ in FitSet}{ 
Score[$w$] $\leftarrow 0$\;
Motifs $\leftarrow$ ExtractMotifs($w$)\;
MotifSet[$w$] $\leftarrow$ sample $r$ motifs from Motifs\;
}
\ForEach{ordered pair $(c,d)$ in FitSet, $c \neq d$}{
  DefenderTemplate $\leftarrow$ ``\texttt{<title(d)> in the style of <artist(d)>}''\;
  $wins\_c \leftarrow 0$; $wins\_d \leftarrow 0$\;
  \For{$r \leftarrow 1$ \KwTo $R$}{
    Selected $\leftarrow$ select $r^\text{th}$ motif from MotifSet[$c$]\;
    Prompt$_r$ $\leftarrow$ ``\texttt{<Selected as phrases>}'' $+$ DefenderTemplate\;
    Images$_r$ $\leftarrow$ Generate $K$ samples via $M$(Prompt$_r$)\;
    prox$_c$ $\leftarrow$ Average(Proximity(Images$_r$, reference\_image($c$)))\;
    prox$_d$ $\leftarrow$ Average(Proximity(Images$_r$, reference\_image($d$)))\;
    \If{(prox$_c$ $-$ prox$_d$) $>$ $\delta$}{ $wins\_c${+}{+} }
    \If{(prox$_d$ $-$ prox$_c$) $>$ $\delta$}{ $wins\_d${+}{+} }
  }
  \eIf{$wins\_c > wins\_d$}{Matches.append(\{pair: $(c,d)$, winner: $c$\}); Score[$c$]{+}{+}}
                           {\eIf{$wins\_d > wins\_c$}{Matches.append(\{pair: $(c,d)$, winner: $d$\}); Score[$d$]{+}{+}}
                                                   {Matches.append(\{pair: $(c,d)$, winner: ``draw''\})}}
}

\StepHeader{StepThreeBg}{Step 3: Influence Ledger (Leaderboard)}
Ledger $\leftarrow$ SortByDescendingScore(Score)\;
\Return \{FitSet, Matches, Ledger\}\;
\end{algorithm}
\paragraph{Membership Inference and Data Leakage Signals}
A third line of work studies leakage through membership inference attacks, which estimate whether a given sample was part of the training dataset \citep{shokri2017membership,hayes2019logan,pang2023black,pang2025white}. These methods typically assign scalar or binary membership scores to individual samples, often relying on auxiliary models or labeled member/non-member data.

While membership scores indicate that training signals persist in the model, they remain inherently instance-level and do not capture how generative outputs are formed through the combination of multiple learned patterns. Diffusion models synthesize images through iterative denoising over distributed latent representations, where multiple sources can jointly influence the final output \citep{ho2020denoising,rombach2022high,gu2023memorization}. Consequently, a high membership score does not imply that a corresponding style will consistently reappear during generation. More importantly, these methods cannot measure whether stylistic patterns emerge without explicit prompting, nor can they characterize how styles compete, combine, or dominate across different generation contexts.

\paragraph{Image Generation Models and Training Datasets}
Large-scale text-to-image diffusion models such as Stable Diffusion v1.5 and Stable Diffusion XL (SDXL) are publicly documented to have been trained on subsets of the LAION-5B dataset, a web-scale corpus of image--text pairs constructed from Common Crawl URLs~\citep{rombach2022high,schuhmann2022laion}. Official model documentation confirms that Stable Diffusion v1.x models were trained on filtered subsets including LAION-2B-en, LAION-high-resolution, and LAION-Aesthetics~\citep{rombach2022high}. The use of LAION-derived data has also been acknowledged in legal proceedings, including \textit{Andersen et al. v. Stability AI} in the United States and \textit{Getty Images v. Stability AI} in the United Kingdom, where it was treated as undisputed that Stable Diffusion models were trained on large-scale web-scraped image--text datasets containing copyrighted material~\citep{andersen2023complaint,lindberg2024applying,getty2023uk,coulter2024aiming}. While SDXL does not release a granular dataset manifest, Stability AI has stated that SDXL follows the same large-scale LAION-based data curation paradigm as earlier Stable Diffusion models~\citep{stabilityai2023sdxl}. In contrast, newer models such as SANA-1.5 do not publicly disclose their training datasets, reflecting a broader lack of dataset transparency in contemporary foundation model research rather than contradicting the established role of LAION-style corpora.

\begin{figure*}[t]
\centering
\small
\begin{tcolorbox}[
  enhanced jigsaw,
  unbreakable,
  colback=blue!5!white,
  colframe=blue!75!black,
  title=Motifs Extraction Prompt
]
\textbf{Instruction.}
Analyze the artwork \texttt{"\{artwork\_name\}"} by \texttt{"\{artist\_name\}"} and identify the most important motifs present in it.
Consult reliable sources (\texttt{\{source\}}) to obtain the commonly used motif names and concise descriptions.
Return the output in the following structure:

\medskip
\noindent\texttt{[}\\
\hspace*{1.2em}\texttt{\{artwork\_name\}: [}\\
\hspace*{2.8em}\texttt{\{motif1: description\},}\\
\hspace*{2.8em}\texttt{\{motif2: description\},}\\
\hspace*{2.8em}\texttt{...}\\
\hspace*{1.2em}\texttt{]}\\
\texttt{]}
\end{tcolorbox}
\vspace{-2pt}
\caption{\textbf{Motif-Extraction Prompt}: Prompt used to extract motif names and natural‑language motif descriptions for an artwork, based on terminology commonly used in reliable art‑historical sources. The sources used are mentioned in Figure \ref{fig:art_sources}.}
\label{fig:motif_extraction}
\end{figure*}
The LAION datasets were constructed through automated scraping of publicly accessible web images and associated text using Common Crawl, followed by weakly supervised filtering based on CLIP similarity, language detection, resolution constraints, and heuristic classifiers for aesthetic quality and NSFW content. Although LAION distributes only URLs and captions rather than hosting images directly, multiple studies note that semantic and stylistic information from the underlying images is nonetheless absorbed into model parameters during training~\citep{rombach2022high,carlini2023extracting}. This emphasis on scale and diversity over manual curation results in datasets containing artworks, illustrations, photographs, stock imagery, and user-generated content spanning a wide range of artistic styles and creators.

Subsequent analyses and investigations have highlighted significant issues with this data collection strategy. Empirical audits and journalistic investigations have shown that LAION subsets contain copyrighted artworks, recognizable characters, and works attributable to living artists, often without consent or licensing~\citep{jiang2023copyright,techcrunch2022}. These findings underpin ongoing copyright litigation against Stability AI and related entities~\citep{andersen2023complaint,getty2023uk}. More critically, independent researchers and watchdog organizations reported the presence of harmful and illegal material, including suspected child sexual abuse material (CSAM), within portions of LAION-5B~\citep{stanford2023laion,404media2023}. In response, LAION removed known CSAM hashes, introduced stricter filtering, and released revised dataset versions with enhanced safety measures~\citep{laion2023statement}. However, these interventions occurred after earlier diffusion models had already been trained and cannot retroactively remove learned representations embedded in model parameters.

This dataset lineage motivates the need for systematic auditing of stylistic behavior in generative models. Training on large-scale, weakly curated datasets such as LAION exposes diffusion models to dense concentrations of artworks and recurring stylistic patterns, which prior work suggests are internalized as distributed statistical regularities rather than explicit memorization~\citep{rombach2022high,carlini2023extracting}. These \emph{stored stylistic traces} may reappear during generation even when stylistic cues are not explicitly requested, leading to the silent appearance of recognizable stylistic features. Because such influence is neither directly observable in model parameters nor reliably captured by prompt-based qualitative analysis, existing evaluation approaches remain insufficient for disentangling explicit attribution, interaction, and unprompted stylistic re-emergence. To address this gap, \textit{ArtArena} formalizes stylistic influence as a measurable construct, converting it into an empirical signal through structured comparisons. By doing so, \textit{ArtArena} enables systematic auditing of stylistic leakage and provides insight into how large-scale training data shapes generative behavior, with applicability beyond text-to-image models to other generative modalities.

\subsection{Motif Extraction}
\begin{figure*}[t]
\centering
\begin{tcolorbox}[
  enhanced,
  unbreakable,
  colback=blue!5!white,
  colframe=blue!75!black,
  title= Museum \& Institutional Sources for Motifs Extraction,
]
\scriptsize

{\bfseries Museum \& Institutional Sources for Motifs Extraction}\\
{\itshape Best for authoritative descriptions, curatorial language, and visual motifs}

\begin{enumerate}
  \item \textbf{The Metropolitan Museum of Art (The Met) — Heilbrunn Timeline of Art History}\\
  \url{https://www.metmuseum.org}\\
  \textbf{Strengths:} clear visual descriptions; motif‑level analysis of gesture, composition, light, and material; especially strong for Caravaggio, Raphael, Warhol, and Monet.

  \item \textbf{Van Gogh Museum (Amsterdam)}\\
  \url{https://www.vangoghmuseum.nl}\\
  \textbf{Strengths:} exceptionally detailed analysis of brushwork, colour, and rhythm; explicit ground, sky, and vegetation motifs; ideal for prompt‑oriented extraction.

  \item \textbf{National Gallery (London) \& National Gallery of Art (Washington)}\\
  \url{https://www.nationalgallery.org.uk}\quad|\quad\url{https://www.nga.gov}\\
  \textbf{Strengths:} balanced formal and emotional descriptions; strong for portraits, interiors, and compositional motifs.

  \item \textbf{Musée d’Orsay (Paris)}\\
  \url{https://www.musee-orsay.fr}\\
  \textbf{Strengths:} Impressionist \& Post‑Impressionist motif language; atmosphere, light, and seasonal landscape descriptions.
\end{enumerate}

\medskip
\hrule height 0.4pt \vspace{0.6em}

{\bfseries High‑Quality Secondary Art Databases}\\
{\itshape Best for concise motif phrasing and comparative analysis}

\begin{enumerate}[start=5]
  \item \textbf{WikiArt}\\
  \url{https://www.wikiart.org}\\
  \textbf{Strengths:} aggregated descriptions across styles; quick identification of recurring motifs; good cross‑artist consistency.

  \item \textbf{Artchive}\\
  \url{http://www.artchive.com}\\
  \textbf{Strengths:} short, prompt‑friendly summaries; emphasis on composition and movement.

  \item \textbf{WahooArt}\\
  \url{https://www.wahooart.com}\\
  \textbf{Strengths:} plain‑language visual descriptions; helpful for classical and Renaissance works.
\end{enumerate}

\medskip
\hrule height 0.4pt \vspace{0.6em}

{\bfseries Scholarly / Curatorial Commentary}\\
{\itshape Best for deeper motif interpretation}

\begin{enumerate}[start=8]
  \item \textbf{Tate (UK)}\\
  \url{https://www.tate.org.uk}\\
  \textbf{Strengths:} excellent for modern and contemporary art; clear explanations of abstraction, repetition, and symbolism.

  \item \textbf{MoMA (Museum of Modern Art)}\\
  \url{https://www.moma.org}\\
  \textbf{Strengths:} essential for Pollock and Warhol; focus on process, material, and conceptual motifs.

  \item \textbf{Kröller‑Müller Museum}\\
  \url{https://krollermuller.nl}\\
  \textbf{Strengths:} landscape‑focused Van Gogh analysis; strong treatment of ground, vegetation, and sky motifs.
\end{enumerate}

\medskip
\hrule height 0.4pt \vspace{0.6em}

{\bfseries Art History \& Analysis Sites}\\
{\itshape Best for motif phrasing usable directly in prompts}

\begin{enumerate}[start=11]
  \item \textbf{The Collector}\\
  \url{https://www.thecollector.com}\\
  \textbf{Strengths:} accessible yet rigorous analysis; frequent discussion of symbolic and compositional motifs.

  \item \textbf{ArtUK}\\
  \url{https://artuk.org}\\
  \textbf{Strengths:} British collections with strong descriptive metadata.

  \item \textbf{Google Arts \& Culture}\\
  \url{https://artsandculture.google.com}\\
  \textbf{Strengths:} high‑resolution imagery with curatorial summaries; useful for spatial organisation and colour‑field understanding.
\end{enumerate}
\end{tcolorbox}
\vspace{-2pt}
\caption{Sources used to extract motif names and descriptions as commonly referenced by curators, art historians, and authoritative museum databases.}
\label{fig:art_sources}
\end{figure*}
\begin{figure*}[t]
\centering
\small
\begin{tcolorbox}[
  enhanced jigsaw,
  unbreakable,
  colback=blue!5!white,
  colframe=blue!75!black,
  title=Motifs Blending Prompt (Challenger)
]
\medskip
\textbf{Input:}
You need to generate prompts by blending the motifs from \{content\}.
Motifs represent \textbf{CONTENT ONLY} (objects, structures, symbols, spatial elements).
They do \textbf{NOT} represent style, artist identity, mood, or medium.

\medskip
\textbf{Task:}
\begin{itemize}
  \item Generate \textbf{ALL} non-empty combinations of the given motifs. If there are $N$ motifs, generate exactly $(2^{N}-1)$ combinations.
  \item Each combination must be used to construct \textbf{ONE} prompt.
\end{itemize}

\medskip
\textbf{Style Influence Constraints:}
\begin{itemize}
  \item The prompt MUST describe ONLY the motifs as visible scene content.
  \item The prompt MUST be compatible with later injection of external style or artist tokens.
  \item The prompt must be content-complete but stylistically neutral.
\end{itemize}

\medskip
\textbf{Prompt Construction Rules:}
For each motif combination:
\begin{itemize}
  \item Describe a single coherent scene that includes all motifs.
  \item Use literal, concrete, observational language.
  \item No references to famous artists, art movements, or mediums.
  \item No mood words (e.g., dramatic, surreal, expressive).
  \item No composition embellishment beyond spatial relations strictly required to place motifs.
  \item Do NOT infer symbolism beyond what is visually explicit.
  \item Do not exceed 70 tokens per prompt.
\end{itemize}

\medskip
\textbf{Output Format:}
Return STRICT JSON ONLY with the following structure.

\begin{verbatim}
{
  "num_motifs": N,
  "expected_combinations": 2^N - 1,
  "items": [
    {
      "combo_id": integer,
      "motifs": [exact motif strings],
      "content_prompt": "A single neutral content-only scene description suitable for 
      style leakage testing.",
      "style_injection_slot": "{{STYLE_OR_ARTIST_TO_BE_INJECTED_LATER}}"
    }
  ]
}
\end{verbatim}

\textbf{Ordering Constraints:}
\begin{itemize}
  \item Sort combinations by increasing number of motifs (1 → N).
  \item Preserve the original motif order inside each combination.
  \item Do not repeat or merge combinations.
\end{itemize}

\begin{verbatim}
Generate the output now
\end{verbatim}

\end{tcolorbox}
\vspace{-2pt}
\caption{\textbf{Motif-based challenger prompt construction}: We enumerate motif combinations and blend them into coherent, style-neutral scene descriptions used as challenger prompts in Motif Duels.}
\label{fig:blending_prompt}
\end{figure*}

\label{sec: motif_extraction}

This section describes the process used to extract motifs for running \textbf{Motif Duels}, an essential evaluation method in \textit{ArtArena}. For every artwork in the \texttt{FitSet}, we generate a \textit{content-only} list of motifs using structured prompts (Figure \ref{fig:motif_extraction}) given to GPT-4o. These prompts instruct the model to identify only the concrete, visible elements in the image, such as objects, structures, symbols, and spatial components, while avoiding any references to style, artist identity, mood, medium, or lighting. The result is a clean, style-neutral motif list that represents the artwork’s content in a consistent and comparable way.

\begin{table*}
\centering
\caption{\textbf{Motifs extraction and blending}. Extracted motifs and their blending to construct the prompt for the artwork. The list of extracted motifs and prompts for different artworks is provided in Table \ref{tab:motifs}.}
\label{tab:motif_combinations}
\renewcommand{\arraystretch}{1.15}
\setlength{\tabcolsep}{4pt}
\small  
\begin{tabular}{@{}p{3.5cm} p{4.2cm} p{5.8cm}@{}}
\toprule
\textbf{Artwork} & \textbf{Motifs} & \textbf{Prompt} \\
\midrule
\multirow{5}{=}{\textit{The Starry Night} by Vincent Van Gogh}
& Night sky, Village & A dark sky with many stars above a small cluster of buildings at the base of the hills. \\
& Cypress tree, Swirling cloud band & A tall tree mass in the foreground with a broad curved band of clouds crossing behind it. \\
& Church steeple, Night sky, Village & A pointed steeple rising above clustered houses beneath a star-filled sky. \\
& Swirling cloud band, Night sky, Cypress tree & A curved band of clouds moving across a starry sky with a tall tree mass in the foreground. \\
& Village, Cypress tree, Church steeple, Night sky & A group of small buildings and a narrow steeple near a tall foreground tree under a dark starry sky. \\
\midrule
\multirow{5}{=}{\textit{The Scream} by Edvard Munch}
& Figure on bridge, Waterbody & A central figure standing on a bridge with a stretch of water visible beyond it. \\
& Bridge railing, Two distant figures, Horizon and landform & A walkway with railings receding into depth and two small figures near a low horizon with landforms. \\
& Figure on bridge, Bridge railing, Horizon and landform & A main figure on a bridge with horizontal planks and a distant low horizon of landmasses. \\
& Waterbody, Bridge railing & A wide body of water seen past the receding railings of a wooden bridge. \\
& Figure on bridge, Two distant figures, Waterbody, Horizon and landform & A central figure with two smaller figures farther back on a bridge overlooking water and a low horizon of landforms. \\
\bottomrule
\end{tabular}
\end{table*}






\begin{table*}[t]
\centering
\caption{Examples of the extracted motif for different artworks.}
\label{tab:motifs}
\small
\begin{tabular}{@{}p{6.0cm} p{3.0cm} p{3.2cm}@{}}
\toprule
\multicolumn{1}{c}{\textbf{Artwork Name}} &
\multicolumn{1}{c}{\textbf{Artwork}} &
\multicolumn{1}{c}{\textbf{Motifs}} \\
\midrule

\multirow{5}{=}{\textit{The Starry Night} by Vincent Van Gogh}
  & \artimg{5}{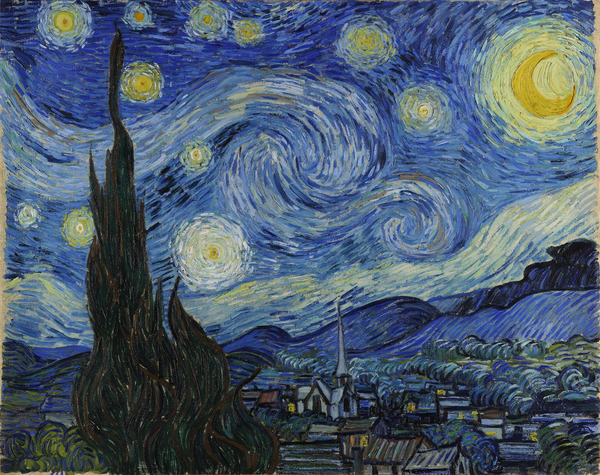} & Night sky \\
  &  & Swirling cloud band \\
  &  & Village \\
  &  & Church steeple \\
  &  & Cypress tree \\
\midrule

\multirow{3}{=}{\textit{Pool with Waterlilies} by Claude Monet}
  & \artimg{3}{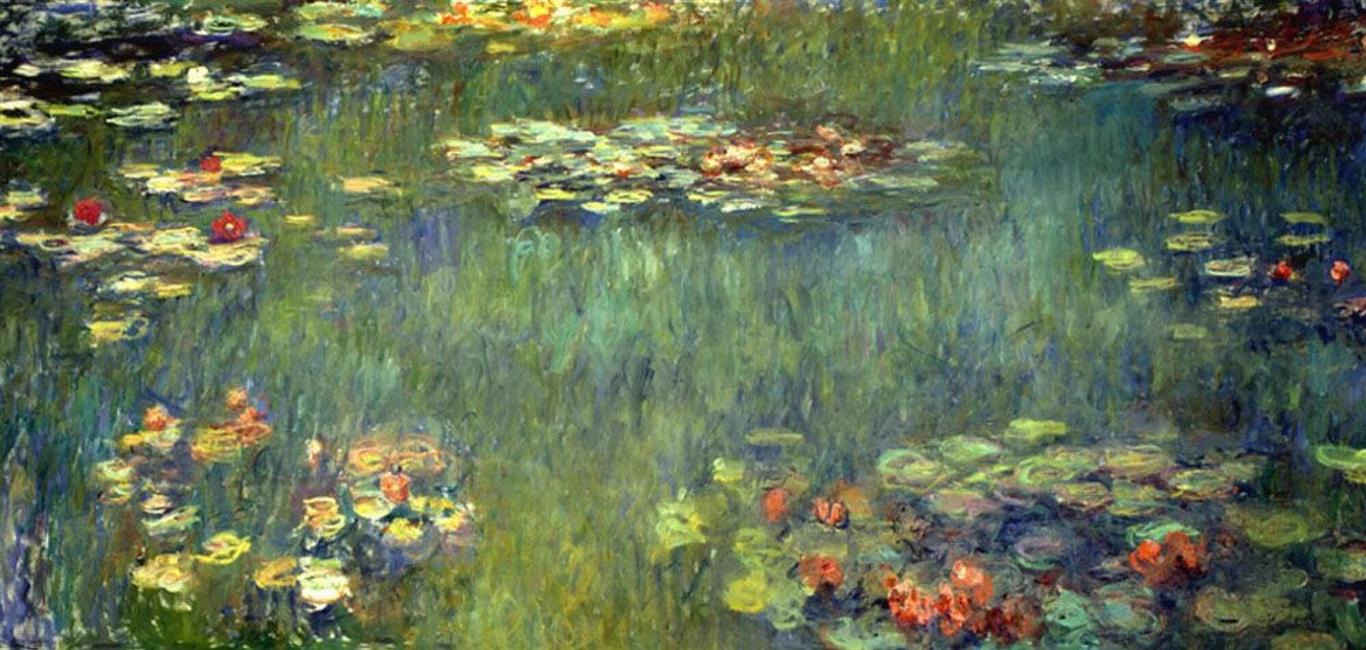} & Water lilies \\
  &  & Water reflections \\
  &  & Natural vegetation \\
\midrule

\multirow{5}{=}{\textit{The Scream} by Edvard Munch}
  & \artimg{5}{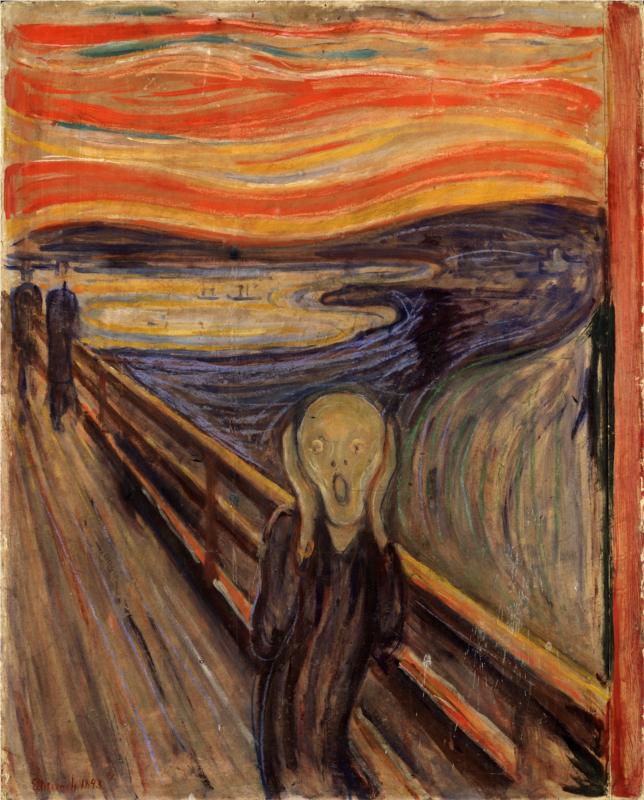} & Figure on bridge \\
  &  & Bridge railing \\
  &  & Two distant figures \\
  &  & Waterbody \\
  &  & Horizon and landform \\
\midrule

\multirow{4}{=}{\textit{The Fall on the Road to Calvary} by Raphael}
  & \artimg{4}{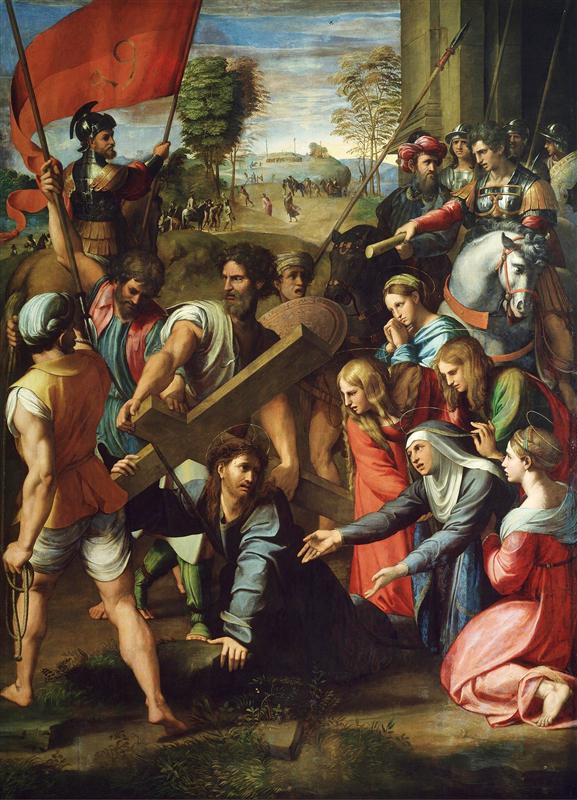} & The Cross \\
  &  & The Fallen Christ \\
  &  & The Mourning Women \\
  &  & Roman Soldiers \\
\midrule

\multirow{4}{=}{\textit{The Ascension Of Christ} by Rembrandt}
  & \artimg{4}{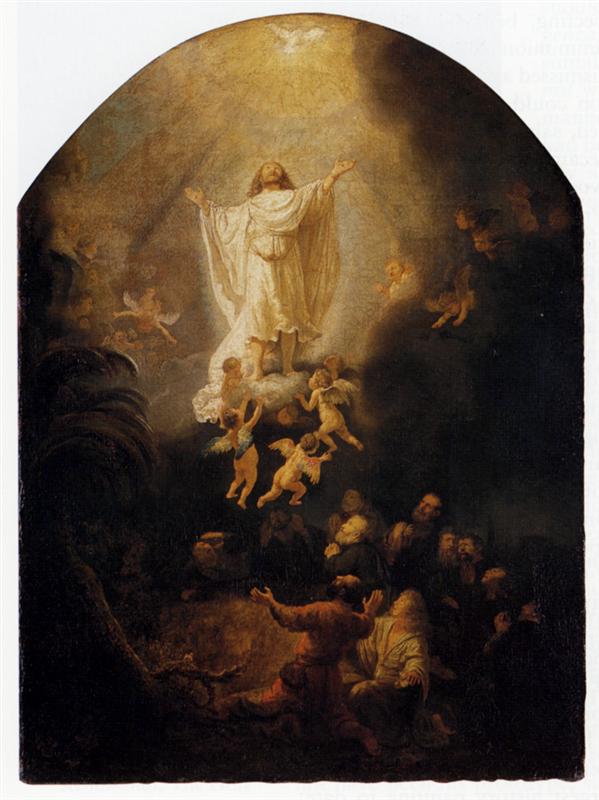} & Central figure of Christ \\
  &  & Cherubic angels \\
  &  & Group of observers \\
  &  & Vegetative motifs \\
\bottomrule
\end{tabular}
\end{table*}

Including clear \textbf{Style Influence Constraints} and \textbf{Prompt Construction Rules} (refer Figure \ref{fig:blending_prompt}) in this process is important for separating content from style. Without these rules, extra clues like artistic terms, emotional language, or medium descriptors could unintentionally influence the behavior of image generation models, making it harder to tell whether the model understands the actual content of the scene. Our framework prevents this in three ways. First, it enforces a strict content-only vocabulary, ensuring that motifs are free of stylistic hints. Second, for a motif set of size $N$, the method generates prompts for all non-empty combinations of motifs, producing exactly $(2^{N} - 1)$ prompts. This full combinatorial coverage helps reveal how models handle different content combinations without style interference. Third, the prompt rules require short, neutral descriptions that use only minimal spatial details, keeping the phrasing simple and style-free while also allowing later style injection when needed.

Together, this motif extraction process and the controlled prompting rules create a reliable testing setup for Motif Duels. Each prompt acts as a focused test of whether a model can represent and combine content correctly, while style remains fully separated. This enables precise measurement of style influence, consistent comparison across artworks, and a stable foundation for later experiments where style can be introduced independently of content.

The following detailed sources (Figure \ref{fig:art_sources}) were used to retrieve the necessary motifs. In the above mentioned prompt, the agent is not just provided with the artwork name and the artist name but also with the artwork image for better capturing of motifs.

\subsection{Artworks and their extracted motifs} 
\label{subsec:Artwork_Extracted_Motifs}

In this section, we provide examples of the extracted motif corresponding to two of the most famous artworks namely "The Starry Night by Vincent Van Gogh" and "The Scream by Edvard Munch". The Table \ref{tab:motifs} describes the types of motifs extracted and the corresponding motifs descriptions for each of the motif. The Table \ref{tab:motif_combinations} represent the crafted prompts corresponding to various combinations using the extracted motifs. The above mentioned Tables help us better understand how different motifs of the same artworks can be carefully combined, such that they semantically are correct.




In the above visual results of the various motifs, their combinations illustrate that simply adding more motifs does not monotonically strengthen an artwork’s stylistic influence. Instead, the influence saturates, plateaus, or can even weaken depending on how well the model has internalized that artwork’s stylistic schema. This reinforces why the proposed Motif Duel setup is necessary: pairwise battleground evaluations reveal whether an artwork’s style genuinely exerts a deep, distributed pull on the model’s parameter space, or whether its influence collapses once motifs are hybridized with competing styles. By forcing artworks to confront one another as both challenger and defender, the duel framework exposes these asymmetric and nonlinear behaviors—showing which styles survive aggressive recombination and which dissipate under motif blending. Consequently, motif duels provide a more faithful measure of leakage potential than raw motif similarity alone, capturing how style representations behave when activated indirectly, competitively, and compositionally.

\subsection{Proximity metrics}
\label{sec:PM}
We measure proximity with three complementary metrics: CLIP cosine similarity \citep{radford2021learning,hessel2021clipscore}, Learned Perceptual Image Patch Similarity (LPIPS) \citep{zhang2018lpips}, and Contrastive Style Descriptors (CSD) \citep{somepalli2024measuring}. The presented set of proximity targets to capture distinct aspect of style. CLIP similarity captures high‑level semantics and composition. It supports content level recall and text–image alignment \citep{radford2021learning,hessel2021clipscore}. It has also been used for data attribution through embedding‑space distribution analysis \citep{joshi2026dota}. It is used widely but is known to have limitations such as sensitivity to prompt wording, dataset biases, and weaker performance on fine‑grained differences \citep{shao2023investigating}. LPIPS measures perceptual closeness via deep feature distances. It reflects aesthetics properties such as structure and texture  \citep{zhang2018lpips}. LPIPS is often included in training objectives for diffusion and restoration to improve realism and to probe the perception–distortion tradeoff \citep{ho2022cascaded,saharia2022photorealistic}. Recent studies have shown that it can be vulnerable to adversarial perturbations and may misalign with human judgments under distribution shift \citep{kettunen2019lpips,ghazanfari2023r}. CSD encodes content‑invariant style cues such as color palettes, texture statistics, and stroke patterns \citep{somepalli2024measuring}. It serves as a style fidelity metric for image‑driven style transfer \citep{xing2024csgo} and as a pre‑trained style embedding for cosine‑similarity scoring against reference image to maintain style consistency \citep{mou2025dreamo}. The quality of the proximity score, depends on how diverse and well-covered the curated training dataset is, which was built by finding style tags in high‑aesthetic LAION captions, removing overly common tags, dropping broken URLs, and deduplicating near‑duplicate images while merging their tags.

\subsection{Robustness of the Influence Ledger}
\label{sec:ledger_robustness}

We test the robustness of the rankings provided by the Influence Ledger. This is important because an artist relies on these rankings to shortlist the canonical artworks treated as leakage candidates (Section~\ref{sec:cs_phases}). If the rankings changed with the random seeds of generation or with the language model that extracts the motifs, the shortlist would reflect these choices rather than the model under audit. For each run of ArtArena, we therefore assess two forms of reliability:
\begin{itemize}[leftmargin=*, itemsep=2pt]
    \item \textbf{Influence Ledger reliability}: We repeat the run under identical settings, changing only the random seeds of generation.
    \item \textbf{Motif Extraction reliability}: We repeat the run with motifs extracted by a different language model in place of GPT-4o.
\end{itemize}

\paragraph{Experimental setup.}
For both assessments, we use the same five WikiArt artworks: \emph{Impression, Sunrise} by Claude Monet, \emph{The Starry Night} by Vincent van Gogh, \emph{The Bellelli Family} by Edgar Degas, \emph{Mona Lisa} by Leonardo da Vinci and \emph{The Scream} by Edvard Munch. We run ArtArena over these artworks on all three models (SD v1.5, SDXL, SANA-1.5) under all three proximity metrics (Semantics, Aesthetics, Fidelity), yielding nine Influence Ledgers. Motif Duels use $R = 5$ rounds per match and $K = 5$ images per round. For Motif Extraction reliability, we use Qwen2.5-VL-32B-Instruct \citep{bai2025qwen25vl} with the same extraction prompts, sources and constraints as GPT-4o (Section~\ref{sec: motif_extraction}), and rerun the Motif Duels. Each assessment produces a second Influence Ledger for each of the nine runs, which we compare against the corresponding ledger obtained using GPT-4o.

\paragraph{Robustness measure.}
We compare the two rankings of each pair of ledgers through the Spearman rank correlation coefficient
{\small
\begin{equation}
\label{eq:spearman}
r_s = 1 - \frac{6\sum_{w}\big(\pi(w) - \pi'(w)\big)^2}{n\,(n^2-1)},
\end{equation}
}
where $\pi(w)$ and $\pi'(w)$ denote the ranks of artwork $w$ in the original and the second ledger, respectively. Here, $n = 5$ is the number of artworks. A value of $r_s = 1$ indicates that both ledgers rank the artworks identically. For Influence Ledger reliability, a high $r_s$ indicates that the ranking is insensitive to the randomness of generation. For Motif Extraction reliability, a high $r_s$ indicates that the ranking is insensitive to the language model that extracts the motifs. A robust Influence Ledger should therefore yield $r_s$ close to $1$ in both assessments.

\paragraph{Observations.}
\textbf{Influence Ledger reliability.}
Changing the random seeds preserves the ranking of artworks in most settings (\cref{tab:ledger_robustness}, top). Reordering remains limited in most settings and is more pronounced under Fidelity. This may stem from CSD, which encodes style cues such as color palettes, texture statistics and stroke patterns (Section~\ref{sec:PM}). Since such cues may vary across generated samples, this sensitivity may reflect a limitation of the proximity metric rather than of ArtArena. \textit{Thus, the shortlisted leakage candidates generally reflect the model under audit rather than the randomness of generation.}

\textbf{Motif Extraction reliability.}
Replacing GPT-4o with Qwen2.5-VL preserves the ranking of artworks in most settings (\cref{tab:ledger_robustness}, bottom). Under Semantics and Aesthetics, the rankings show limited reordering across all three models. Fidelity, in contrast, is more sensitive to the choice of extractor. CSD responds to style cues, so differences in how the two extractors phrase the motifs can shift these cues in the generated images. As ArtArena is not tied to any particular proximity metric, such cases can be addressed by adopting more stable style measures without modifying the protocol. \textit{Thus, under Semantics and Aesthetics, the shortlisted leakage candidates generally reflect the model rather than the choice of motif extractor.}

\begin{table}[h]
\centering
\small
\caption{\textbf{Robustness of the Influence Ledger.} Spearman rank correlation $r_s$ (\cref{eq:spearman}) between the original and the second ledger over five WikiArt artworks. \textit{Top:} Influence Ledger reliability, varying only the random seeds. \textit{Bottom:} Motif Extraction reliability, replacing GPT-4o with Qwen2.5-VL-32B-Instruct.}
\label{tab:ledger_robustness}
\begin{tabular}{@{}lcccc@{}}
\toprule
\textbf{Proximity} & \textbf{SD v1.5} & \textbf{SDXL} & \textbf{SANA-1.5} & \textbf{Mean} \\
\midrule
\multicolumn{5}{@{}l}{\textit{Influence Ledger reliability}} \\
\quad Semantics  & 0.90 & 0.60 & 0.90 & 0.80 \\
\quad Aesthetics & 0.80 & 1.00 & 1.00 & 0.93 \\
\quad Fidelity   & 0.20 & 0.70 & 1.00 & 0.63 \\
\midrule
\multicolumn{5}{@{}l}{\textit{Motif Extraction reliability}} \\
\quad Semantics  & 0.90 & 0.70 & 0.90 & 0.83 \\
\quad Aesthetics & 0.90 & 1.00 & 0.70 & 0.87 \\
\quad Fidelity   & 0.20 & 0.30 & 0.70 & 0.40 \\
\bottomrule
\end{tabular}
\end{table}

\subsection{Experiments}
\label{sec:experiments_details}

\subsubsection{Personalization Leakage Assessment}
\label{sec:pla_details}

\begin{figure*}[p]
  \centering
  \includegraphics[width=0.70\textwidth]{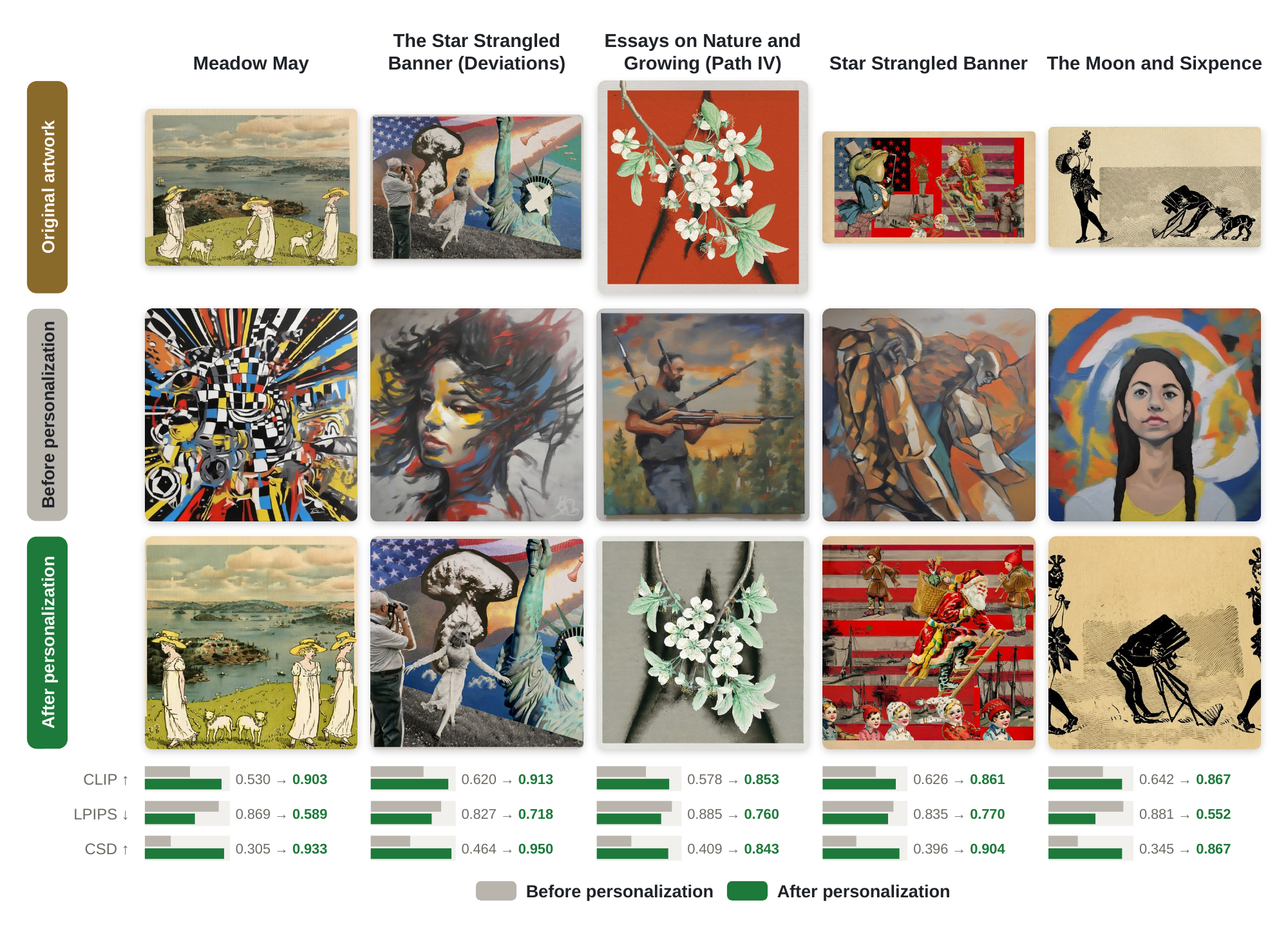}
  \caption{\textbf{Personalization recall of SDXL on the artworks of Trent Anthony Francis.} Top: the artist's original artworks. Middle: images generated by the base model from each artwork's personalization prompt, \texttt{painting in the style of <id>}. Bottom: images generated from the same prompt by the personalized model. Bars report the fitness (\cref{eq:fitness}) of each artwork before (gray) and after (green) personalization under Semantics (CLIP, $\uparrow$), Aesthetics (LPIPS, $\downarrow$), and Fidelity (CSD, $\uparrow$). Each score is averaged over $K=3$ generations, one of which is shown in the middle and bottom rows.}
  \label{fig:recall_SDXL_Trent}
  \vspace{10pt}
  \includegraphics[width=0.70\textwidth]{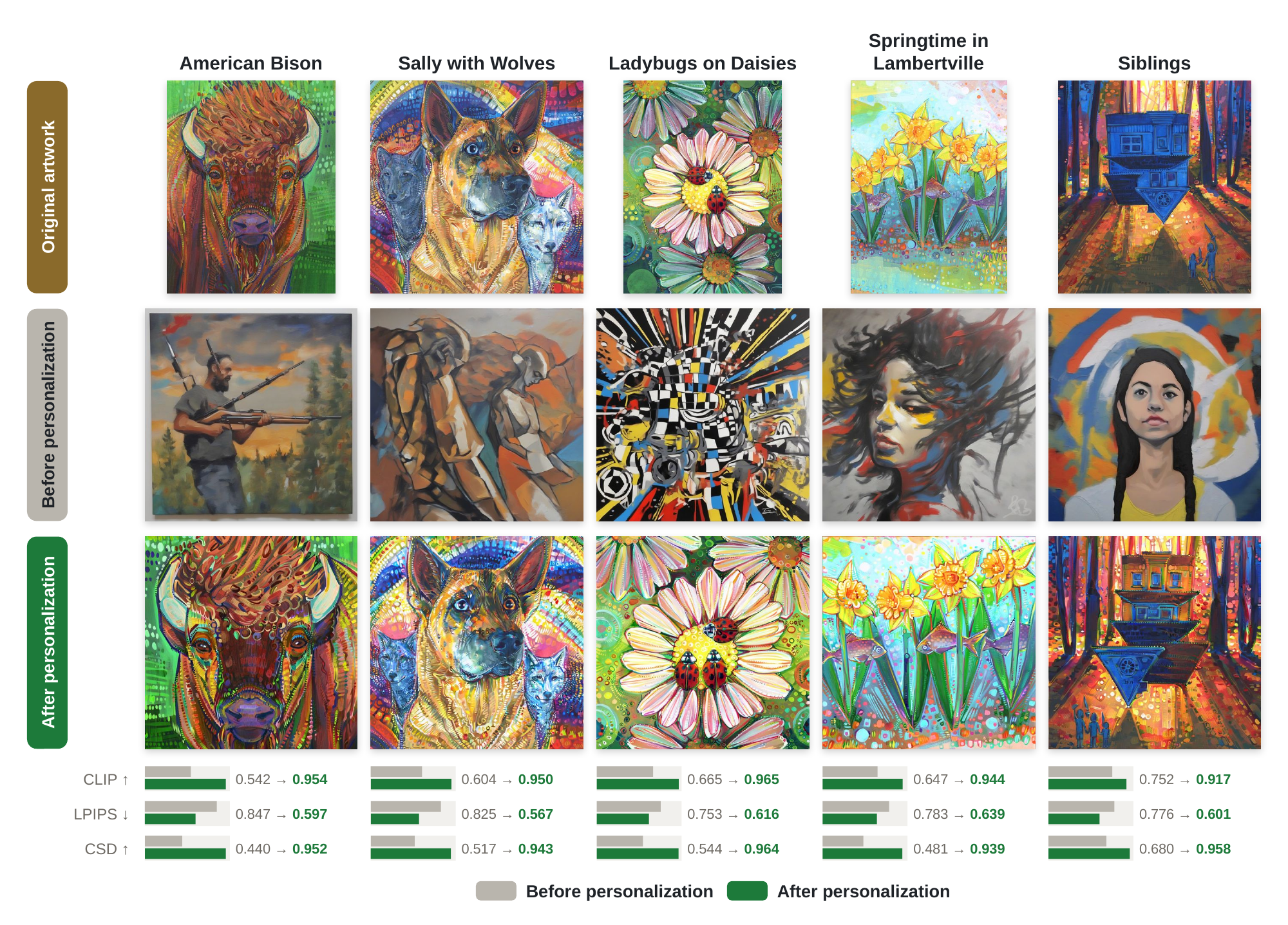}
  \caption{\textbf{Personalization recall of SDXL on the artworks of Gwenn Seemel.} Top: the artist's original artworks. Middle: images generated by the base model from each artwork's personalization prompt, \texttt{painting in the style of <id>}. Bottom: images generated from the same prompt by the personalized model. Bars report the fitness (\cref{eq:fitness}) of each artwork before (gray) and after (green) personalization under Semantics (CLIP, $\uparrow$), Aesthetics (LPIPS, $\downarrow$), and Fidelity (CSD, $\uparrow$). Each score is averaged over $K=3$ generations, one of which is shown in the middle and bottom rows.}
  \label{fig:recall_SDXL_Gwenn}
\end{figure*}

\begin{figure*}[p]
  \centering
  \includegraphics[width=0.70\textwidth]{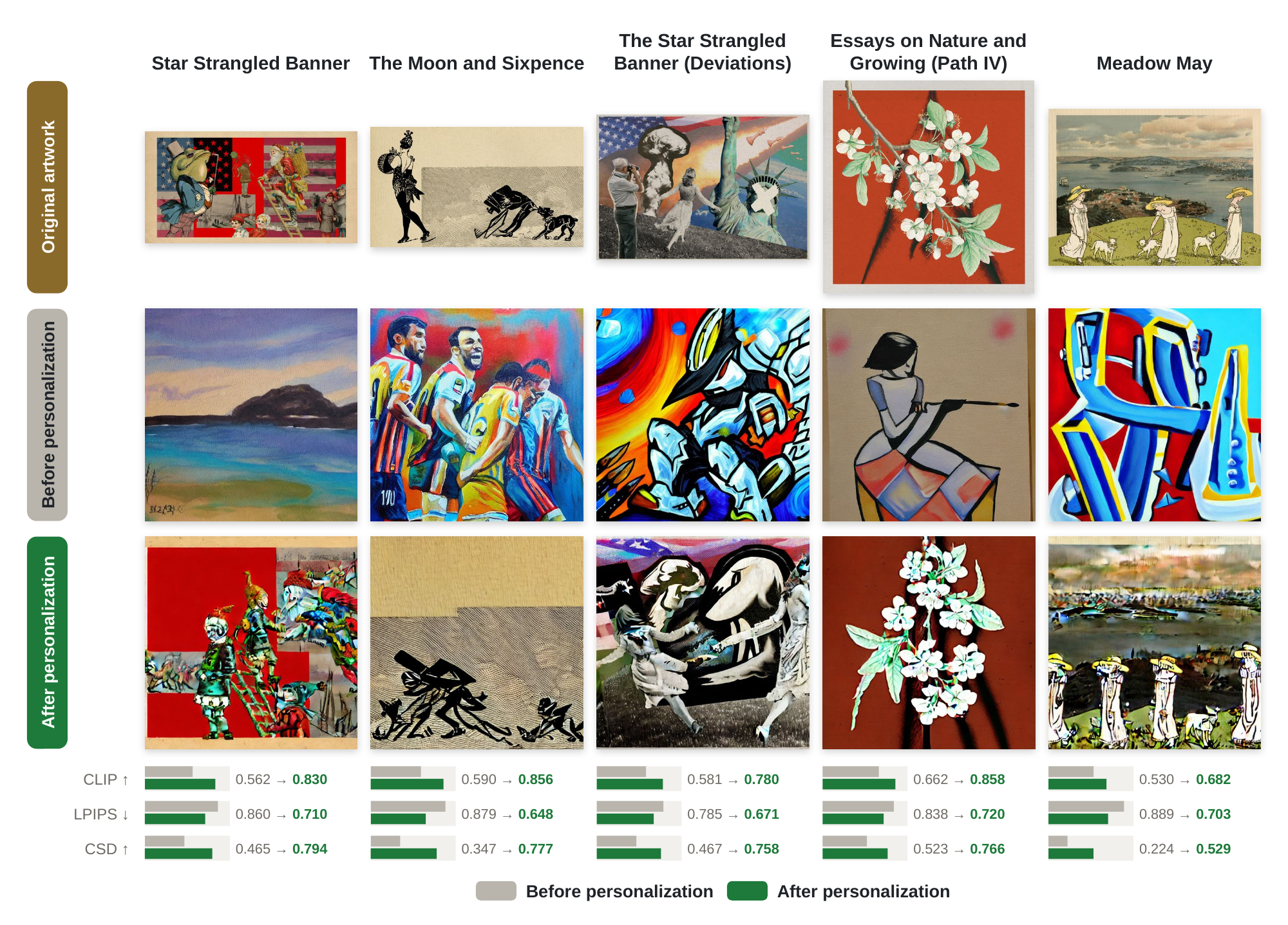}
  \caption{\textbf{Personalization recall of SD v1.5 on the artworks of Trent Anthony Francis.} Top: the artist's original artworks. Middle: images generated by the base model from each artwork's personalization prompt, \texttt{painting in the style of <id>}. Bottom: images generated from the same prompt by the personalized model. Bars report the fitness (\cref{eq:fitness}) of each artwork before (gray) and after (green) personalization under Semantics (CLIP, $\uparrow$), Aesthetics (LPIPS, $\downarrow$), and Fidelity (CSD, $\uparrow$). Each score is averaged over $K=3$ generations, one of which is shown in the middle and bottom rows.}
  \label{fig:recall_SD_Trent}
  \vspace{10pt}
  \includegraphics[width=0.70\textwidth]{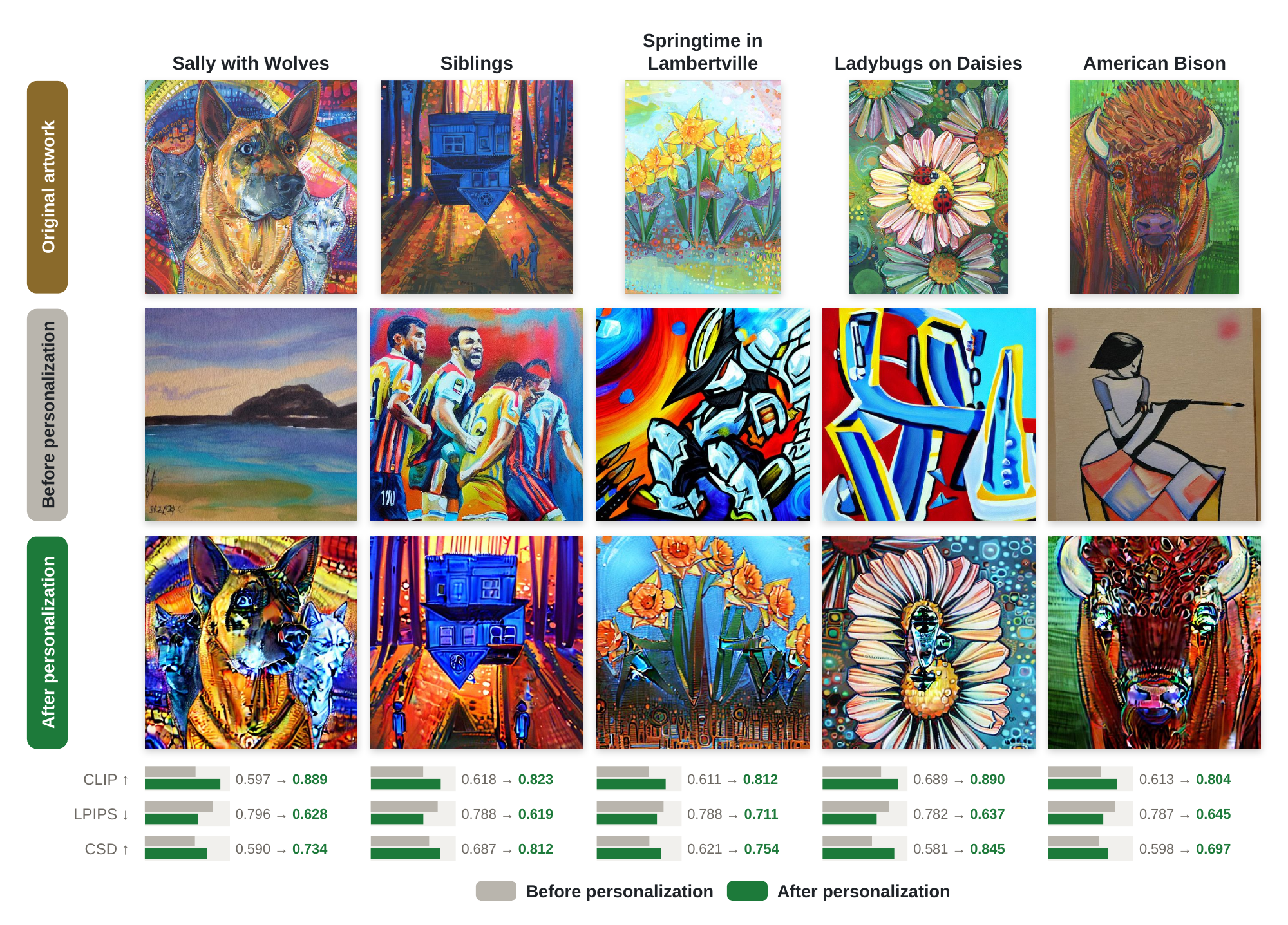}
  \caption{\textbf{Personalization recall of SD v1.5 on the artworks of Gwenn Seemel.} Top: the artist's original artworks. Middle: images generated by the base model from each artwork's personalization prompt, \texttt{painting in the style of <id>}. Bottom: images generated from the same prompt by the personalized model. Bars report the fitness (\cref{eq:fitness}) of each artwork before (gray) and after (green) personalization under Semantics (CLIP, $\uparrow$), Aesthetics (LPIPS, $\downarrow$), and Fidelity (CSD, $\uparrow$). Each score is averaged over $K=3$ generations, one of which is shown in the middle and bottom rows.}
  \label{fig:recall_SD_Gwenn}
\end{figure*}

\begin{figure*}[p]
  \centering
  \includegraphics[width=0.70\textwidth]{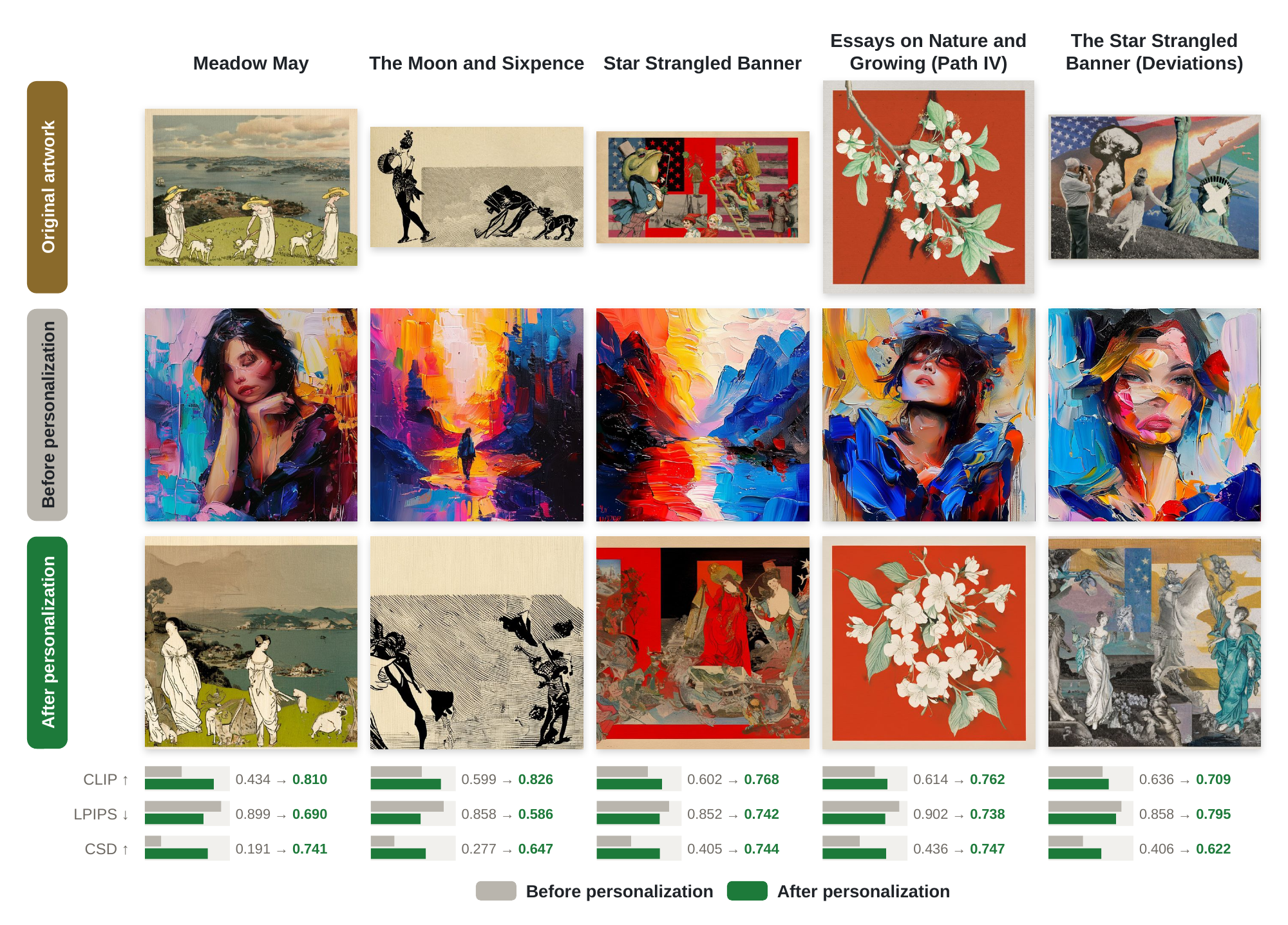}
  \caption{\textbf{Personalization recall of SANA-1.5 on the artworks of Trent Anthony Francis.} Top: the artist's original artworks. Middle: images generated by the base model from each artwork's personalization prompt, \texttt{painting in the style of <id>}. Bottom: images generated from the same prompt by the personalized model. Bars report the fitness (\cref{eq:fitness}) of each artwork before (gray) and after (green) personalization under Semantics (CLIP, $\uparrow$), Aesthetics (LPIPS, $\downarrow$), and Fidelity (CSD, $\uparrow$). Each score is averaged over $K=3$ generations, one of which is shown in the middle and bottom rows.}
  \label{fig:recall_SANA_Trent}
  \vspace{10pt}
  \includegraphics[width=0.70\textwidth]{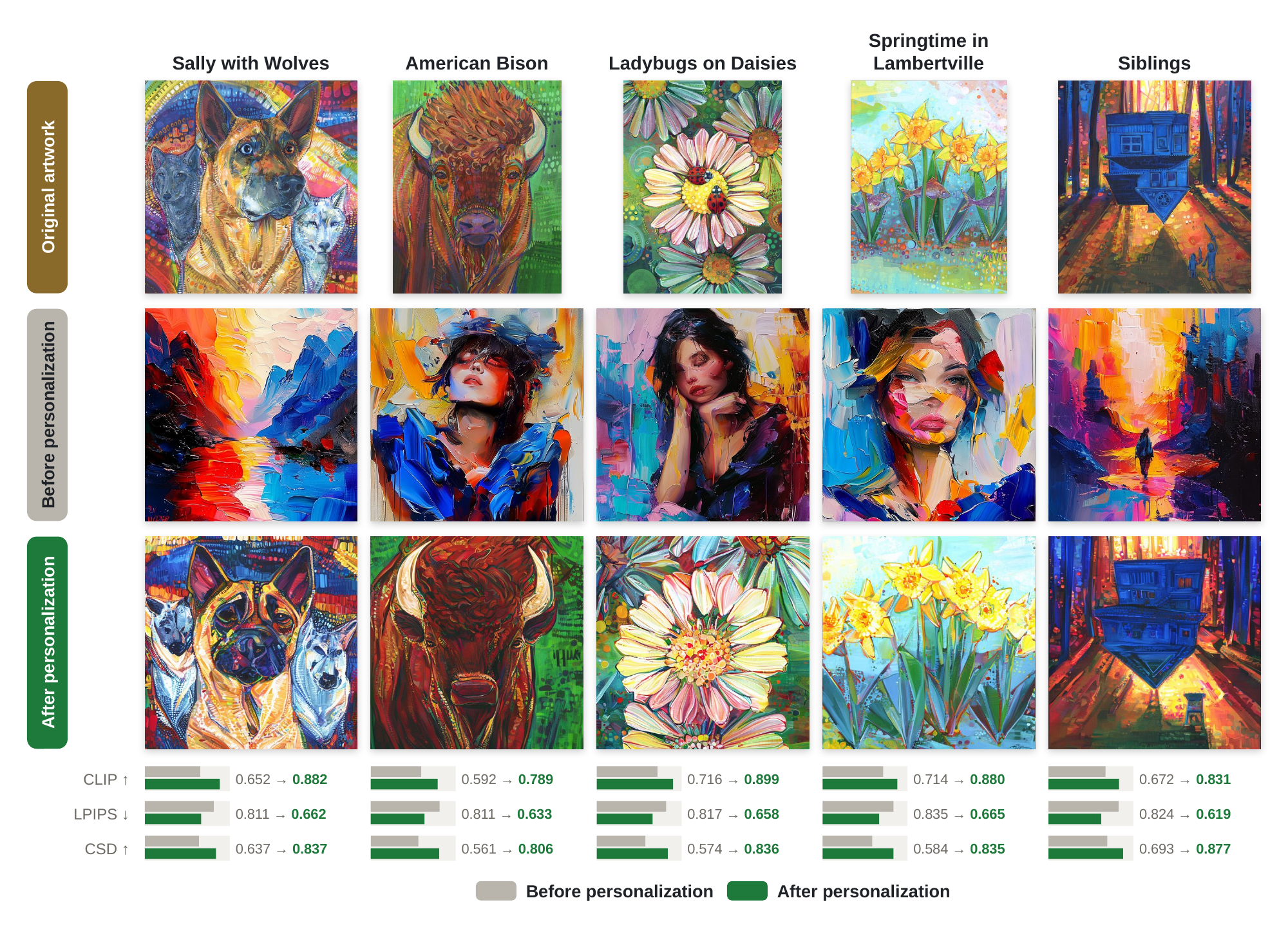}
  \caption{\textbf{Personalization recall of SANA-1.5 on the artworks of Gwenn Seemel.} Top: the artist's original artworks. Middle: images generated by the base model from each artwork's personalization prompt, \texttt{painting in the style of <id>}. Bottom: images generated from the same prompt by the personalized model. Bars report the fitness (\cref{eq:fitness}) of each artwork before (gray) and after (green) personalization under Semantics (CLIP, $\uparrow$), Aesthetics (LPIPS, $\downarrow$), and Fidelity (CSD, $\uparrow$). Each score is averaged over $K=3$ generations, one of which is shown in the middle and bottom rows.}
  \label{fig:recall_SANA_Gwenn}
\end{figure*}

\paragraph{Personalization recall.} Figs.~\ref{fig:recall_SDXL_Trent}--\ref{fig:recall_SANA_Gwenn} complement Block~A of the leakage cards (Section~\ref{sec:leakage_card}) by showing the generations of the base and personalized models for each artwork's personalization prompt (Section~\ref{sec:exp}). While the base model generates images unrelated to the artworks, the personalized model recalls their content and style, increasing the fitness of every artwork under all three proximity metrics.

\begin{figure*}[p]
  \centering
  \begin{minipage}[t]{0.48\textwidth}
    \centering
    \resizebox{0.85\linewidth}{!}{\includegraphics[]{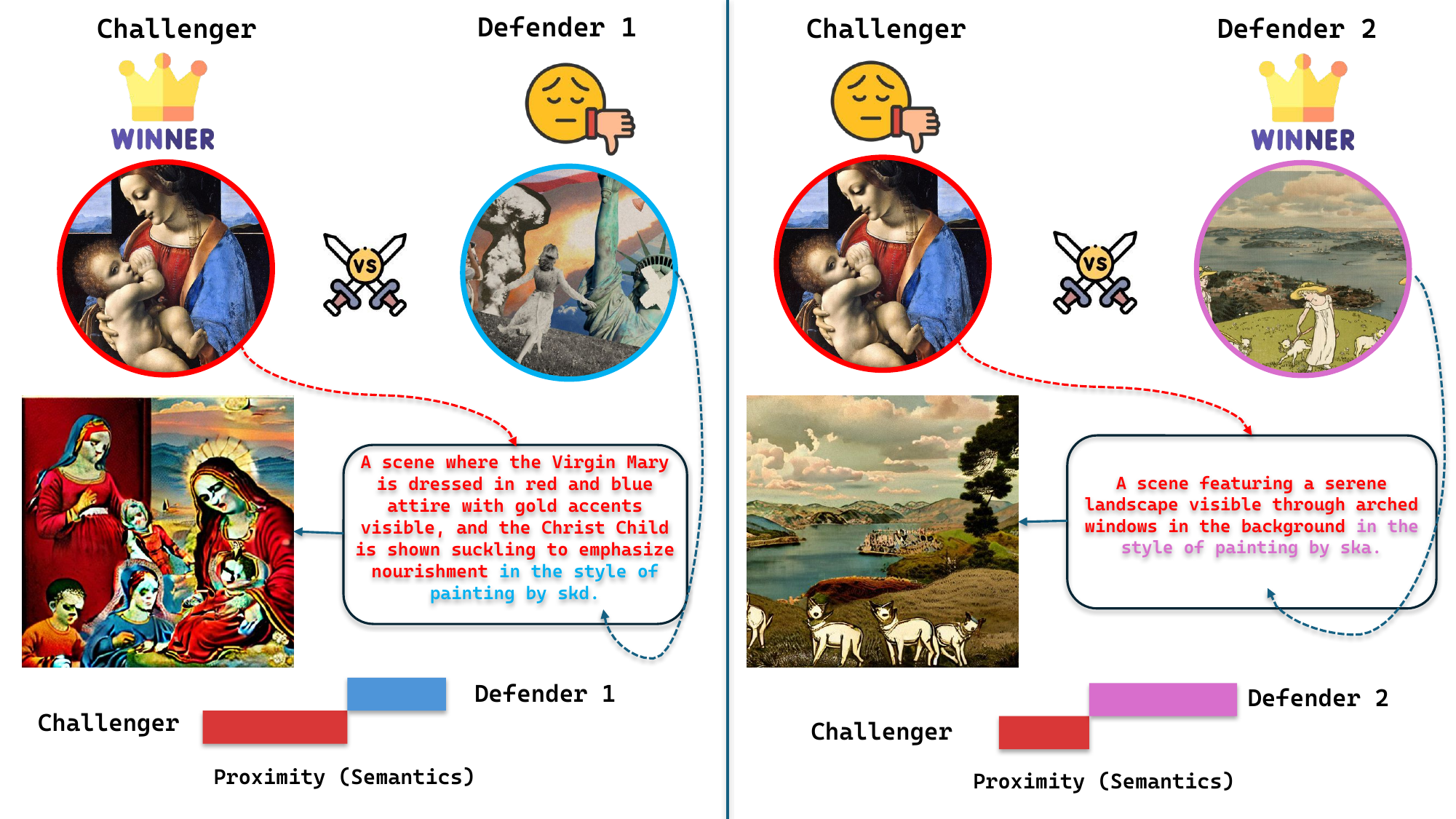}}
    \caption{Illustration of a \emph{Motif Duel} for SD v1.5 personalized on artworks by Trent Anthony Francis, evaluated using semantics-based proximity (CLIP). The results demonstrate how combining the motifs of the canonical challenger \textit{Madonna Litta (Madonna and the Child)} by Leonardo da Vinci with the artist's \textit{The Star Strangled Banner (Deviations)} (Defender 1, \texttt{skd}) and \textit{Meadow May} (Defender 2, \texttt{ska}) produces stylistically distinct outputs.}
    \label{fig:pduel_sd_sem_trent}
  \end{minipage}\hfill
  \begin{minipage}[t]{0.48\textwidth}
    \centering
    \resizebox{0.85\linewidth}{!}{\includegraphics[]{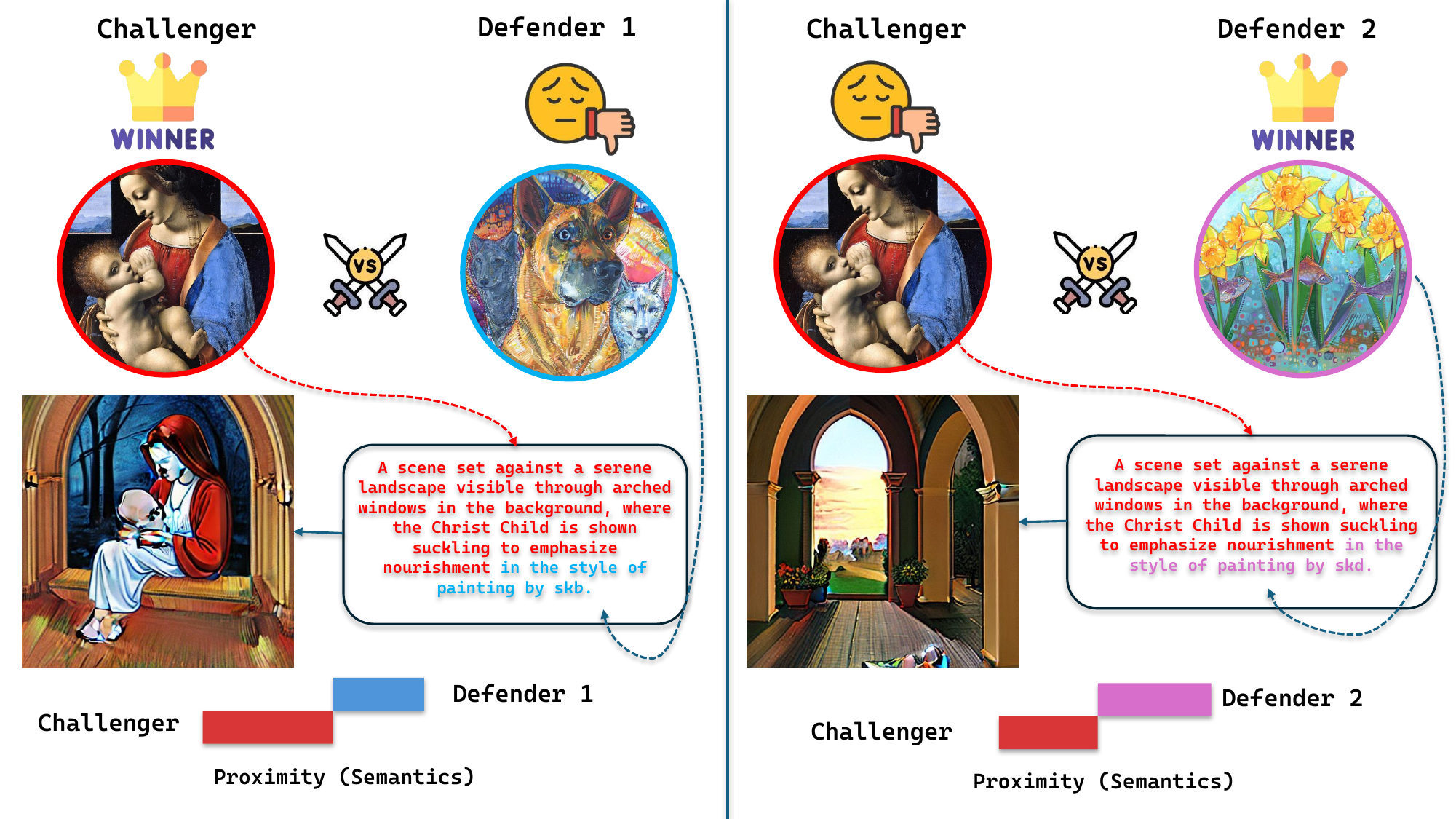}}
    \caption{Illustration of a \emph{Motif Duel} for SD v1.5 personalized on artworks by Gwenn Seemel, evaluated using semantics-based proximity (CLIP). The results demonstrate how combining the motifs of the canonical challenger \textit{Madonna Litta (Madonna and the Child)} by Leonardo da Vinci with the artist's \textit{Sally with Wolves} (Defender 1, \texttt{skb}) and \textit{Springtime in Lambertville} (Defender 2, \texttt{skd}) produces stylistically distinct outputs.}
    \label{fig:pduel_sd_sem_gwenn}
  \end{minipage}
  \par\vspace{1pt}
  \begin{minipage}[t]{0.48\textwidth}
    \centering
    \resizebox{0.85\linewidth}{!}{\includegraphics[]{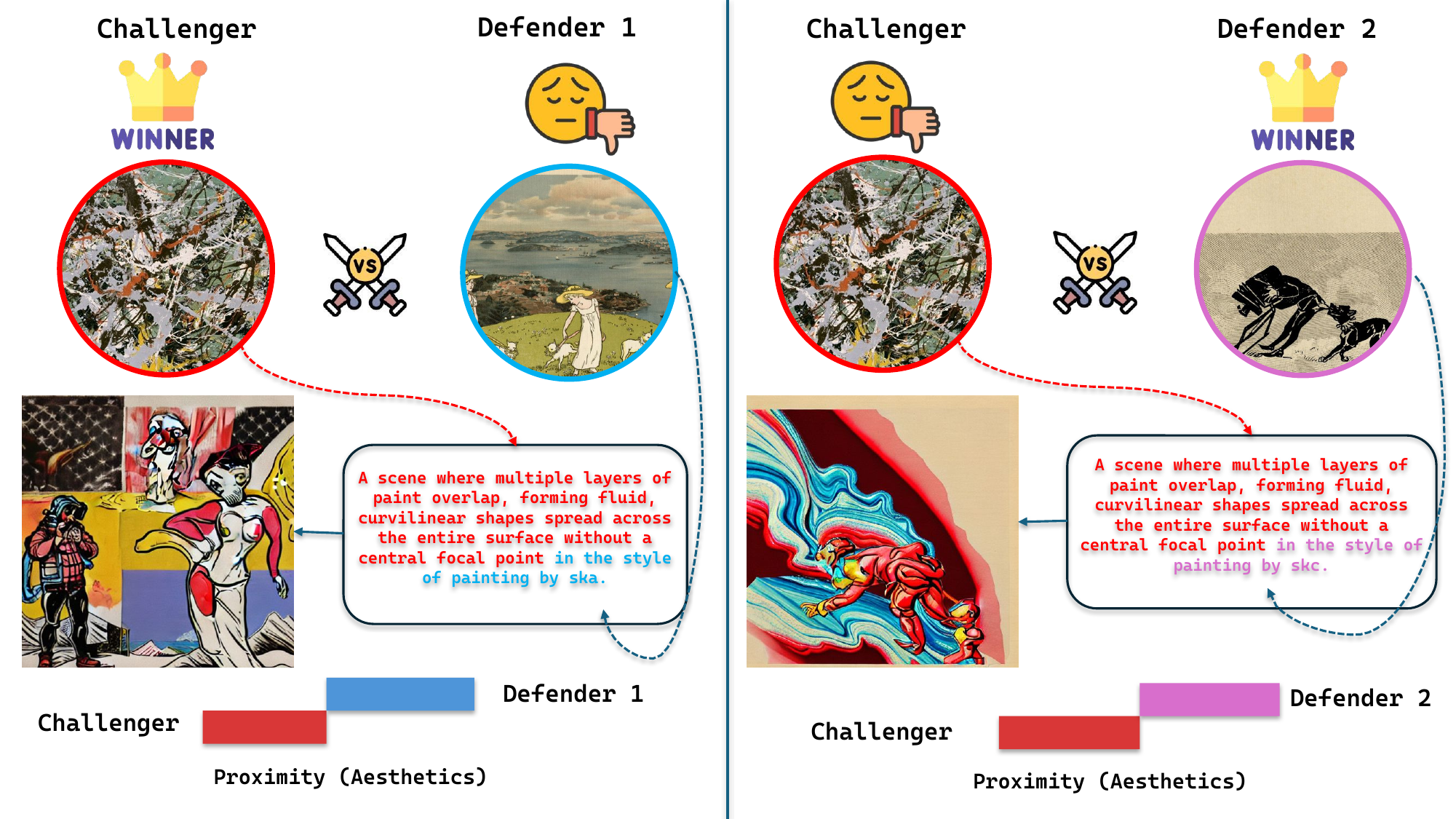}}
    \caption{Illustration of a \emph{Motif Duel} for SD v1.5 personalized on artworks by Trent Anthony Francis, evaluated using aesthetics-based proximity (LPIPS; lower scores indicate closer proximity). The results demonstrate how combining the motifs of the canonical challenger \textit{Untitled (Green Silver)} by Jackson Pollock with the artist's \textit{Meadow May} (Defender 1, \texttt{ska}) and \textit{The Moon and Sixpence} (Defender 2, \texttt{skc}) produces stylistically distinct outputs.}
    \label{fig:pduel_sd_aes_trent}
  \end{minipage}\hfill
  \begin{minipage}[t]{0.48\textwidth}
    \centering
    \resizebox{0.85\linewidth}{!}{\includegraphics[]{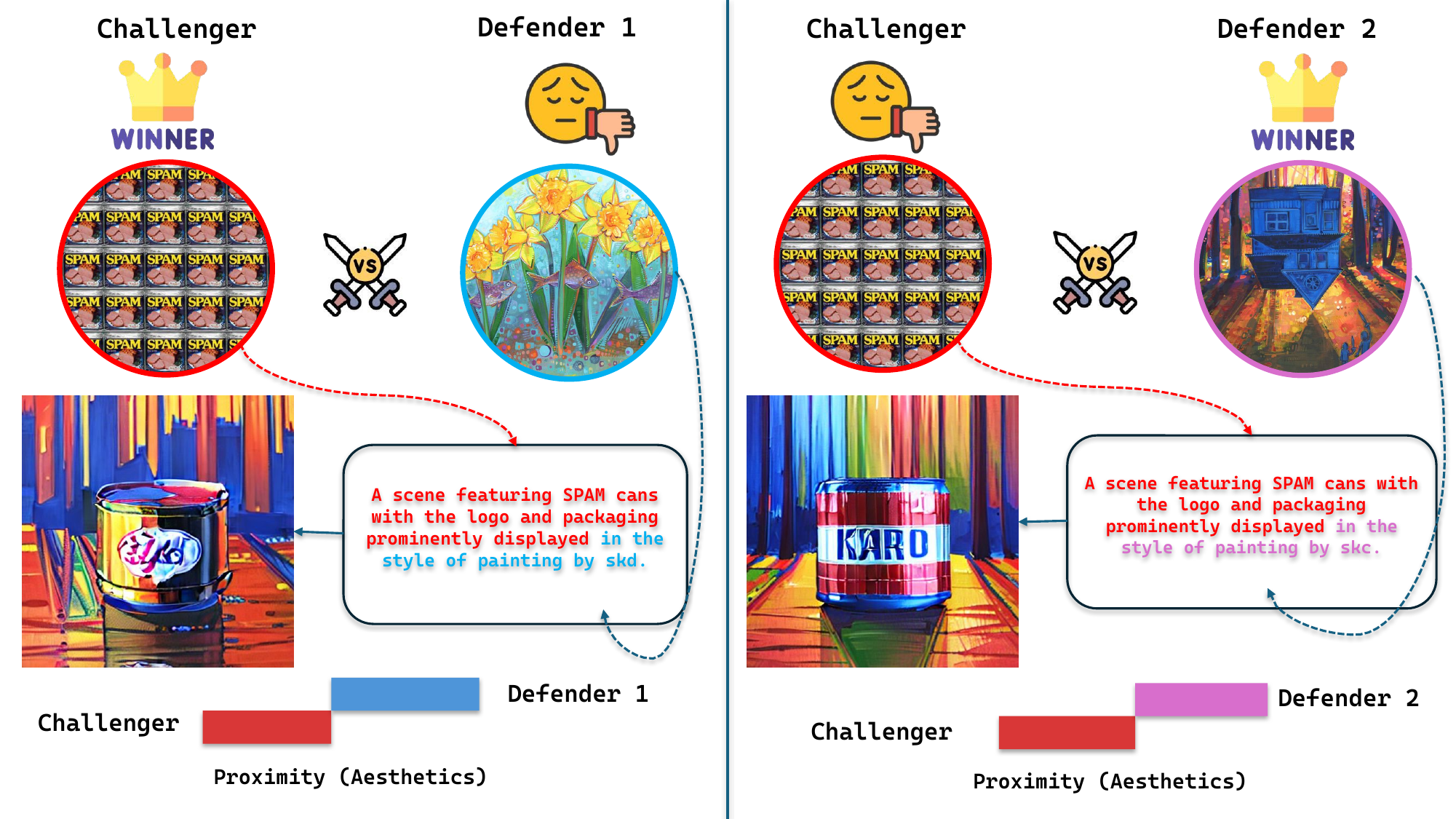}}
    \caption{Illustration of a \emph{Motif Duel} for SD v1.5 personalized on artworks by Gwenn Seemel, evaluated using aesthetics-based proximity (LPIPS; lower scores indicate closer proximity). The results demonstrate how combining the motifs of the canonical challenger \textit{Spam} by Andy Warhol with the artist's \textit{Springtime in Lambertville} (Defender 1, \texttt{skd}) and \textit{Siblings} (Defender 2, \texttt{skc}) produces stylistically distinct outputs.}
    \label{fig:pduel_sd_aes_gwenn}
  \end{minipage}
  \par\vspace{1pt}
  \begin{minipage}[t]{0.48\textwidth}
    \centering
    \resizebox{0.85\linewidth}{!}{\includegraphics[]{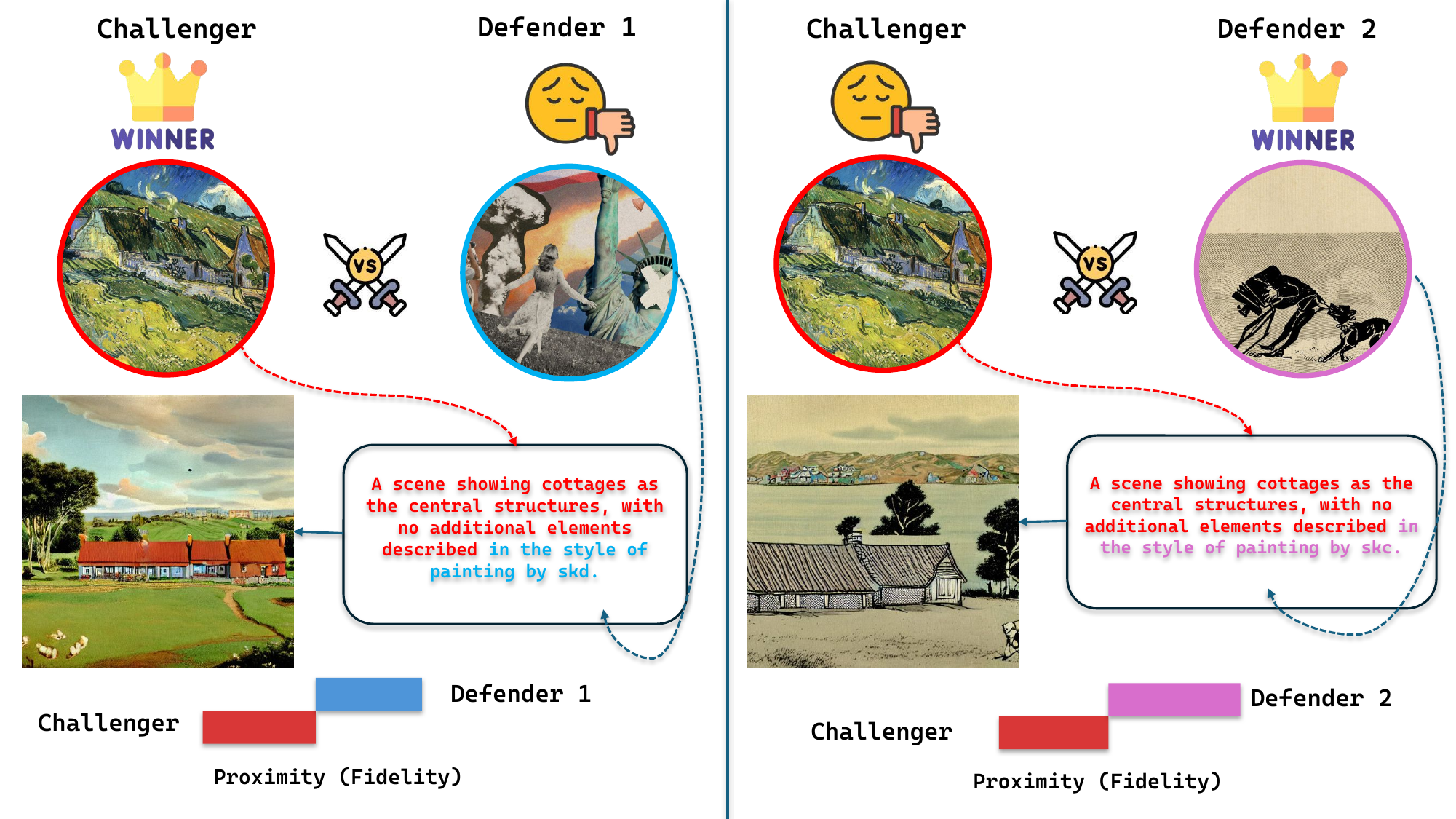}}
    \caption{Illustration of a \emph{Motif Duel} for SD v1.5 personalized on artworks by Trent Anthony Francis, evaluated using fidelity-based proximity (CSD). The results demonstrate how combining the motifs of the canonical challenger \textit{A Group of Cottages} by Vincent van Gogh with the artist's \textit{The Star Strangled Banner (Deviations)} (Defender 1, \texttt{skd}) and \textit{The Moon and Sixpence} (Defender 2, \texttt{skc}) produces stylistically distinct outputs.}
    \label{fig:pduel_sd_fid_trent}
  \end{minipage}\hfill
  \begin{minipage}[t]{0.48\textwidth}
    \centering
    \resizebox{0.85\linewidth}{!}{\includegraphics[]{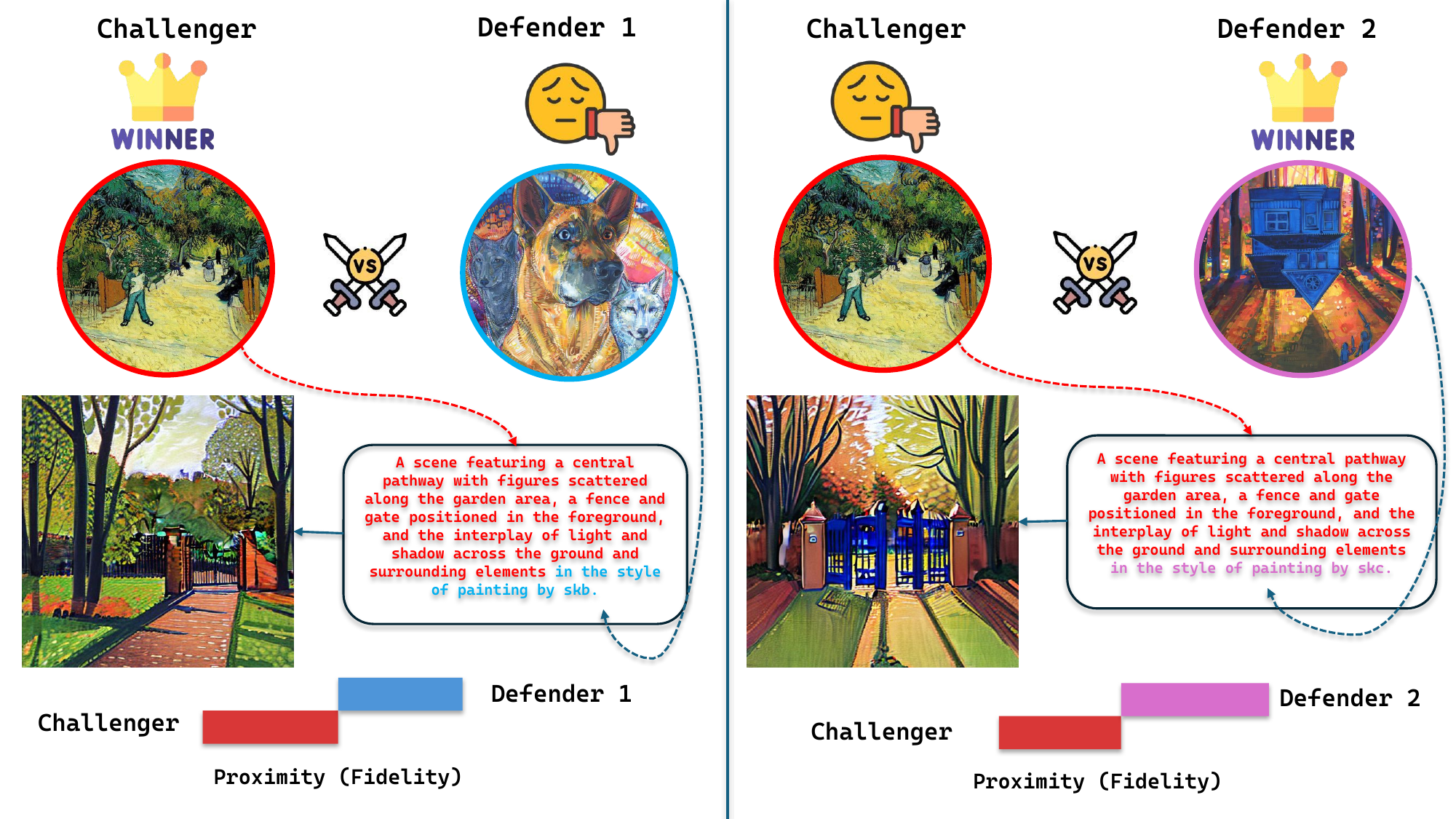}}
    \caption{Illustration of a \emph{Motif Duel} for SD v1.5 personalized on artworks by Gwenn Seemel, evaluated using fidelity-based proximity (CSD). The results demonstrate how combining the motifs of the canonical challenger \textit{Entrance to the Public Garden in Arles} by Vincent van Gogh with the artist's \textit{Sally with Wolves} (Defender 1, \texttt{skb}) and \textit{Siblings} (Defender 2, \texttt{skc}) produces stylistically distinct outputs.}
    \label{fig:pduel_sd_fid_gwenn}
  \end{minipage}
\end{figure*}

\begin{figure*}[p]
  \centering
  \begin{minipage}[t]{0.48\textwidth}
    \centering
    \resizebox{0.85\linewidth}{!}{\includegraphics[]{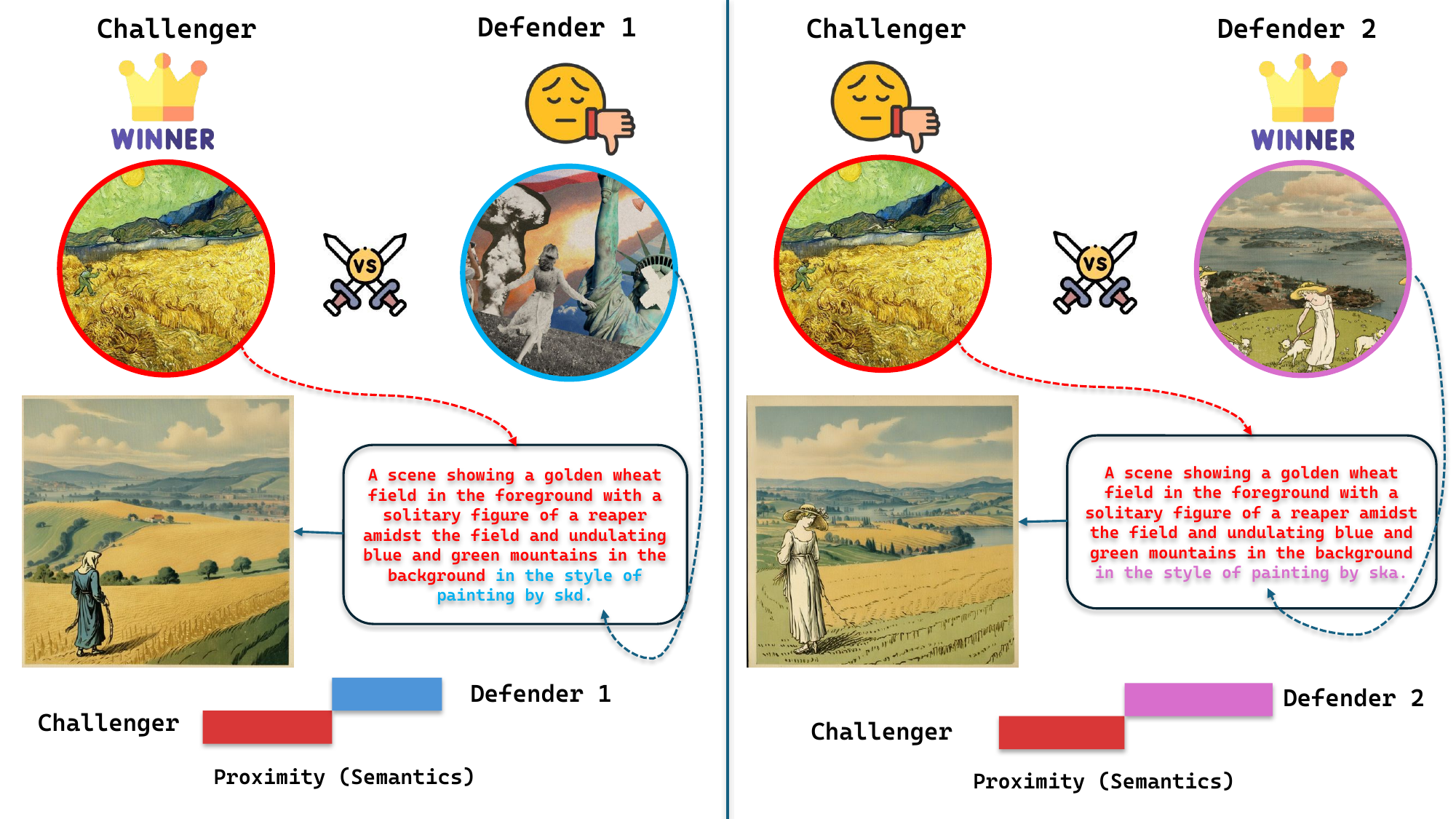}}
    \caption{Illustration of a \emph{Motif Duel} for SDXL personalized on artworks by Trent Anthony Francis, evaluated using semantics-based proximity (CLIP). The results demonstrate how combining the motifs of the canonical challenger \textit{Wheat Field with Reaper and Sun} by Vincent van Gogh with the artist's \textit{The Star Strangled Banner (Deviations)} (Defender 1, \texttt{skd}) and \textit{Meadow May} (Defender 2, \texttt{ska}) produces stylistically distinct outputs.}
    \label{fig:pduel_sdxl_sem_trent}
  \end{minipage}\hfill
  \begin{minipage}[t]{0.48\textwidth}
    \centering
    \resizebox{0.85\linewidth}{!}{\includegraphics[]{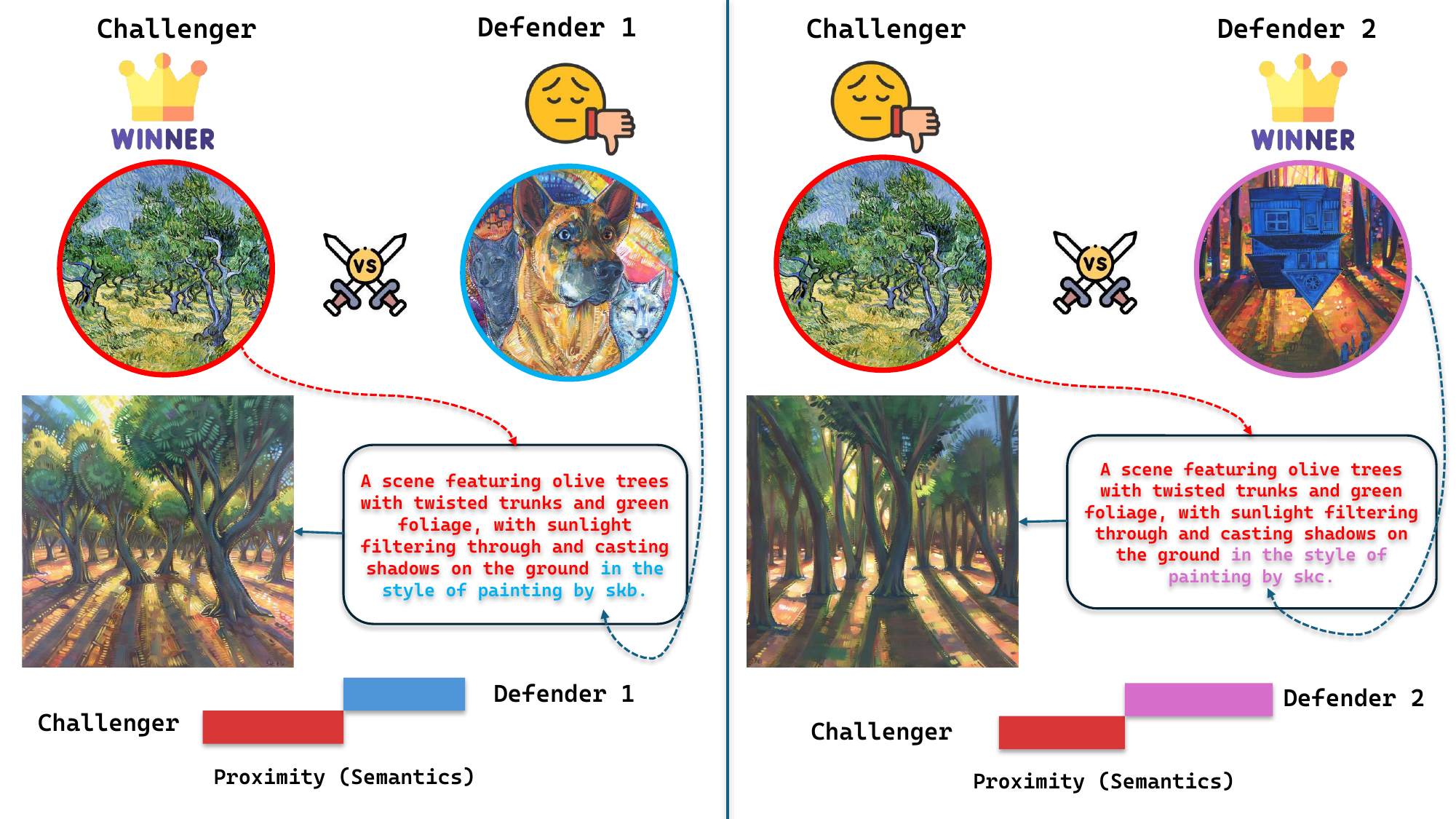}}
    \caption{Illustration of a \emph{Motif Duel} for SDXL personalized on artworks by Gwenn Seemel, evaluated using semantics-based proximity (CLIP). The results demonstrate how combining the motifs of the canonical challenger \textit{Olive Grove} by Vincent van Gogh with the artist's \textit{Sally with Wolves} (Defender 1, \texttt{skb}) and \textit{Siblings} (Defender 2, \texttt{skc}) produces stylistically distinct outputs.}
    \label{fig:pduel_sdxl_sem_gwenn}
  \end{minipage}
  \par\vspace{1pt}
  \begin{minipage}[t]{0.48\textwidth}
    \centering
    \resizebox{0.85\linewidth}{!}{\includegraphics[]{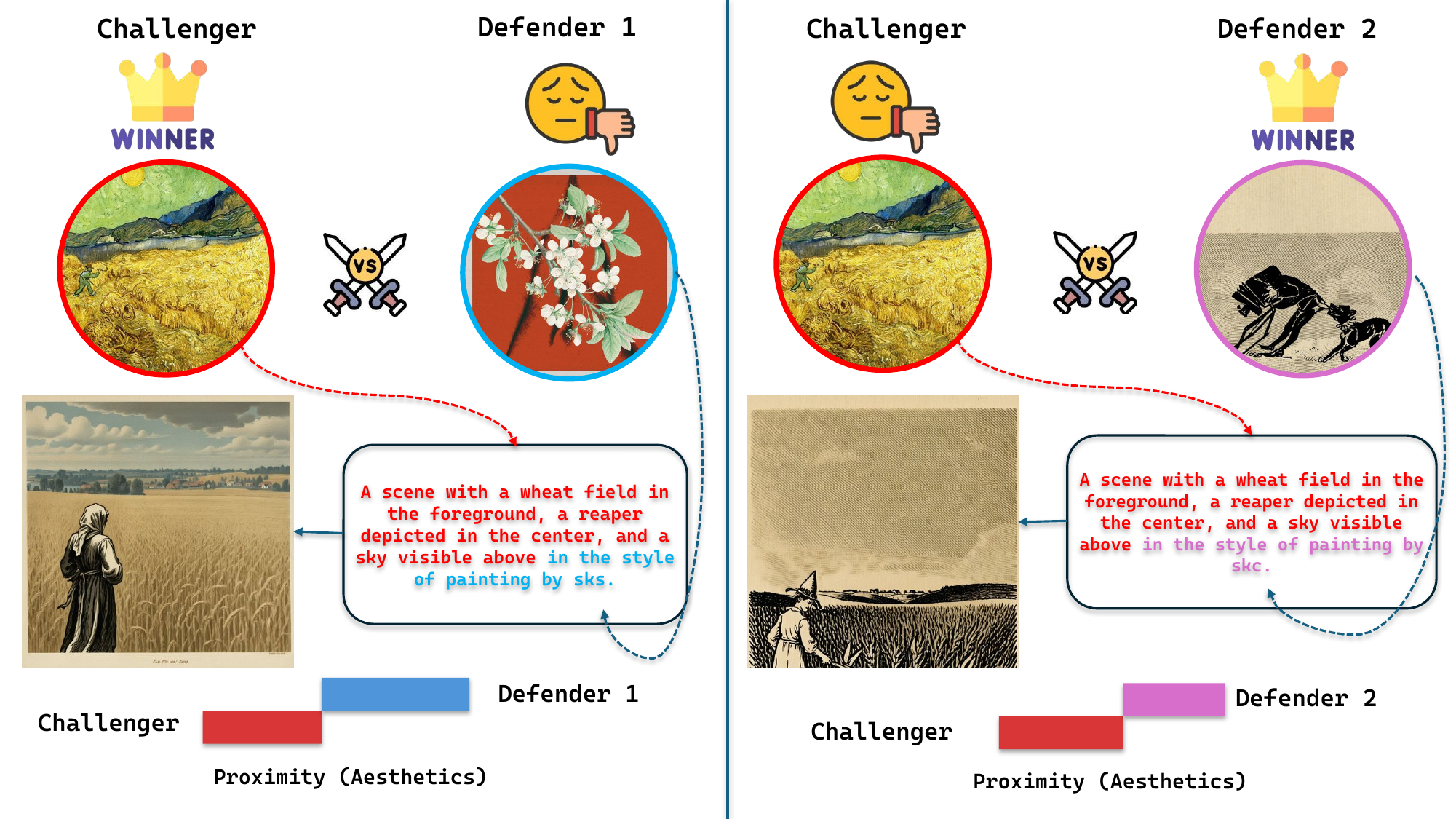}}
    \caption{Illustration of a \emph{Motif Duel} for SDXL personalized on artworks by Trent Anthony Francis, evaluated using aesthetics-based proximity (LPIPS; lower scores indicate closer proximity). The results demonstrate how combining the motifs of the canonical challenger \textit{Wheat Field with Reaper and Sun} by Vincent van Gogh with the artist's \textit{Essays on Nature and Growing (Path IV)} (Defender 1, \texttt{sks}) and \textit{The Moon and Sixpence} (Defender 2, \texttt{skc}) produces stylistically distinct outputs.}
    \label{fig:pduel_sdxl_aes_trent}
  \end{minipage}\hfill
  \begin{minipage}[t]{0.48\textwidth}
    \centering
    \resizebox{0.85\linewidth}{!}{\includegraphics[]{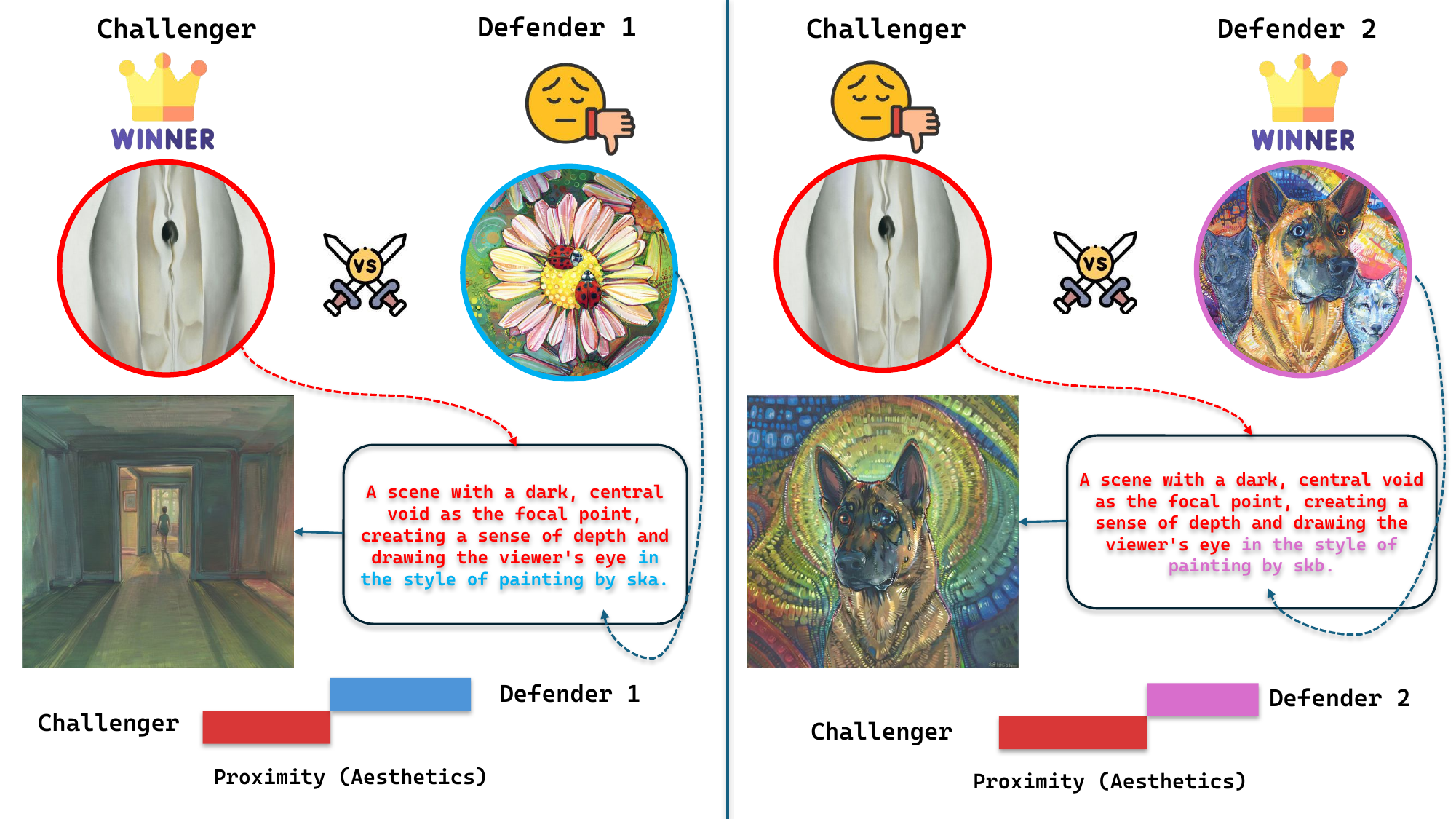}}
    \caption{Illustration of a \emph{Motif Duel} for SDXL personalized on artworks by Gwenn Seemel, evaluated using aesthetics-based proximity (LPIPS; lower scores indicate closer proximity). The results demonstrate how combining the motifs of the canonical challenger \textit{Slightly Open Clam Shell} by Georgia O'Keeffe with the artist's \textit{Ladybugs on Daisies} (Defender 1, \texttt{ska}) and \textit{Sally with Wolves} (Defender 2, \texttt{skb}) produces stylistically distinct outputs.}
    \label{fig:pduel_sdxl_aes_gwenn}
  \end{minipage}
  \par\vspace{1pt}
  \begin{minipage}[t]{0.48\textwidth}
    \centering
    \resizebox{0.85\linewidth}{!}{\includegraphics[]{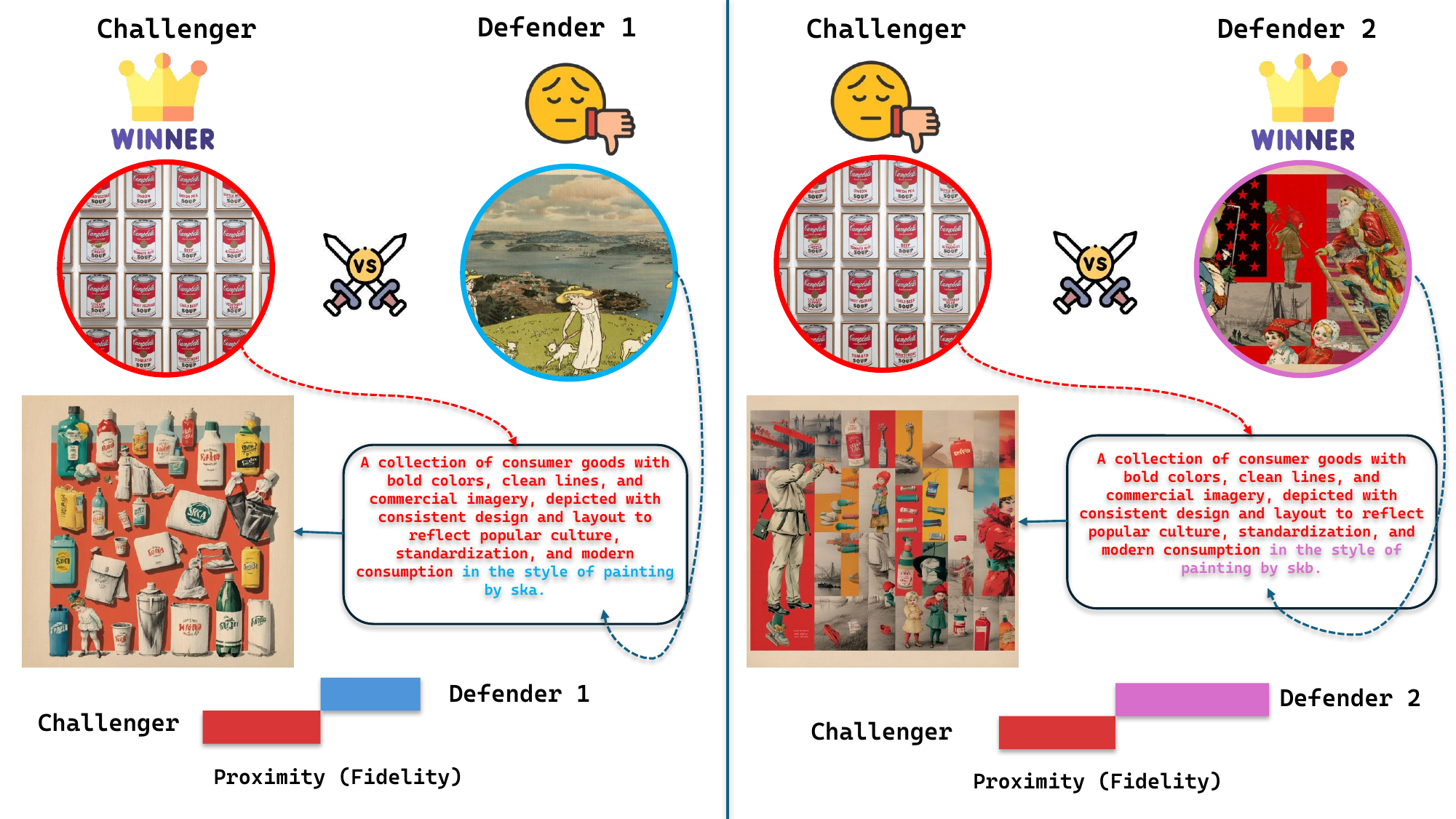}}
    \caption{Illustration of a \emph{Motif Duel} for SDXL personalized on artworks by Trent Anthony Francis, evaluated using fidelity-based proximity (CSD). The results demonstrate how combining the motifs of the canonical challenger \textit{Campbell's Soup Cans} by Andy Warhol with the artist's \textit{Meadow May} (Defender 1, \texttt{ska}) and \textit{Star Strangled Banner} (Defender 2, \texttt{skb}) produces stylistically distinct outputs.}
    \label{fig:pduel_sdxl_fid_trent}
  \end{minipage}\hfill
  \begin{minipage}[t]{0.48\textwidth}
    \centering
    \resizebox{0.85\linewidth}{!}{\includegraphics[]{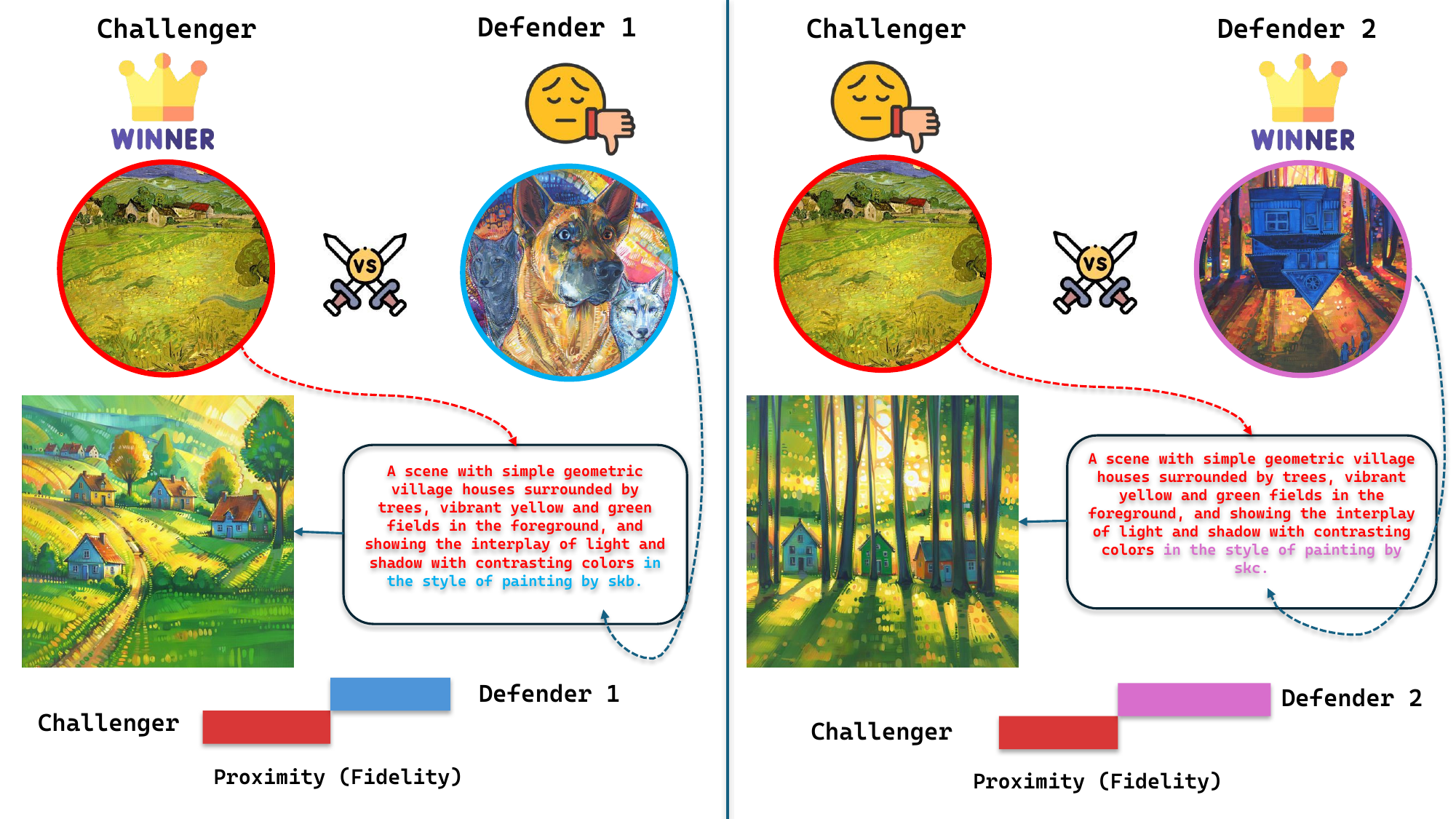}}
    \caption{Illustration of a \emph{Motif Duel} for SDXL personalized on artworks by Gwenn Seemel, evaluated using fidelity-based proximity (CSD). The results demonstrate how combining the motifs of the canonical challenger \textit{View of Vessenots near Auvers} by Vincent van Gogh with the artist's \textit{Sally with Wolves} (Defender 1, \texttt{skb}) and \textit{Siblings} (Defender 2, \texttt{skc}) produces stylistically distinct outputs.}
    \label{fig:pduel_sdxl_fid_gwenn}
  \end{minipage}
\end{figure*}

\begin{figure*}[p]
  \centering
  \begin{minipage}[t]{0.48\textwidth}
    \centering
    \resizebox{0.85\linewidth}{!}{\includegraphics[]{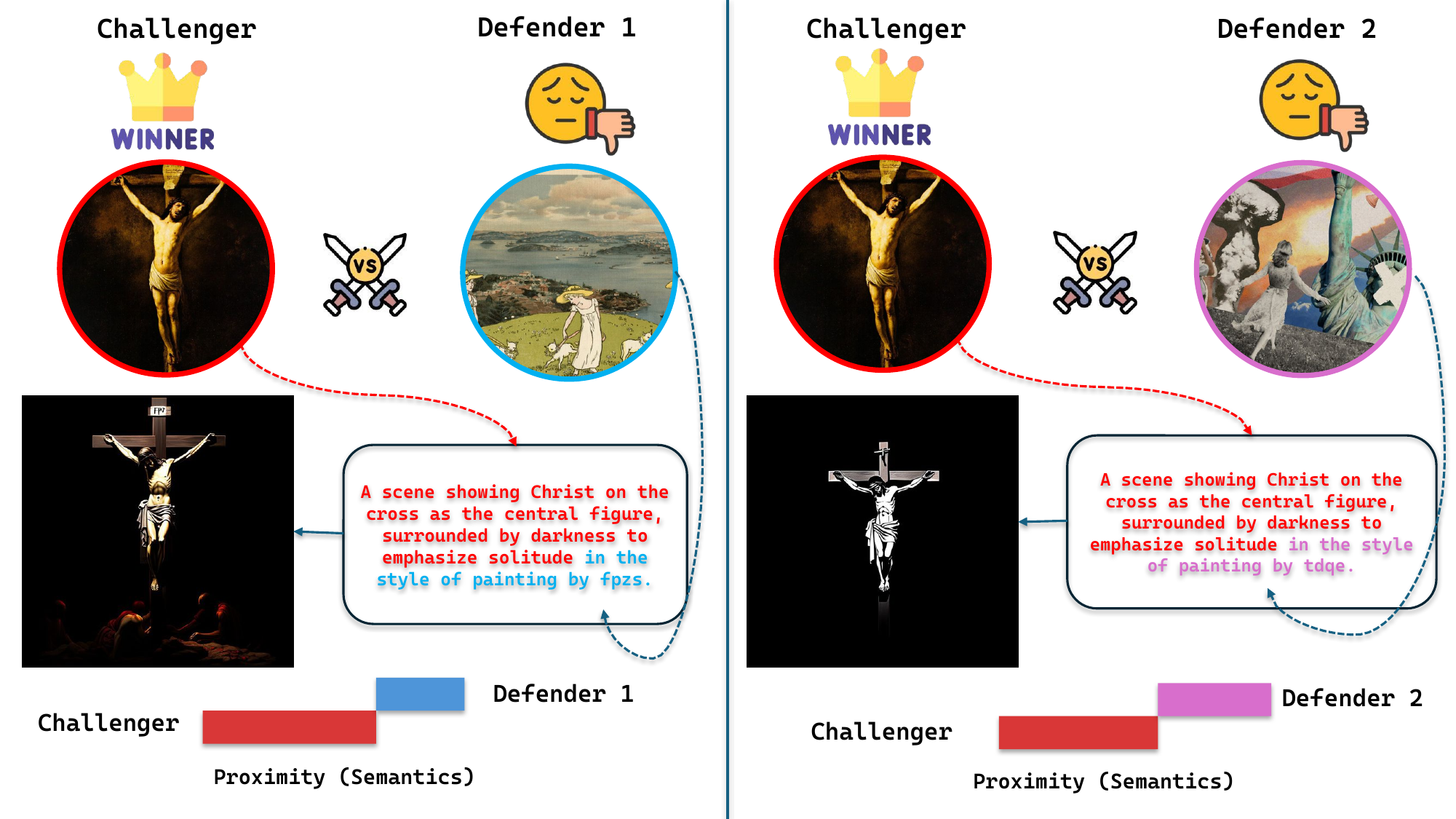}}
    \caption{Illustration of a \emph{Motif Duel} for SANA-1.5 personalized on artworks by Trent Anthony Francis, evaluated using semantics-based proximity (CLIP). The results demonstrate how combining the motifs of the canonical challenger \textit{Christ on the Cross} by Rembrandt with the artist's \textit{Meadow May} (Defender 1, \texttt{fpzs}) and \textit{The Star Strangled Banner (Deviations)} (Defender 2, \texttt{tdqe}) produces stylistically distinct outputs.}
    \label{fig:pduel_sana_sem_trent}
  \end{minipage}\hfill
  \begin{minipage}[t]{0.48\textwidth}
    \centering
    \resizebox{0.85\linewidth}{!}{\includegraphics[]{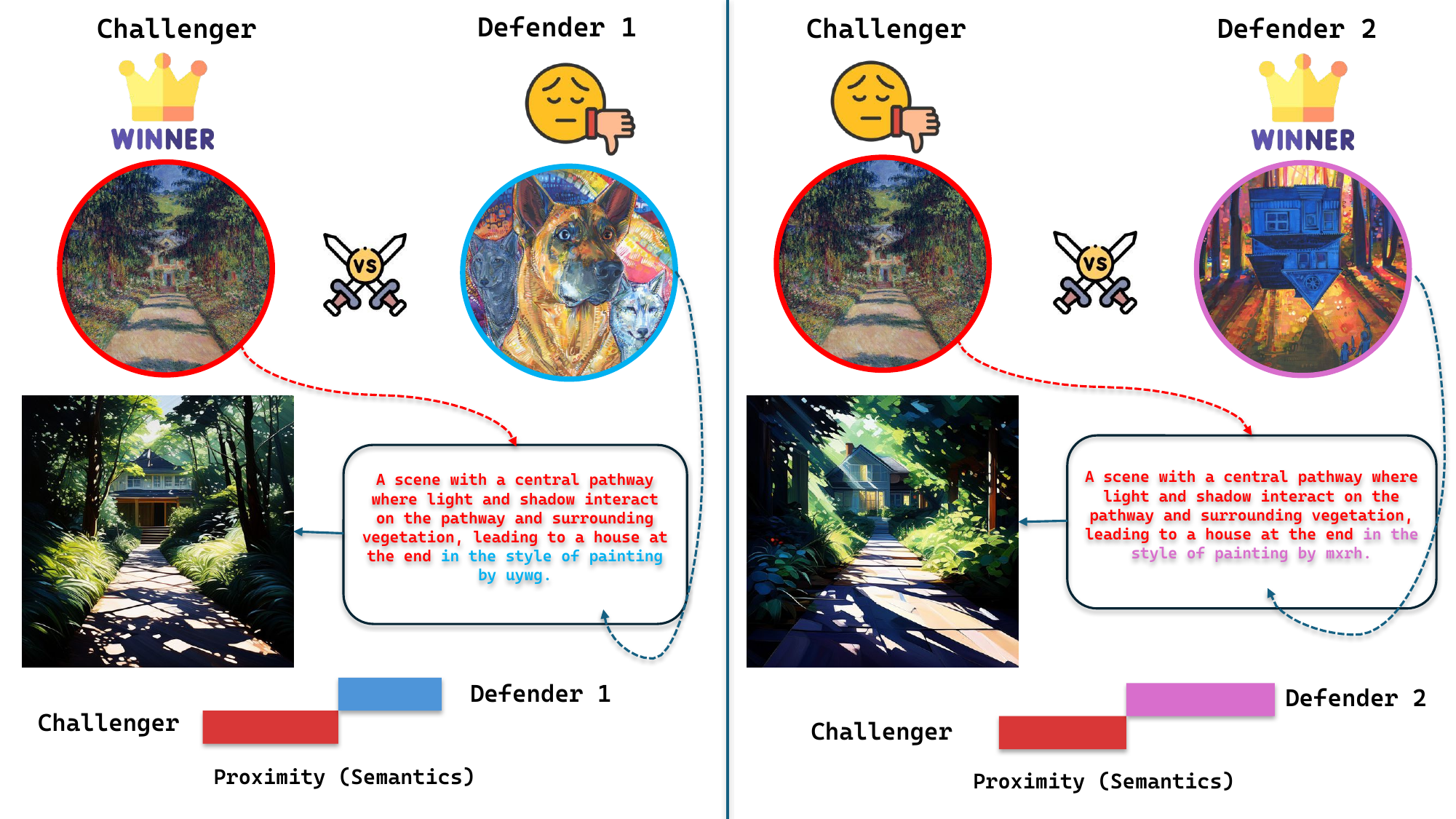}}
    \caption{Illustration of a \emph{Motif Duel} for SANA-1.5 personalized on artworks by Gwenn Seemel, evaluated using semantics-based proximity (CLIP). The results demonstrate how combining the motifs of the canonical challenger \textit{Pathway in Monet's Garden at Giverny} by Claude Monet with the artist's \textit{Sally with Wolves} (Defender 1, \texttt{uywg}) and \textit{Siblings} (Defender 2, \texttt{mxrh}) produces stylistically distinct outputs.}
    \label{fig:pduel_sana_sem_gwenn}
  \end{minipage}
  \par\vspace{1pt}
  \begin{minipage}[t]{0.48\textwidth}
    \centering
    \resizebox{0.85\linewidth}{!}{\includegraphics[]{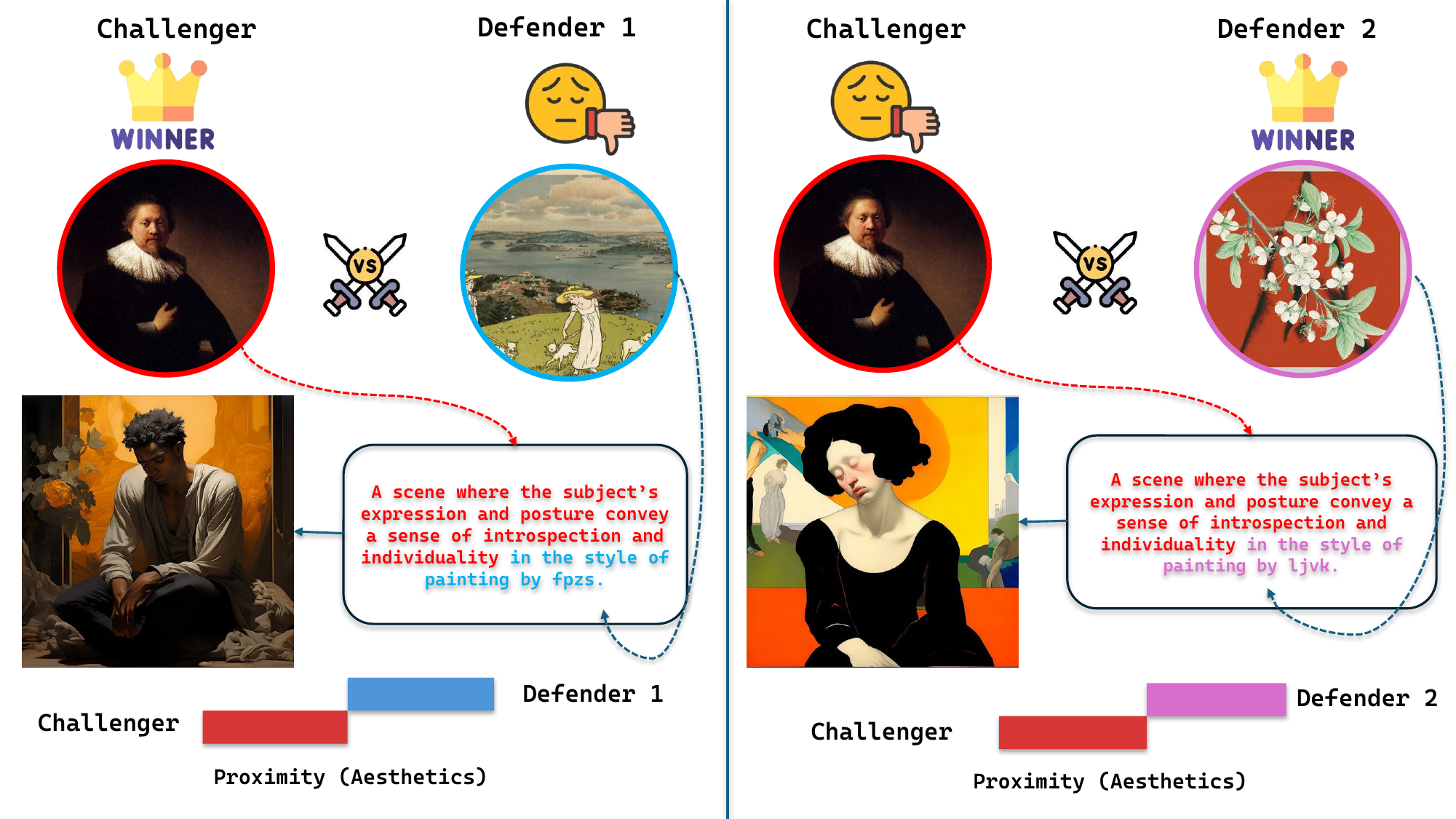}}
    \caption{Illustration of a \emph{Motif Duel} for SANA-1.5 personalized on artworks by Trent Anthony Francis, evaluated using aesthetics-based proximity (LPIPS; lower scores indicate closer proximity). The results demonstrate how combining the motifs of the canonical challenger \textit{Portrait of a Man} by Rembrandt with the artist's \textit{Meadow May} (Defender 1, \texttt{fpzs}) and \textit{Essays on Nature and Growing (Path IV)} (Defender 2, \texttt{ljvk}) produces stylistically distinct outputs.}
    \label{fig:pduel_sana_aes_trent}
  \end{minipage}\hfill
  \begin{minipage}[t]{0.48\textwidth}
    \centering
    \resizebox{0.85\linewidth}{!}{\includegraphics[]{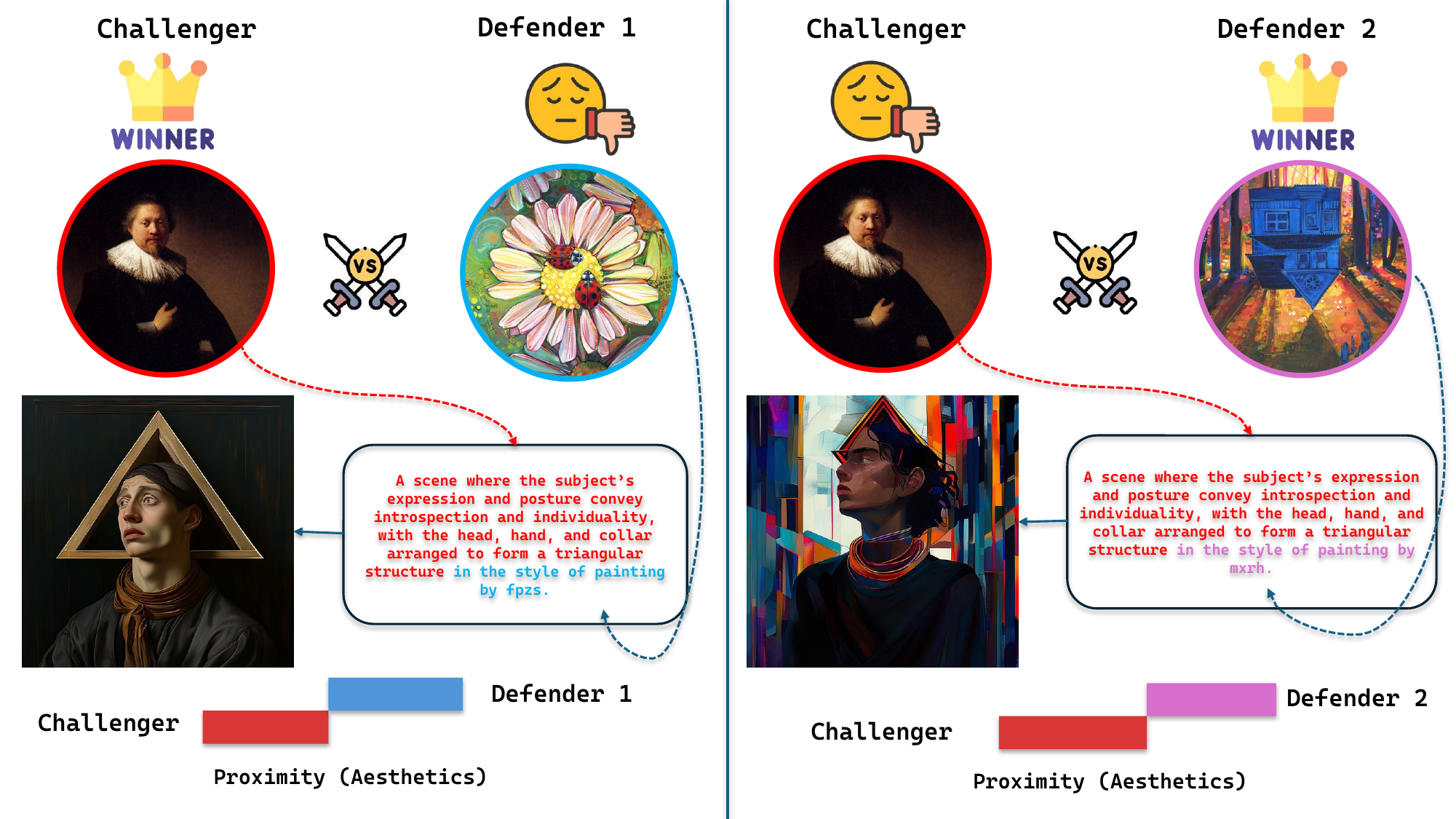}}
    \caption{Illustration of a \emph{Motif Duel} for SANA-1.5 personalized on artworks by Gwenn Seemel, evaluated using aesthetics-based proximity (LPIPS; lower scores indicate closer proximity). The results demonstrate how combining the motifs of the canonical challenger \textit{Portrait of a Man} by Rembrandt with the artist's \textit{Ladybugs on Daisies} (Defender 1, \texttt{fpzs}) and \textit{Siblings} (Defender 2, \texttt{mxrh}) produces stylistically distinct outputs.}
    \label{fig:pduel_sana_aes_gwenn}
  \end{minipage}
  \par\vspace{1pt}
  \begin{minipage}[t]{0.48\textwidth}
    \centering
    \resizebox{0.85\linewidth}{!}{\includegraphics[]{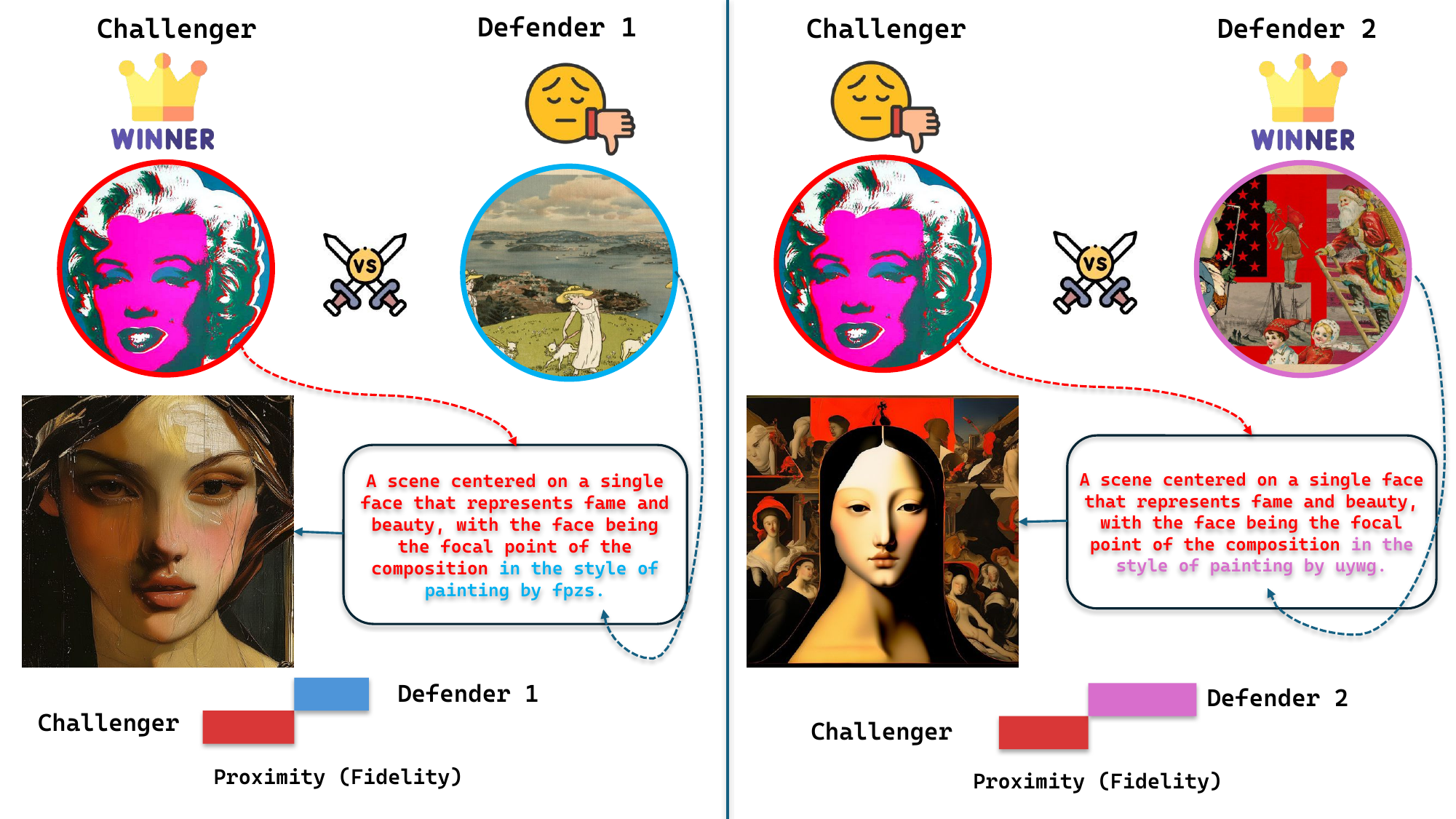}}
    \caption{Illustration of a \emph{Motif Duel} for SANA-1.5 personalized on artworks by Trent Anthony Francis, evaluated using fidelity-based proximity (CSD). The results demonstrate how combining the motifs of the canonical challenger \textit{Marilyn Blue} by Andy Warhol with the artist's \textit{Meadow May} (Defender 1, \texttt{fpzs}) and \textit{Star Strangled Banner} (Defender 2, \texttt{uywg}) produces stylistically distinct outputs.}
    \label{fig:pduel_sana_fid_trent}
  \end{minipage}\hfill
  \begin{minipage}[t]{0.48\textwidth}
    \centering
    \resizebox{0.85\linewidth}{!}{\includegraphics[]{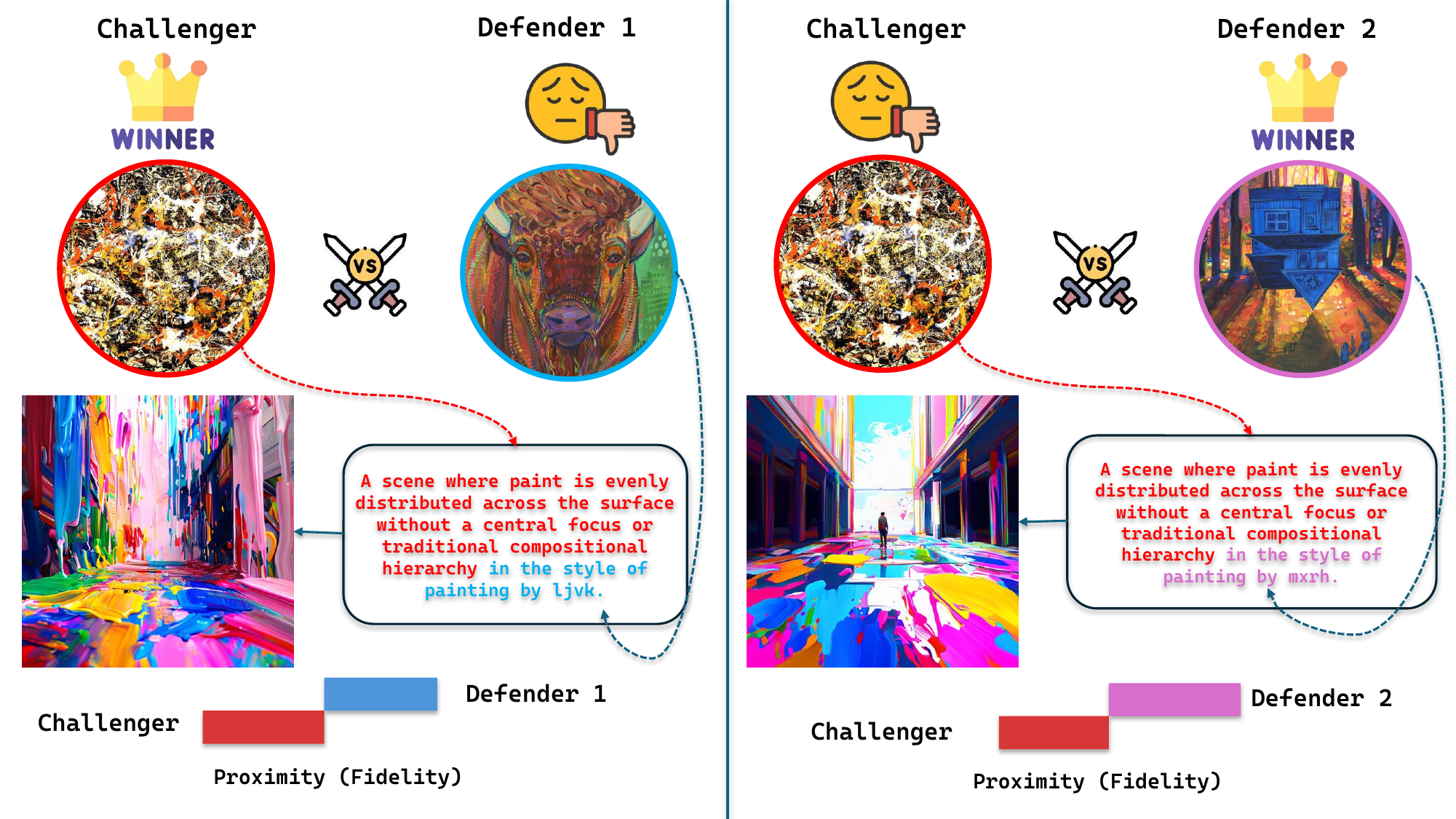}}
    \caption{Illustration of a \emph{Motif Duel} for SANA-1.5 personalized on artworks by Gwenn Seemel, evaluated using fidelity-based proximity (CSD). The results demonstrate how combining the motifs of the canonical challenger \textit{Convergence (Number 10)} by Jackson Pollock with the artist's \textit{American Bison} (Defender 1, \texttt{ljvk}) and \textit{Siblings} (Defender 2, \texttt{mxrh}) produces stylistically distinct outputs.}
    \label{fig:pduel_sana_fid_gwenn}
  \end{minipage}
\end{figure*}

\paragraph{Motif Duels on the personalized models.} Figs.~\ref{fig:pduel_sd_sem_trent}--\ref{fig:pduel_sana_fid_gwenn} illustrate instances of Motif Duels on the personalized models, whose outcomes are aggregated in Block~B of the leakage cards. In each instance, the motif of a canonical challenger is combined with two of the artist's artworks, showing the canonical artwork blending into the artist's artworks to varying degrees.

\subsubsection{Validation}
\label{sec:val_details}

This section details the strengthening and weakening interventions used to validate ArtArena (Section~\ref{sec:Val_AA}). For each model and proximity metric, both interventions start from the same base model and target an artwork $w$ of the corresponding Influence Ledger obtained during Canonical Influence Profiling. Both use LoRA adapters~\citep{hu2022lora} of rank $64$ and the training prompt \texttt{<title> in the style of <artist>}, matching the artwork's defender template in Motif Duels. The fine-tuned and unlearned models are evaluated as described in Section~\ref{sec:Val_AA}, with results reported in Tabs.~\ref{tab:val_sd_clip_imit}--\ref{tab:val_sana_csd_def}.

\paragraph{Strengthening (fine-tuning).} To strengthen the bottom-1 artwork $w$, we fine-tune the base model on $w$ using DreamBooth~\citep{ruiz2023dreambooth}, following the configuration of the personalization experiments (Table~\ref{tab:setup}): LoRA adapters of rank $64$ with $\alpha = 32$, batch size $1$, and learning rate $1 \times 10^{-4}$, at a resolution of $512 \times 512$ for SD v1.5 and $1024 \times 1024$ for SDXL and SANA-1.5. We select the checkpoint based on the balance between strengthening the target artwork and preserving non-target artworks. Checkpoint selection considers the gain in the target artwork's fitness alongside the fitness of the remaining \texttt{FitSet} artworks, favoring checkpoints that raise the fitness of the target artwork while minimizing changes to unrelated artworks. Each checkpoint is evaluated using the corresponding proximity metric to identify the strengthen--retain trade-off.

\paragraph{Weakening (unlearning).} To weaken the top-1 artwork $w$, we unlearn it from the base model using SurgUn~\citep{ranjan2026forgetting}, applied to attention blocks of each backbone: \texttt{up\_blocks.1.attentions.0} for SD v1.5, \texttt{up\_blocks.0.attentions.2} for SDXL, and \texttt{transformer\_blocks.7} for SANA-1.5. We select the checkpoint based on the balance between suppressing the target artwork and preserving non-target artworks. Checkpoint selection considers unlearning effectiveness alongside retention metrics, favoring checkpoints that suppress the target artwork while minimizing degradation of unrelated artworks. Each checkpoint is evaluated using the corresponding benchmark metrics to identify the erase--retain trade-off.

\begin{figure*}[t]
\centering

\begin{small} 
\begin{subfigure}[t]{0.22\linewidth}
    \centering
    \includegraphics[width=\linewidth]{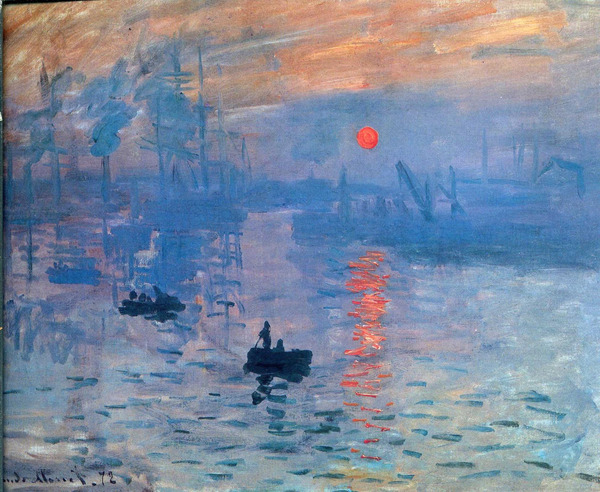}
    \caption{Reference}
\end{subfigure}
\hfill
\begin{subfigure}[t]{0.22\linewidth}
    \centering
    \includegraphics[width=\linewidth]{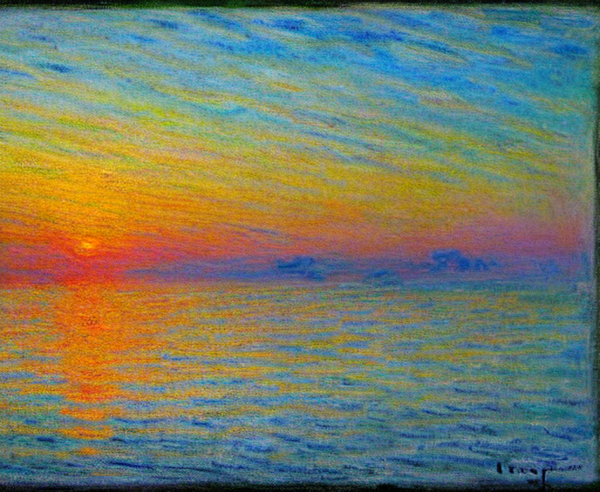}
    \caption{Generated with SD v1.5}
\end{subfigure}
\hfill
\begin{subfigure}[t]{0.22\linewidth}
    \centering
    \includegraphics[width=\linewidth]{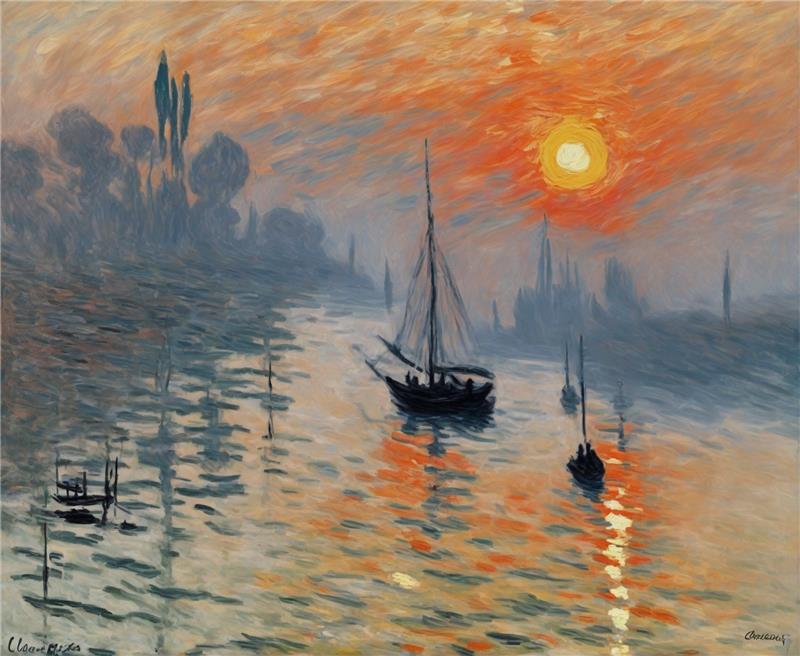}
    \caption{Generated with SDXL}
\end{subfigure}
\hfill
\begin{subfigure}[t]{0.22\linewidth}
    \centering
    \includegraphics[width=\linewidth]{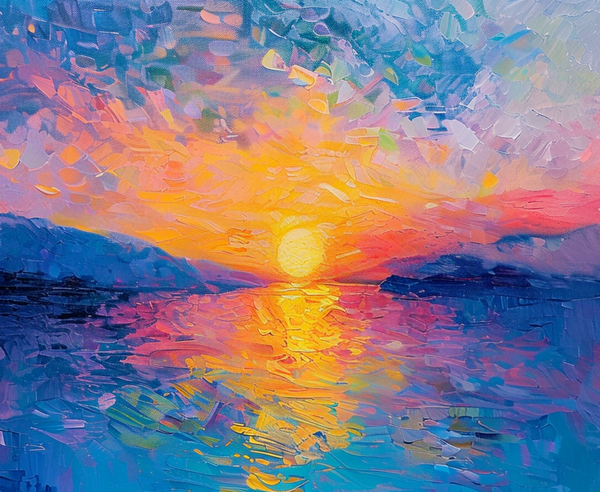}
    \caption{Generated with SANA-1.5}
\end{subfigure}
\end{small}

\vspace{0.8em}

\begin{tikzpicture}[x=1cm,y=0.7cm]

\def\barwidth{3.0}
\def\barheight{0.45}
\def\gap{1.0}
\def\scoreoffset{0.12}

\def\xB{\barwidth+\gap}            
\def\xA{0}                         
\def\xC{2*\barwidth+2*\gap}        


\def\VSB{0.89}
\def\SIB{0.60}
\def\ATB{0.64}

\def\VSA{0.39}
\def\SIA{0.84}
\def\ATA{0.70}

\def\VSC{0.75}
\def\SIC{0.81}
\def\ATC{0.22}

\definecolor{bgcolor}{RGB}{238,236,232}
\definecolor{stylecolor}{RGB}{186,108,83}
\definecolor{semanticcolor}{RGB}{108,123,158}
\definecolor{affectivecolor}{RGB}{122,146,118}

\node[font=\small] at (\xA+0.5*\barwidth,3.3) {Reference vs SD v1.5};
\node[font=\small] at (\xB+0.5*\barwidth,3.3) {Reference vs SDXL}; 
\node[font=\small] at (\xC+0.5*\barwidth,3.3) {Reference vs SANA-1.5};


\node[anchor=east] at (-0.3,2.6) {Semantics     };
\node[anchor=east] at (-0.3,1.6) {Aesthetics    };
\node[anchor=east] at (-0.3,0.6) {Fidelity      };

\foreach \y in {2.6,1.6,0.6} {
    \draw[rounded corners=3pt, fill=bgcolor, draw=none]
        (\xA,\y) rectangle (\xA+\barwidth,\y+\barheight);
    \draw[rounded corners=3pt, fill=bgcolor, draw=none]
        (\xB,\y) rectangle (\xB+\barwidth,\y+\barheight);
    \draw[rounded corners=3pt, fill=bgcolor, draw=none]
        (\xC,\y) rectangle (\xC+\barwidth,\y+\barheight);
}

\draw[rounded corners=3pt, fill=stylecolor]
    (\xB,2.6) rectangle ({\xB+\VSB*\barwidth},2.6+\barheight);
\draw[rounded corners=3pt, fill=semanticcolor]
    (\xB,1.6) rectangle ({\xB+\SIB*\barwidth},1.6+\barheight);
\draw[rounded corners=3pt, fill=affectivecolor]
    (\xB,0.6) rectangle ({\xB+\ATB*\barwidth},0.6+\barheight);

\node[anchor=west, font=\small\color{stylecolor!80!black}]
    at ({\xB+\VSB*\barwidth+\scoreoffset},2.80) {0.89};
\node[anchor=west, font=\small\color{semanticcolor!80!black}]
    at ({\xB+\SIB*\barwidth+\scoreoffset},1.83) {0.60};
\node[anchor=west, font=\small\color{affectivecolor!80!black}]
    at ({\xB+\ATB*\barwidth+\scoreoffset},0.83) {0.64};
    
\draw[rounded corners=3pt, fill=stylecolor]
    (\xA,2.6) rectangle ({\xA+\VSA*\barwidth},2.6+\barheight);
\draw[rounded corners=3pt, fill=semanticcolor]
    (\xA,1.6) rectangle ({\xA+\SIA*\barwidth},1.6+\barheight);
\draw[rounded corners=3pt, fill=affectivecolor]
    (\xA,0.6) rectangle ({\xA+\ATA*\barwidth},0.6+\barheight);

\node[anchor=west, font=\small\color{stylecolor!80!black}]
    at ({\xA+\VSA*\barwidth+\scoreoffset},2.83) {0.39};
\node[anchor=west, font=\small\color{semanticcolor!80!black}]
    at ({\xA+\SIA*\barwidth+\scoreoffset},1.83) {0.84};
\node[anchor=west, font=\small\color{affectivecolor!80!black}]
    at ({\xA+\ATA*\barwidth+\scoreoffset},0.83) {0.70};    

\draw[rounded corners=3pt, fill=stylecolor]
    (\xC,2.6) rectangle ({\xC+\VSC*\barwidth},2.6+\barheight);
\draw[rounded corners=3pt, fill=semanticcolor]
    (\xC,1.6) rectangle ({\xC+\SIC*\barwidth},1.6+\barheight);
\draw[rounded corners=3pt, fill=affectivecolor]
    (\xC,0.6) rectangle ({\xC+\ATC*\barwidth},0.6+\barheight);

\node[anchor=west, font=\small\color{stylecolor!80!black}]
    at ({\xC+\VSC*\barwidth+\scoreoffset},2.80) {0.75};
\node[anchor=west, font=\small\color{semanticcolor!80!black}]
    at ({\xC+\SIC*\barwidth+\scoreoffset},1.83) {0.81};
\node[anchor=west, font=\small\color{affectivecolor!80!black}]
    at ({\xC+\ATC*\barwidth+\scoreoffset},0.83) {0.22};

\end{tikzpicture}

\caption{\textbf{Imitation of the artwork} \textit{Impression, Sunrise} by Claude Monet using the SDXL, SD v1.5, and SANA-1.5 text-to-image models. The prompt used for imitation was: ``Impression, Sunrise in the style of Claude Monet.'' We evaluate the degree of imitation using CLIP (Semantics), LPIPS (Aesthetics), and CSD (Fidelity). \textit{Higher CLIP and CSD scores indicate stronger imitation, whereas lower LPIPS scores indicate stronger imitation.} Among the generated samples, image (c) appears visually closest to the reference image (a), exhibiting similar structural and stylistic characteristics, including color palette, objects placement, and finer details such as the reflection of the sun on the water, as collectively captured by the proximity-based metrics.}
\label{fig: imitation_example}
\end{figure*}
\begin{figure*}[t]
    \centering
    \begin{subfigure}{0.32\textwidth}
        \centering
        \includegraphics[width=\linewidth]{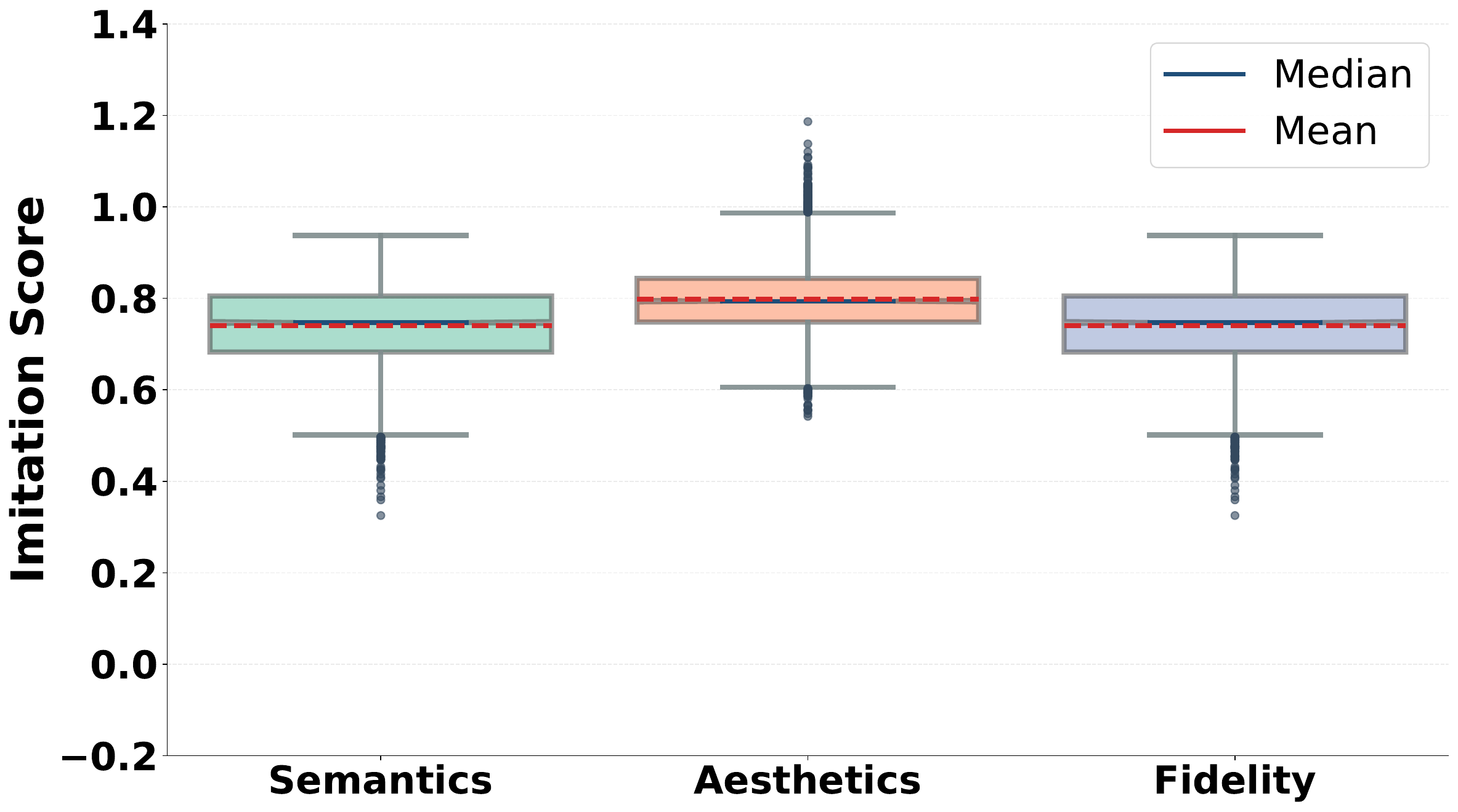}
        \caption{SD v1.5}
    \end{subfigure}
    \begin{subfigure}{0.32\textwidth}
        \centering
        \includegraphics[width=\linewidth]{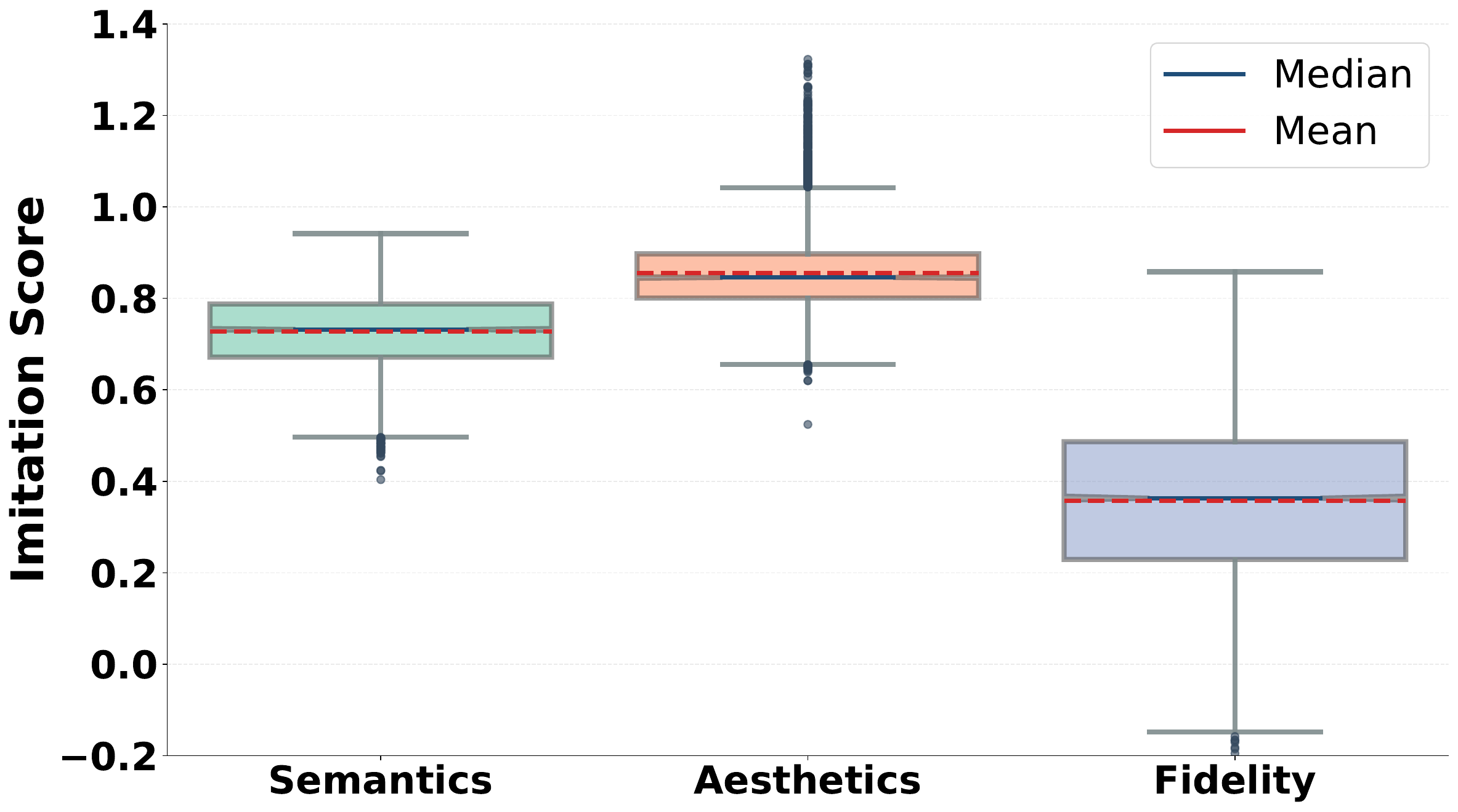}
        \caption{SDXL}
    \end{subfigure}
    \begin{subfigure}{0.32\textwidth}
        \centering
        \includegraphics[width=\linewidth]{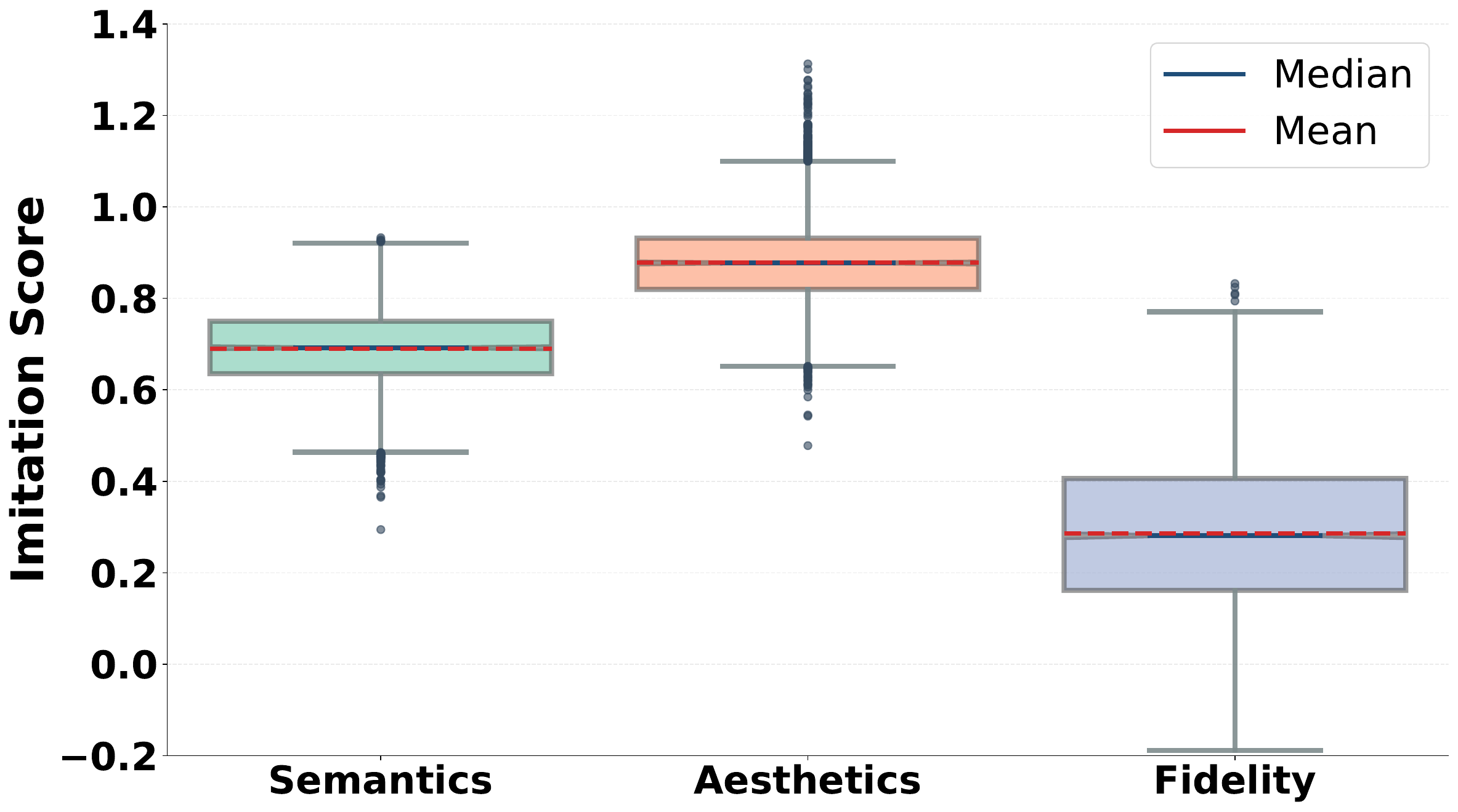}
        \caption{SANA-1.5}
    \end{subfigure}
    \caption{We present the distribution of fitness scores for SD v1.5, SDXL, and SANA‑1.5 across three proximity metrics, evaluated on 8,737 artworks from twenty popular artists. For the Semantics and Fidelity metrics, a higher mean indicates better performance, whereas for Aesthetics, a lower mean indicates better performance. A tighter interquartile range reflects more concentrated scores, while a wider range indicates greater variability. Outliers correspond to artworks whose scores deviate substantially from the overall distribution, suggesting that the models strongly or weakly recall a small subset of artworks with extreme confidence levels.}
    \label{fig: imitation_box_plot}
\end{figure*}

\begin{figure*}[tp]
  \centering
  \begin{minipage}[t]{0.48\textwidth}
    \centering
    \input{figures/battle_fig.tex}
  \end{minipage}\hfill
  \begin{minipage}[t]{0.48\textwidth}
    \centering
    \input{figures/battle/SD_sem}
  \end{minipage}
  \par\vspace{8pt}
  \begin{minipage}[t]{0.48\textwidth}
    \centering
    \input{figures/battle/SD_aes}
  \end{minipage}\hfill
  \begin{minipage}[t]{0.48\textwidth}
    \centering
    \input{figures/battle/SDXL_sem}
  \end{minipage}
  \par\vspace{8pt}
  \begin{minipage}[t]{0.48\textwidth}
    \centering
    \input{figures/battle/SDXL_aes}
  \end{minipage}\hfill
  \begin{minipage}[t]{0.48\textwidth}
    \centering
    \input{figures/battle/SANA_sem}
  \end{minipage}
\end{figure*}
\begin{figure*}[tp]
  \centering
  \begin{minipage}[t]{0.48\textwidth}
    \centering
    \input{figures/battle/SANA_fid}
  \end{minipage}\hfill
  \begin{minipage}[t]{0.48\textwidth}
    \centering
    \input{figures/battle/SANA_aes}
  \end{minipage}
\end{figure*}

\captionsetup[subfigure]{position=top, justification=centering, font=small}

\begin{figure*}[t]
  \centering
  \begin{subfigure}[t]{0.15\textwidth}
    \centering
    \begingroup
      \setlength{\fboxsep}{0pt}%
      \setlength{\fboxrule}{1pt}%
      \includegraphics[width=\linewidth]{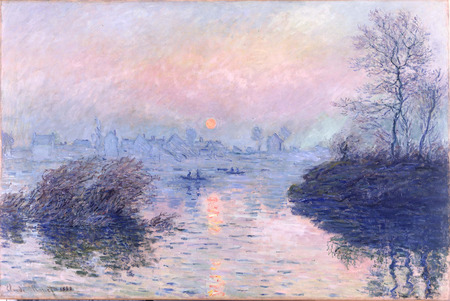}%
    \endgroup
    \caption{Challenger Artwork}
  \end{subfigure}\hfill
  \begin{subfigure}[t]{0.15\textwidth}
    \centering
    \includegraphics[width=\linewidth]{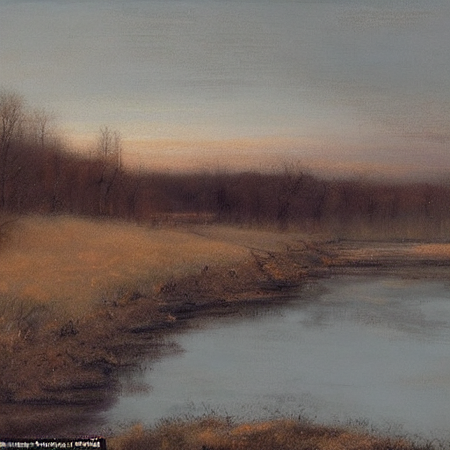}
    \caption{Without defender template (DT)}
  \end{subfigure}\hfill
  \begin{subfigure}[t]{0.15\textwidth}
    \centering
    \includegraphics[width=\linewidth]{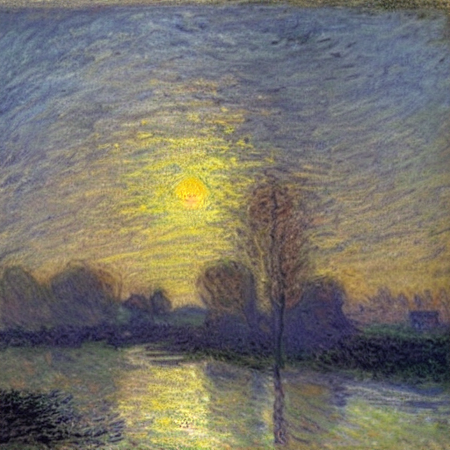}
    \caption{DT: The Three Trees, Autumn by Claude Monet}
  \end{subfigure}\hfill
  \begin{subfigure}[t]{0.15\textwidth}
    \centering
    \includegraphics[width=\linewidth]{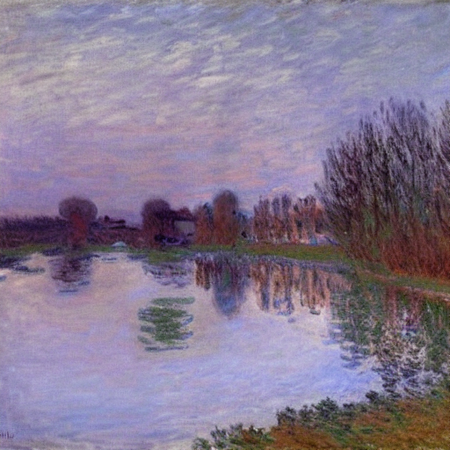}
    \caption{DT: Autumn Effect at Argenteuil by Claude Monet}
  \end{subfigure}\hfill
  \begin{subfigure}[t]{0.15\textwidth}
    \centering
    \includegraphics[width=\linewidth]{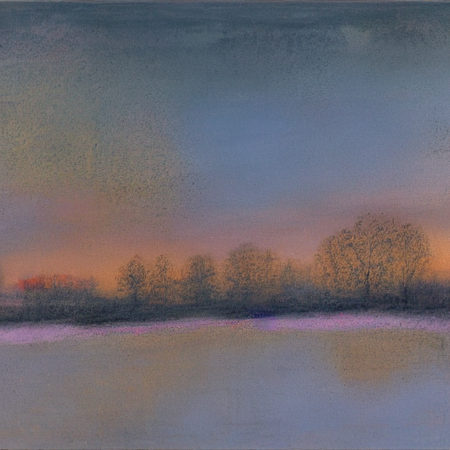}
    \caption{DT: Number 3 by Jackson Pollock}
  \end{subfigure}\hfill
  \begin{subfigure}[t]{0.15\textwidth}
    \centering
    \includegraphics[width=\linewidth]{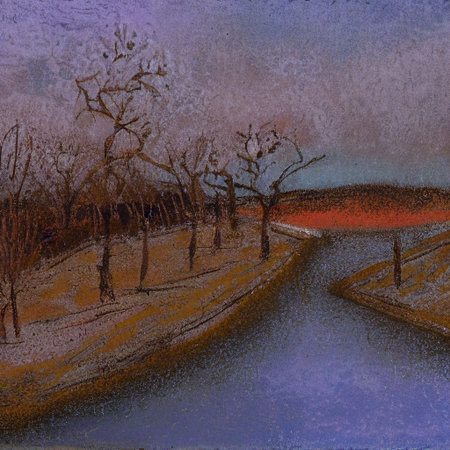}
    \caption{DT: Number 48 by Jackson Pollock}
  \end{subfigure}

\caption{Representative Motif Duel instance for \textbf{SD v1.5} under semantics-based proximity, with a challenger artwork tested against four distinct defenders. The challenger artwork is \textit{Sunset on the Seine at Lavacourt, Winter Effect} by Claude Monet and the motif-derived prompt is given by: ``The scene features muted winter sunset light with pastel tones, a river reflecting shimmering light, a hazy softness over structures, and a winter landscape with sparse, bare trees.'' Column (a) shows the challenger artwork. Column (b) shows the image generated using only the motif-derived prompt without any \textbf{defender template (DT)}. Columns (c) to (f) show images generated by combining the motif-derived prompt with different defender templates, where each defender template specifies a particular artwork and artist (refer Algorithm \ref{algo: art_arena}, step 2).}
\label{fig:sd_sem_motif}

\end{figure*}

\captionsetup[subfigure]{position=top, justification=centering, font=small}

\begin{figure*}[t]
  \centering
  \begin{subfigure}[t]{0.15\textwidth}
    \centering
    \begingroup
      \setlength{\fboxsep}{0pt}%
      \setlength{\fboxrule}{1pt}%
      \includegraphics[width=\linewidth]{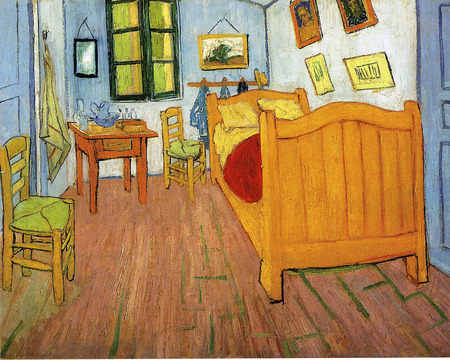}%
    \endgroup
    \caption{Challenger Artwork}
  \end{subfigure}\hfill
  \begin{subfigure}[t]{0.15\textwidth}
    \centering
    \includegraphics[width=\linewidth]{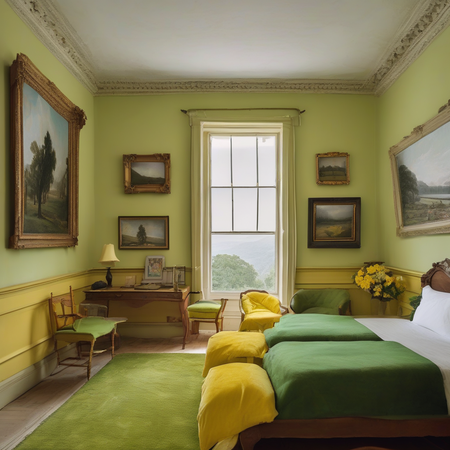}
    \caption{Without defender template (DT)}
  \end{subfigure}\hfill
  \begin{subfigure}[t]{0.15\textwidth}
    \centering
    \includegraphics[width=\linewidth]{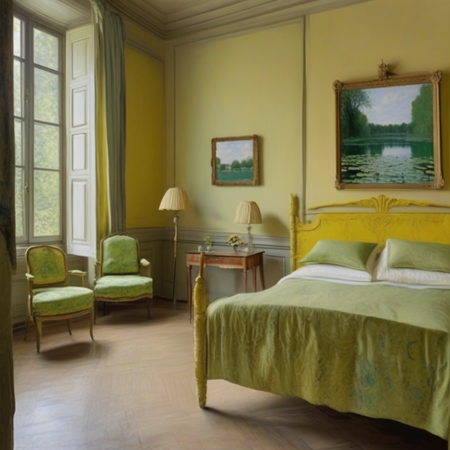}
    \caption{DT: Water Lilies by Claude Monet}
  \end{subfigure}\hfill
  \begin{subfigure}[t]{0.15\textwidth}
    \centering
    \includegraphics[width=\linewidth]{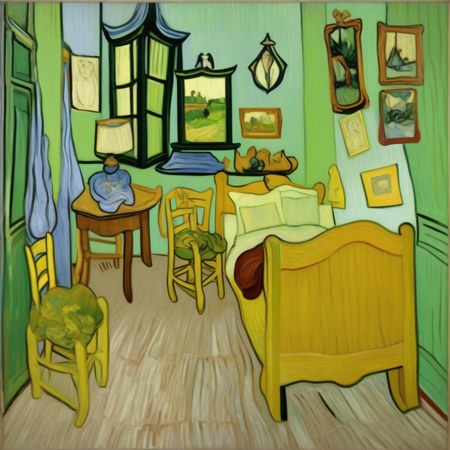}
    \caption{DT: Irises by Vincent van Gogh}
  \end{subfigure}\hfill
  \begin{subfigure}[t]{0.15\textwidth}
    \centering
    \includegraphics[width=\linewidth]{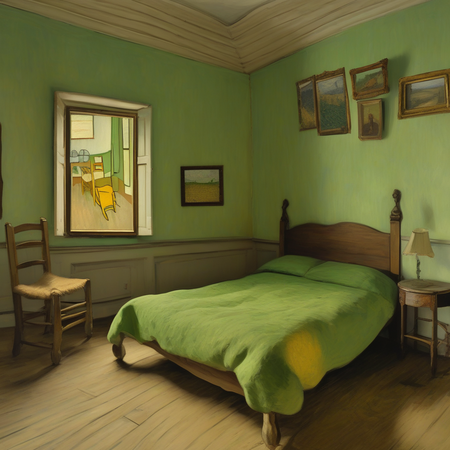}
    \caption{DT: Self Portrait with Felt Hat by Vincent van Gogh}
  \end{subfigure}\hfill
  \begin{subfigure}[t]{0.15\textwidth}
    \centering
    \includegraphics[width=\linewidth]{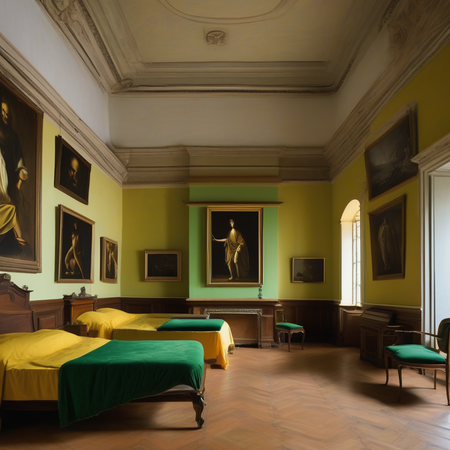}
    \caption{DT: Incredulity of Saint Thomas by Caravaggio}
  \end{subfigure}
\caption{Representative Motif Duel instance for \textbf{SDXL} under semantics based proximity, with a challenger artwork tested against four distinct defenders. The challenger artwork is \textbf{Vincent's Bedroom in Arles} by Vincent van Gogh and the motif-derived prompt is given by: ``A room with a prominent yellow bed, plain wooden chairs with green cushions, and framed pictures on the walls featuring portraits and landscapes.'' Column (a) shows the challenger artwork. Column (b) shows the image generated using only the motif-derived prompt without any defender template (DT). Columns (c) to (f) show images generated by combining the motif-derived prompt with different defender templates, where each defender template specifies a particular artwork and artist.}
\label{fig:sdxl_sem_motif}
\end{figure*}

\captionsetup[subfigure]{position=top, justification=centering, font=small}

\begin{figure*}[t]
  \centering
  \begin{subfigure}[t]{0.15\textwidth}
    \centering
    \begingroup
      \setlength{\fboxsep}{0pt}%
      \setlength{\fboxrule}{1pt}%
      \includegraphics[width=\linewidth]{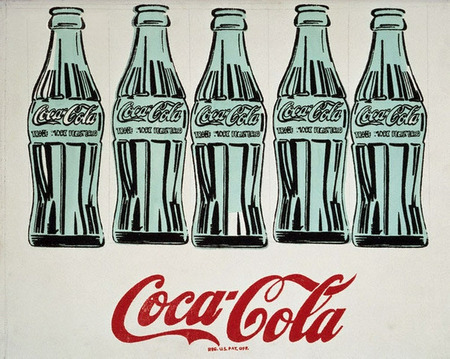}%
    \endgroup
    \caption{Challenger Artwork}
  \end{subfigure}\hfill
  \begin{subfigure}[t]{0.15\textwidth}
    \centering
    \includegraphics[width=\linewidth]{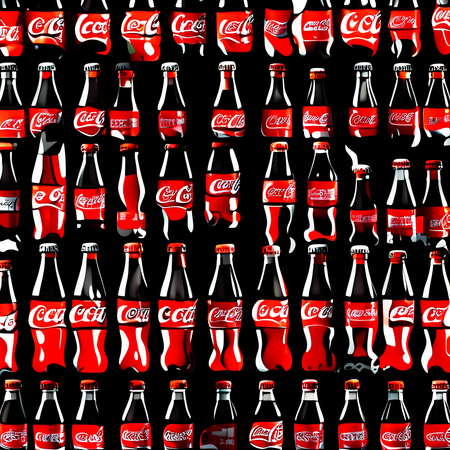}
    \caption{Without defender template (DT)}
  \end{subfigure}\hfill
  \begin{subfigure}[t]{0.15\textwidth}
    \centering
    \includegraphics[width=\linewidth]{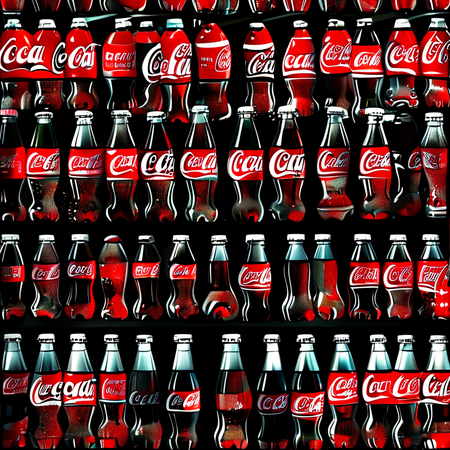}
    \caption{DT: Christ on the Cross by Rembrandt}
  \end{subfigure}\hfill
  \begin{subfigure}[t]{0.15\textwidth}
    \centering
    \includegraphics[width=\linewidth]{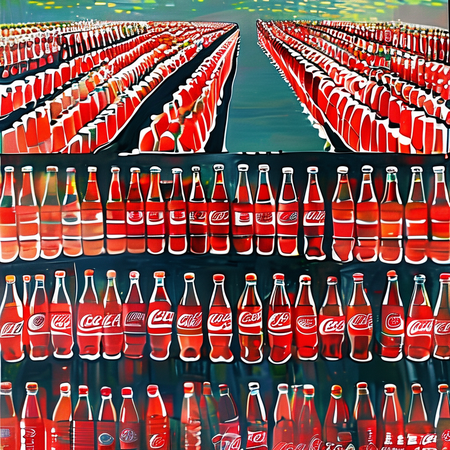}
    \caption{DT: The Japanese Bridge by Claude Monet}
  \end{subfigure}\hfill
  \begin{subfigure}[t]{0.15\textwidth}
    \centering
    \includegraphics[width=\linewidth]{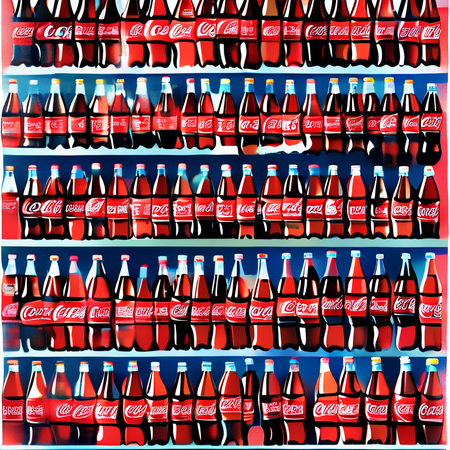}
    \caption{DT: Marilyn Blue by Andy Warhol}
  \end{subfigure}\hfill
  \begin{subfigure}[t]{0.15\textwidth}
    \centering
    \includegraphics[width=\linewidth]{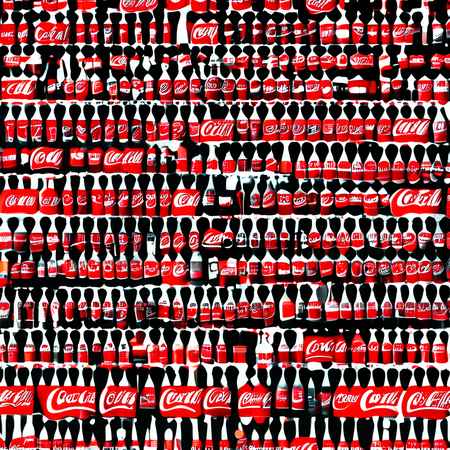}
    \caption{DT: Mickey by Andy Warhol}
  \end{subfigure}
  \caption{Representative Motif Duel instance for \textbf{SANA-1.5} under semantics-based proximity, with a challenger artwork tested against four distinct defenders. The challenger artwork is \textbf{Green Coca Cola Bottles} by Andy Warhol and the motif-derived prompt is given by: ``Rows of repeating Coca-Cola bottles arranged in a flat layout, emphasizing a lack of visual depth or dimensional perspective.'' Column (a) shows the challenger artwork. Column (b) shows the image generated using only the motif-derived prompt without any defender template (DT). Columns (c) to (f) show images generated by combining the motif-derived prompt with different defender templates, where each defender template specifies a particular artwork and artist.}
\label{fig:sana_sem_motif}
\end{figure*}

\begin{figure*}[t]
  \centering

  \begin{subfigure}[t]{0.32\textwidth}
    \centering
    \includegraphics[width=\linewidth]{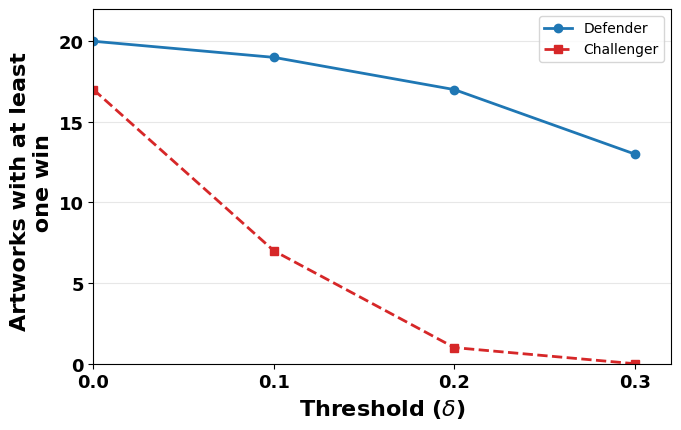} 
    \caption{Semantics}
  \end{subfigure}
  \hfill
  \begin{subfigure}[t]{0.32\textwidth}
    \centering
    \includegraphics[width=\linewidth]{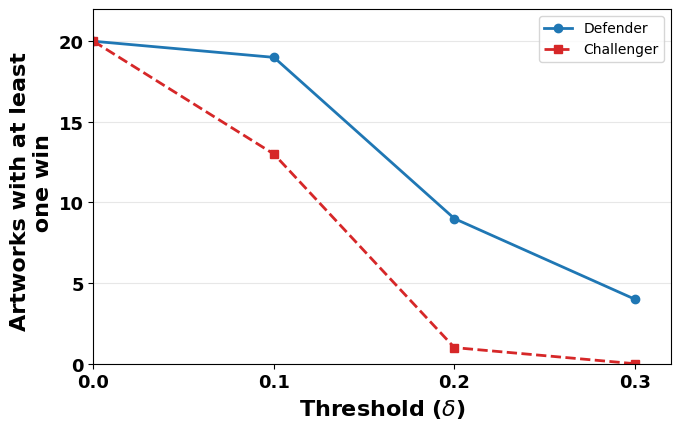}
    \caption{Aesthetics}
  \end{subfigure}
  \hfill
  \begin{subfigure}[t]{0.32\textwidth}
    \centering
    \includegraphics[width=\linewidth]{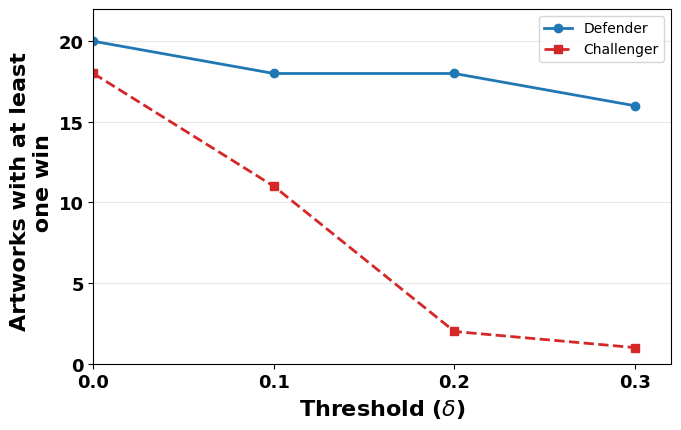}
    \caption{Fidelity}
  \end{subfigure}

  \caption{\textbf{Sensitivity analysis of threshold} ($\delta$) used to award rounds for \textbf{SD v1.5} with proximity (a) Semantics, (b) Aesthetics, and (c) Fidelity. The x-axis represents the different threshold values and the y-axis represent the number of artworks (defender and challenger) with at least one win.}
  \label{fig:SD_delta}
\end{figure*}

\begin{figure*}[t]
  \centering

  \begin{subfigure}[t]{0.32\textwidth}
    \centering
    \includegraphics[width=\linewidth]{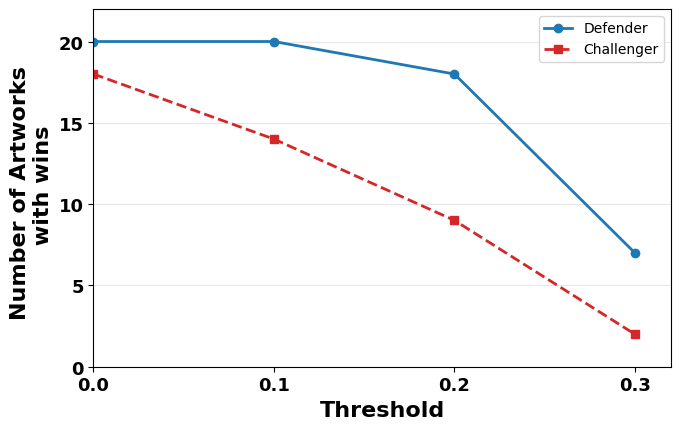} 
    \caption{Semantics}
  \end{subfigure}
  \hfill
  \begin{subfigure}[t]{0.32\textwidth}
    \centering
    \includegraphics[width=\linewidth]{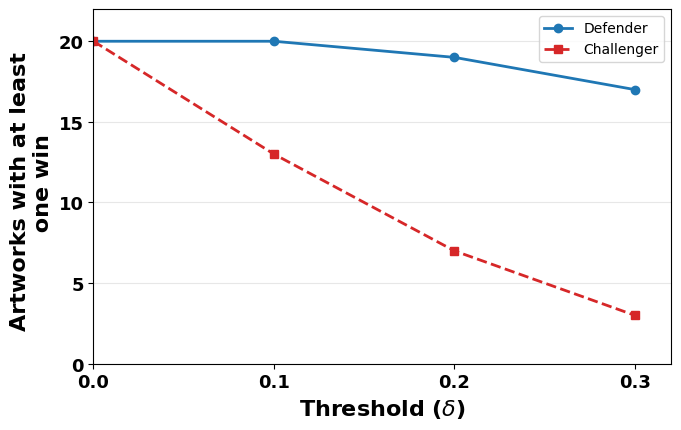}
    \caption{Aesthetics}
  \end{subfigure}
  \hfill
  \begin{subfigure}[t]{0.32\textwidth}
    \centering
    \includegraphics[width=\linewidth]{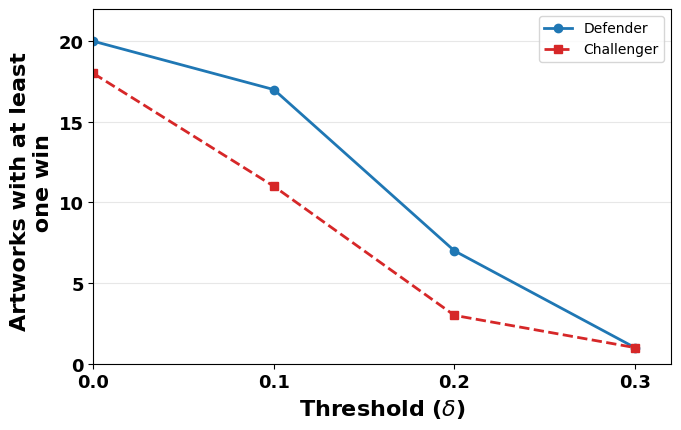}
    \caption{Fidelity}
  \end{subfigure}

    \caption{\textbf{Sensitivity analysis of threshold} ($\delta$) used to award rounds for \textbf{SDXL} with proximity (a) Semantics, (b) Aesthetics, and (c) Fidelity. The x-axis represents the different threshold values and the y-axis represent the number of artworks (defender and challenger) with at least one win.}
  \label{fig:SDXL_delta}
\end{figure*}
\begin{figure*}[t]
  \centering

  \begin{subfigure}[t]{0.32\textwidth}
    \centering
    \includegraphics[width=\linewidth]{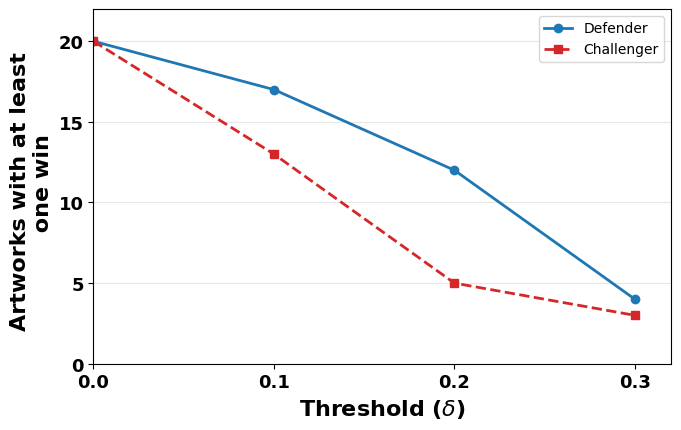} 
    \caption{Semantics}
  \end{subfigure}
  \hfill
  \begin{subfigure}[t]{0.32\textwidth}
    \centering
    \includegraphics[width=\linewidth]{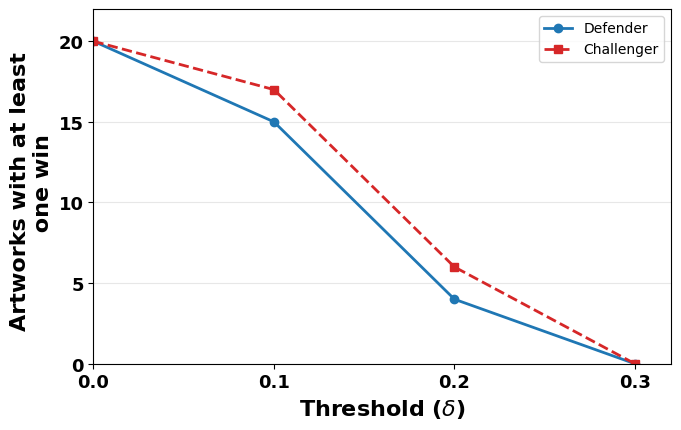}
    \caption{Aesthetics}
  \end{subfigure}
  \hfill
  \begin{subfigure}[t]{0.32\textwidth}
    \centering
    \includegraphics[width=\linewidth]{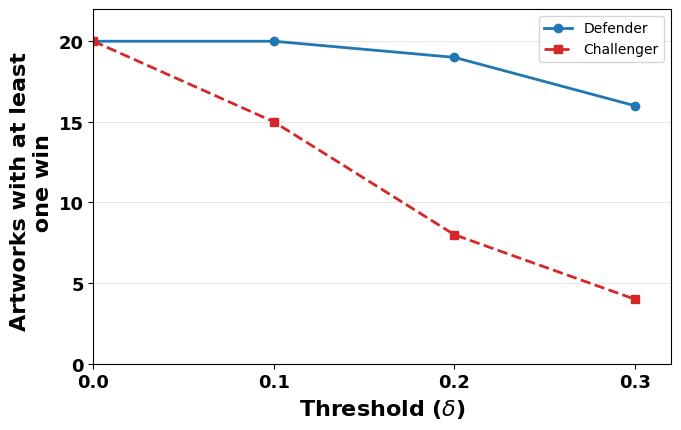}
    \caption{Fidelity}
  \end{subfigure}

    \caption{\textbf{Sensitivity analysis of threshold} ($\delta$) used to award rounds for \textbf{SANA-1.5} with proximity (a) Semantics, (b) Aesthetics, and (c) Fidelity. The x-axis represents the different threshold values and the y-axis represent the number of artworks (defender and challenger) with at least one win.}
  \label{fig:SANA_delta}
\end{figure*}

\begin{table*}[t]
  \begin{minipage}[t]{0.485\textwidth}
\centering
\scriptsize
\setlength{\tabcolsep}{3pt}
\renewcommand{\arraystretch}{0.9}
\setlength{\aboverulesep}{0.25ex}
\setlength{\belowrulesep}{0.25ex}

\caption{Influence Ledgers from Motif Duels for \textbf{SD v1.5}, for \textbf{Semantics}, \textbf{Aesthetics}, and \textbf{Fidelity} (top to bottom). Ranks are assigned by total match wins, with challenger wins used as a tie-breaker.}
\label{tab:sd15_ledgers}

\begin{tabularx}{\columnwidth}{@{}r>{\raggedright\arraybackslash}Xcc@{}}
\toprule
\textbf{Rank} & \textbf{Artwork (base model)} & \textbf{Chall.} & \textbf{Def.} \\
\midrule
\multicolumn{4}{@{}c}{\textit{Semantics (CLIP)}} \\ \hline
\addlinespace[1pt]
1  & Leonardo da Vinci, \textit{Madonna Litta (Madonna and the Child)} & 16 & 13 \\
2  & Vincent van Gogh, \textit{Wheat Fields at Auvers Under Clouded Sky} & 13 & 15 \\
3  & Claude Monet, \textit{Pool with Waterlilies} & 13 & 15 \\
16 & Claude Monet, \textit{Yachts at Argenteuil} & 4 & 7 \\
17 & Jackson Pollock, \textit{Number 3} & 1 & 10 \\
18 & Claude Monet, \textit{Small Branch of the Seine} & 4 & 6 \\
19 & Jackson Pollock, \textit{Number 48} & 1 & 8 \\
20 & Raphael, \textit{The Fall on the Road to Calvary} & 0 & 9 \\
\midrule
\multicolumn{4}{@{}c}{\textit{Aesthetics (LPIPS)}} \\ \hline
\addlinespace[1pt]
1  & Jackson Pollock, \textit{Circumcision January} & 17 & 15 \\
2  & Jackson Pollock, \textit{Number 17} & 17 & 15 \\
3  & Andy Warhol, \textit{Spam} & 19 & 12 \\
16 & Claude Monet, \textit{Lavacourt, Sun and Snow} & 7 & 5 \\
17 & Jackson Pollock, \textit{Enchanted Forest} & 8 & 3 \\
18 & Claude Monet, \textit{Houses of Parliament, Fog Effect} & 8 & 2 \\
19 & Jackson Pollock, \textit{Number 4} & 6 & 4 \\
20 & Andy Warhol, \textit{Rorschach} & 2 & 0 \\
\midrule
\multicolumn{4}{@{}c}{\textit{Fidelity (CSD)}} \\ \hline
\addlinespace[1pt]
1  & Vincent van Gogh, \textit{Olive Grove} & 19 & 12 \\
2  & Vincent van Gogh, \textit{A Group of Cottages} & 14 & 12 \\
3  & Vincent van Gogh, \textit{Wheat Field at Auvers with White House} & 14 & 12 \\
16 & Vincent van Gogh, \textit{Daubigny's Garden} & 7 & 8 \\
17 & Claude Monet, \textit{Waterloo Bridge, London} & 3 & 12 \\
18 & Pablo Picasso, \textit{Weeping Woman} & 0 & 13 \\
19 & Claude Monet, \textit{The Sea Seen from the Cliffs of Fecamp} & 0 & 12 \\
20 & Vincent van Gogh, \textit{Flowering Garden} & 0 & 7 \\
\bottomrule
\end{tabularx}


  \end{minipage}\hfill
  \begin{minipage}[t]{0.485\textwidth}
\centering
\scriptsize
\setlength{\tabcolsep}{3pt}
\renewcommand{\arraystretch}{0.9}
\setlength{\aboverulesep}{0.25ex}
\setlength{\belowrulesep}{0.25ex}

\caption{Influence Ledgers from Motif Duels for \textbf{SANA-1.5}, for \textbf{Semantics}, \textbf{Aesthetics}, and \textbf{Fidelity} (top to bottom). Ranks are assigned by total match wins, with challenger wins used as a tie-breaker.}
\label{tab:Sana_clip_tr}

\begin{tabularx}{\columnwidth}{@{}r>{\raggedright\arraybackslash}Xcc@{}}
\toprule
\textbf{Rank} & \textbf{Artwork (base model)} & \textbf{Chall.} & \textbf{Def.} \\
\midrule
\multicolumn{4}{@{}c}{\textit{Semantics (CLIP)}} \\ \hline
\addlinespace[1pt]
1  & Andy Warhol, \textit{Green Coca Cola Bottles} & 19 & 16 \\
2  & Rembrandt, \textit{Christ on the Cross} & 19 & 12 \\
3  & Claude Monet, \textit{The Japanese Bridge} & 14 & 16 \\
16 & Andy Warhol, \textit{Marilyn Monroe} & 5 & 9 \\
17 & Claude Monet, \textit{A Corner of the Garden at Montgeron} & 5 & 8 \\
18 & Andy Warhol, \textit{After Marilyn Pink} & 3 & 9 \\
19 & Andy Warhol, \textit{Marilyn Blue} & 4 & 5 \\
20 & Andy Warhol, \textit{Mickey} & 1 & 1 \\
\midrule
\multicolumn{4}{@{}c}{\textit{Aesthetics (LPIPS)}} \\ \hline
\addlinespace[1pt]
1  & Jackson Pollock, \textit{Echo (Number 25)} & 19 & 15 \\
2  & Jackson Pollock, \textit{Black \& White (Number 20)} & 17 & 15 \\
3  & Jackson Pollock, \textit{Number 32} & 17 & 13 \\
16 & Rembrandt, \textit{Lighting Study of an Elderly Woman in a White Cap} & 4 & 8 \\
17 & Andy Warhol, \textit{Crushed Campbell's Soup Can (Beef Noodle)} & 7 & 3 \\
18 & Rembrandt, \textit{Portrait of a Bearded Man in Black Beret} & 4 & 6 \\
19 & Rembrandt, \textit{Portrait of a Woman Wearing a Gold Chain} & 6 & 2 \\
20 & Rembrandt, \textit{Portrait of a seated man rising from his chair} & 6 & 1 \\
\midrule
\multicolumn{4}{@{}c}{\textit{Fidelity (CSD)}} \\ \hline
\addlinespace[1pt]
1  & Jackson Pollock, \textit{Convergence (Number 10)} & 19 & 13 \\
2  & Andy Warhol, \textit{Marilyn Monroe} & 17 & 13 \\
3  & Andy Warhol, \textit{Cross} & 19 & 9 \\
16 & Vincent van Gogh, \textit{Trees in the garden of the Hospital Saint-Paul} & 7 & 6 \\
17 & Vincent van Gogh, \textit{Daubigny's Garden} & 4 & 8 \\
18 & Andy Warhol, \textit{Untitled (From Marilyn Monroe)} & 5 & 6 \\
19 & Vincent van Gogh, \textit{Sunny Lawn in a Public Park} & 2 & 9 \\
20 & Vincent van Gogh, \textit{Mademoiselle Gachet in her garden at Auvers-sur-Oise} & 1 & 10 \\
\bottomrule
\end{tabularx}


  \end{minipage}
\end{table*}




\begin{table*}[t]
\centering
\small
\caption{Artworks in the FitSet for \textbf{SD v1.5} ranked for proximity: Semantics (CLIP) fitness score (mean $\pm$ variance, with variance computed over $K = 5$ runs of the Fitness test).}
\label{tab:SD_CLIP_fitness}
\begin{tabular}{r p{0.5\textwidth} p{0.3\textwidth}}
\toprule
Rank & Artwork Name & Fitness Score (mean $\pm$ variance) \\ \midrule
1  & Crucifixion by Raphael & $0.93750 \pm 6.414 \times 10^{-5}$ \\
2  & Number 48 by Jackson Pollock & $0.93732 \pm 4.284 \times 10^{-4}$ \\
3  & Pool with Waterlilies by Claude Monet & $0.93701 \pm 2.576 \times 10^{-4}$ \\
4  & Vase with Carnations and Other Flowers by Vincent van Gogh & $0.93164 \pm 3.725 \times 10^{-4}$ \\
5  & Yachts At Argenteuil by Claude Monet & $0.92920 \pm 1.324 \times 10^{-4}$ \\
6  & Water Lilies Red by Claude Monet & $0.92871 \pm 6.565 \times 10^{-5}$ \\
7  & The Three Trees, Autumn by Claude Monet & $0.92822 \pm 8.951 \times 10^{-5}$ \\
8  & Autumn Effect at Argenteuil by Claude Monet & $0.92725 \pm 1.210 \times 10^{-4}$ \\
9  & Madonna Litta (Madonna and the Child) by Leonardo da Vinci & $0.92676 \pm 5.468 \times 10^{-5}$ \\
10 & Olive Grove - Orange Sky by Vincent van Gogh & $0.92676 \pm 2.997 \times 10^{-4}$ \\
11 & Poppy Field in Giverny by Claude Monet & $0.92529 \pm 7.819 \times 10^{-4}$ \\
12 & Sunset on the Seine at Lavacourt, Winter Effect by Claude Monet & $0.92529 \pm 8.303 \times 10^{-4}$ \\
13 & Number 3 by Jackson Pollock & $0.92480 \pm 4.921 \times 10^{-5}$ \\
14 & Small Branch of the Seine by Claude Monet & $0.92334 \pm 2.392 \times 10^{-4}$ \\
15 & Wheat Fields at Auvers Under Clouded Sky by Vincent van Gogh & $0.92334 \pm 2.246 \times 10^{-4}$ \\
16 & The Fall on the Road to Calvary by Raphael & $0.92285 \pm 3.188 \times 10^{-4}$ \\
17 & Nude woman with turkish bonnet by Pablo Picasso & $0.92285 \pm 2.611 \times 10^{-3}$ \\
18 & Wheat Field with the Alpilles Foothills in the Background by Vincent van Gogh & $0.92139 \pm 7.965 \times 10^{-4}$ \\
19 & Vase of Peonies on a Small Pedestal by Edouard Manet & $0.92041 \pm 2.317 \times 10^{-4}$ \\
20 & Waves Breaking by Claude Monet & $0.92041 \pm 2.291 \times 10^{-4}$ \\
\bottomrule
\end{tabular}
\end{table*}

\begin{table*}[t]
\centering
\small
\caption{Artworks in the FitSet for \textbf{SD v1.5} ranked for proximity: Aesthetics (LPIPS) fitness score (mean $\pm$ variance, with variance computed over $K = 5$ runs of the Fitness test).}
\label{tab:SD_LPIPS_fitness}
\begin{tabular}{r p{0.5\textwidth} p{0.3\textwidth}}
\toprule
Rank & Artwork Name & Fitness Score (mean $\pm$ variance) \\ \midrule
1  & Circumcision January by Jackson Pollock & $0.54193 \pm 1.386 \times 10^{-4}$ \\
2  & Enchanted Forest by Jackson Pollock & $0.54782 \pm 2.029 \times 10^{-3}$ \\
3  & Number 29 by Jackson Pollock & $0.55431 \pm 1.256 \times 10^{-3}$ \\
4  & Ocean Greyness by Jackson Pollock & $0.55560 \pm 1.621 \times 10^{-2}$ \\
5  & Number 48 by Jackson Pollock & $0.55670 \pm 4.698 \times 10^{-4}$ \\
6  & Untitled (Green Silver) by Jackson Pollock & $0.56490 \pm 1.604 \times 10^{-4}$ \\
7  & Mural by Jackson Pollock & $0.56646 \pm 3.354 \times 10^{-4}$ \\
8  & Houses of Parliament, Fog Effect by Claude Monet & $0.56746 \pm 4.114 \times 10^{-3}$ \\
9  & Spam by Andy Warhol & $0.58048 \pm 1.135 \times 10^{-3}$ \\
10 & Mask by Jackson Pollock & $0.58515 \pm 7.405 \times 10^{-4}$ \\
11 & Sibyl Erithraea (Sistine Chapel Ceiling) by Michelangelo & $\text{0.58612} \pm \text{2.385} \times 10^{-\text{-3}}$ \\
12 & The Beach at Sainte-Adresse by Claude Monet & $0.58741 \pm 1.435 \times 10^{-3}$ \\
13 & Number 17 by Jackson Pollock & $0.58817 \pm 9.817 \times 10^{-4}$ \\
14 & Chestnut Tree in Blossom by Vincent van Gogh (1890) & $0.59207 \pm 2.743 \times 10^{-4}$ \\
15 & Birth by Jackson Pollock & $0.59270 \pm 1.705 \times 10^{-4}$ \\
16 & Number 4 by Jackson Pollock & $0.59378 \pm 6.409 \times 10^{-5}$ \\
17 & Rorschach by Andy Warhol & $0.59394 \pm 1.727 \times 10^{-3}$ \\
18 & Lavacourt, Sun and Snow by Claude Monet & $0.59395 \pm 1.480 \times 10^{-3}$ \\
19 & Trees with Ivy by Vincent van Gogh & $0.59746 \pm 1.589 \times 10^{-3}$ \\
20 & Chestnut Trees in Blossom by Vincent van Gogh (1887) & $0.59793 \pm 1.630 \times 10^{-4}$ \\
\bottomrule
\end{tabular}
\end{table*}

\begin{table*}[t]
\centering
\small
\caption{Artworks in the FitSet for \textbf{SD v1.5} ranked for proximity: Fidelity (CSD) fitness score (mean $\pm$ variance, with variance computed over $K = 5$ runs of the Fitness test).}
\label{tab:SD_CSD_fitness}
\begin{tabular}{r p{0.5\textwidth} p{0.3\textwidth}}
\toprule
Rank & Artwork Name & Fitness Score (mean $\pm$ variance) \\ \midrule
1  & Olive Grove by Vincent van Gogh & $0.89784 \pm 8.083 \times 10^{-4}$ \\
2  & Le Mas de Saint-Paul (A Meadow in the Mountains) by Vincent Van Gogh & $0.87941 \pm 1.455 \times 10^{-3}$ \\
3  & Poppy Field in Giverny by Claude Monet & $0.86426 \pm 2.024 \times 10^{-4}$ \\
4  & The Church at Auvers by Vincent van Gogh & $0.86373 \pm 1.369 \times 10^{-3}$ \\
5  & The Garden of St. Paul's Hospital at St. Remy by Vincent van Gogh & $0.85926 \pm 1.394 \times 10^{-3}$ \\
6  & Public Garden with Couple and Blue Fir Tree (The Poet's Garden III) by Vincent van Gogh & $0.85904 \pm 4.480 \times 10^{-4}$ \\
7  & Wheat Field at Auvers with White House by Vincent van Gogh & $0.85795 \pm 1.268 \times 10^{-3}$ \\
8  & Waterloo Bridge, London by Claude Monet & $0.85207 \pm 3.097 \times 10^{-3}$ \\
9  & Flower Garden by Gustav Klimt & $0.84759 \pm 2.146 \times 10^{-3}$ \\
10  & Entrance to the Public Garden in Arles by Vincent van Gogh & $0.84324 \pm 5.746 \times 10^{-4}$ \\
11 & The Sea Seen from the Cliffs of Fecamp by Claude Monet & $0.84285 \pm 5.397 \times 10^{-4}$ \\
12 & Entrance to a Quarry near Saint Remy by Vincent van Gogh & $0.84129 \pm 3.083 \times 10^{-4}$ \\
13 & Flowers on the Banks of Seine near Vetheuil by Claude Monet & $0.84059 \pm 7.077 \times 10^{-4}$ \\
14 & A Group of Cottages by Vincent van Gogh & $0.83874 \pm 8.440 \times 10^{-4}$ \\
15 & Weeping Woman by Pablo Picasso & $0.83824 \pm 1.998 \times 10^{-3}$ \\
16 & Avenue in the Park by Vincent van Gogh & $0.83733 \pm 1.117 \times 10^{-3}$ \\
17 & Daubigny's Garden by Vincent van Gogh & $0.83620 \pm 7.955 \times 10^{-4}$ \\
18 & Camille Monet and a Child in the Artist's Garden in Argenteuil by Claude Monet & $0.83496 \pm 1.164 \times 10^{-3}$ \\
19 & Green Wheat Fields by Vincent van Gogh & $0.83110 \pm 1.661 \times 10^{-3}$ \\
20 & Flowering Garden by Vincent van Gogh & $0.82969 \pm 1.308 \times 10^{-3}$ \\
\bottomrule
\end{tabular}
\end{table*}

\begin{table*}[t]
\centering
\small
\caption{Artworks in the FitSet for \textbf{SDXL} ranked for proximity: Semantics (CLIP) fitness score (mean $\pm$ variance, with variance computed over $K = 5$ runs of the Fitness test).}
\label{tab:SDXL_CLIP_fitness}
\begin{tabular}{r p{0.5\textwidth} p{0.3\textwidth}}
\toprule
Rank & Artwork Name & Fitness Score (mean $\pm$ variance) \\ \midrule
1  & Blumengarten by Gustav Klimt & $0.94141 \pm 6.135 \times 10^{-4}$ \\
2  & Vase with Zinnias by Vincent van Gogh & $0.93994 \pm 4.396 \times 10^{-5}$ \\
3  & Marilyn Monroe by Andy Warhol & $0.93555 \pm 2.906 \times 10^{-4}$ \\
4  & Number 48 by Jackson Pollock & $0.93164 \pm 2.564 \times 10^{-3}$ \\
5  & Olive Grove by Vincent van Gogh & $0.93115 \pm 1.273 \times 10^{-4}$ \\
6  & Self Portrait with a Grey Felt Hat by Vincent van Gogh & $0.93018 \pm 6.999 \times 10^{-4}$ \\
7  & Water Lilies by Claude Monet & $0.92822 \pm 4.752 \times 10^{-4}$ \\
8  & Irises by Vincent van Gogh & $0.92822 \pm 4.086 \times 10^{-5}$ \\
9  & Olive Trees with Yellow Sky and Sun by Vincent van Gogh & $0.92676 \pm 1.591 \times 10^{-4}$ \\
10 & After Marilyn Pink by Andy Warhol & $0.92578 \pm 3.346 \times 10^{-4}$ \\
11 & Still Life - Vase with Twelve Sunflowers by Vincent van Gogh & $0.92529 \pm 6.580 \times 10^{-6}$ \\
12 & Self Portrait with Felt Hat by Vincent van Gogh & $0.92432 \pm 4.202 \times 10^{-5}$ \\
13 & David by Michelangelo & $0.92432 \pm 1.012 \times 10^{-3}$ \\
14 & The Fall on the Road to Calvary by Raphael & $0.92139 \pm 9.944 \times 10^{-4}$ \\
15 & Bouquet of Sunflowers by Claude Monet & $0.91797 \pm 1.381 \times 10^{-3}$ \\
16 & Vincent's Bedroom in Arles by Vincent van Gogh & $0.91748 \pm 5.655 \times 10^{-5}$ \\
17 & Incredulity of Saint Thomas by Caravaggio & $0.91699 \pm 5.120 \times 10^{-4}$ \\
18 & Orchard in Bloom by Claude Monet & $0.91650 \pm 6.647 \times 10^{-5}$ \\
19 & Number 4 by Jackson Pollock & $0.91650 \pm 8.009 \times 10^{-4}$ \\
20 & Wheat Field with Reaper and Sun by Vincent van Gogh & $0.91504 \pm 1.238 \times 10^{-4}$ \\
\bottomrule
\end{tabular}
\end{table*}

\begin{table*}[t]
\centering
\small
\caption{Artworks in the FitSet for \textbf{SDXL} ranked for proximity: Aesthetics (LPIPS) fitness score (mean $\pm$ variance, with variance computed over $K = 5$ runs of the Fitness test).}
\label{tab:SDXL_LPIPS_fitness}
\begin{tabular}{r p{0.5\textwidth} p{0.3\textwidth}}
\toprule
Rank & Artwork Name & Fitness Score (mean $\pm$ variance) \\ \midrule
1  & Slightly Open Clam Shell by Georgia O'Keeffe & $0.52429 \pm 7.136 \times 10^{-4}$ \\
2  & Bust of Brutus by Michelangelo & $0.61980 \pm 9.270 \times 10^{-4}$ \\
3  & Sky Above Clouds IV by Georgia O'Keeffe & $0.62005 \pm 2.112 \times 10^{-4}$ \\
4  & White Aphrodisiac Telephone by Salvador Dali & $0.63866 \pm 4.949 \times 10^{-4}$ \\
5  & Shadows on the Sea at Pourville by Claude Monet & $0.64155 \pm 6.627 \times 10^{-4}$ \\
6  & Three female heads with one sleeping by Rembrandt & $0.64307 \pm 1.588 \times 10^{-3}$ \\
7  & Eyes in the Heat by Jackson Pollock & $0.64559 \pm 7.907 \times 10^{-4}$ \\
8  & Wheat Field with Reaper and Sun by Vincent van Gogh & $0.64791 \pm 1.552 \times 10^{-4}$ \\
9  & Irises by Vincent van Gogh & $0.64810 \pm 8.238 \times 10^{-4}$ \\
10 & Impression, sunrise by Claude Monet & $0.64971 \pm 7.145 \times 10^{-4}$ \\
11 & Tree with Ivy in the Asylum Garden by Vincent van Gogh & $0.65076 \pm 1.240 \times 10^{-4}$ \\
12 & Black \& White (Number 20) by Jackson Pollock & $0.65157 \pm 3.207 \times 10^{-4}$ \\
13 & Mademoiselle Gachet in her garden at Auvers-sur-Oise by Vincent van Gogh & $0.65371 \pm 7.231 \times 10^{-5}$ \\
14 & Lady with a Lap Dog by Rembrandt & $0.65434 \pm 3.068 \times 10^{-3}$ \\
15 & The Ploughed Field by Vincent van Gogh & $0.65435 \pm 2.517 \times 10^{-4}$ \\
16 & Ice Floes on the Seine at Bougival by Claude Monet & $0.65498 \pm 6.431 \times 10^{-5}$ \\
17 & Mulberry Tree by Vincent van Gogh & $0.65562 \pm 3.079 \times 10^{-4}$ \\
18 & Tomb of Giuliano de Medici by Michelangelo & $0.65589 \pm 3.106 \times 10^{-4}$ \\
19 & Design for Julius II tomb (first version) by Michelangelo & $0.65971 \pm 9.643 \times 10^{-5}$ \\
20 & The Sea at Saint-Adresse by Claude Monet & $0.65992 \pm 6.843 \times 10^{-5}$ \\
\bottomrule
\end{tabular}
\end{table*}

\begin{table*}[t]
\centering
\small
\caption{Artworks in the FitSet for \textbf{SDXL} ranked for proximity: Fidelity (CSD) fitness score (mean $\pm$ variance, with variance computed over $K = 5$ runs of the Fitness test).}
\label{tab:SDXL_CSD_fitness}
\begin{tabular}{r p{0.5\textwidth} p{0.3\textwidth}}
\toprule
Rank & Artwork Name & Fitness Score (mean $\pm$ variance) \\ \midrule
1  & Still Life - Vase with Twelve Sunflowers by Vincent van Gogh & $0.85814 \pm 5.813 \times 10^{-4}$ \\
2  & Wheat Field with Reaper and Sun by Vincent van Gogh & $0.85079 \pm 1.979 \times 10^{-4}$ \\
3  & View of Vessenots near Auvers by Vincent van Gogh & $0.83458 \pm 3.376 \times 10^{-4}$ \\
4  & Self Portrait with Pallette by Vincent van Gogh & $0.81932 \pm 5.011 \times 10^{-4}$ \\
5  & After Marilyn Pink by Andy Warhol & $0.81885 \pm 6.420 \times 10^{-4}$ \\
6  & Field and Ploughman and Mill by Vincent van Gogh & $0.81820 \pm 3.031 \times 10^{-4}$ \\
7  & Campbell's Soup Cans by Andy Warhol & $0.81511 \pm 2.998 \times 10^{-3}$ \\
8  & The Kiss by Gustav Klimt & $0.80719 \pm 5.261 \times 10^{-3}$ \\
9  & Wheat Field at Auvers with White House by Vincent van Gogh & $0.80699 \pm 1.690 \times 10^{-4}$ \\
10 & Mademoiselle Gachet in her garden at Auvers-sur-Oise by Vincent van Gogh & $0.80657 \pm 1.508 \times 10^{-4}$ \\
11 & Vincent's Bedroom in Arles by Vincent van Gogh & $0.80615 \pm 3.517 \times 10^{-4}$ \\
12 & Starry Night Over the Rhone by Vincent van Gogh & $0.80287 \pm 1.158 \times 10^{-3}$ \\
13 & Marilyn Monroe by Andy Warhol & $0.80145 \pm 6.923 \times 10^{-4}$ \\
14 & Composition (White, Black, Blue and Red on White) by Jackson Pollock & $0.80135 \pm 6.594 \times 10^{-4}$ \\
15 & Wheat Fields at Auvers Under Clouded Sky by Vincent van Gogh & $0.79764 \pm 1.074 \times 10^{-3}$ \\
16 & Untitled (From Marilyn Monroe) by Andy Warhol & $0.79740 \pm 1.134 \times 10^{-3}$ \\
17 & The Garden of Saint-Paul Hospital by Vincent van Gogh & $0.79027 \pm 1.277 \times 10^{-4}$ \\
18 & Number 4 by Jackson Pollock & $0.78989 \pm 3.991 \times 10^{-3}$ \\
19 & Three Marilyns by Andy Warhol & $0.78799 \pm 2.719 \times 10^{-3}$ \\
20 & A Group of Cottages by Vincent van Gogh & $0.78793 \pm 1.591 \times 10^{-4}$ \\
\bottomrule
\end{tabular}
\end{table*}

\begin{table*}[t]
\centering
\small
\caption{Artworks in the FitSet for \textbf{SANA-1.5} ranked for proximity: Semantics (CLIP) fitness score (mean $\pm$ variance, with variance computed over $K = 5$ runs of the Fitness test).}
\label{tab:SANA_CLIP_fitness}
\begin{tabular}{r p{0.5\textwidth} p{0.3\textwidth}}
\toprule
Rank & Artwork Name & Fitness Score (mean $\pm$ variance) \\ \midrule
1  & Christ on the Cross by Rembrandt & $0.93262 \pm 2.609 \times 10^{-4}$ \\
2  & The Alpilles with Olive Trees in the Foreground by Vincent van Gogh & $0.92822 \pm 1.338 \times 10^{-3}$ \\
3  & After Marilyn Pink by Andy Warhol & $0.92480 \pm 7.807 \times 10^{-5}$ \\
4  & Pathway in Monet's Garden at Giverny by Claude Monet & $0.92334 \pm 3.405 \times 10^{-4}$ \\
5  & Marilyn Monroe by Andy Warhol & $0.92090 \pm 6.866 \times 10^{-5}$ \\
6  & The Ascension Of Christ by Rembrandt & $0.91699 \pm 9.685 \times 10^{-4}$ \\
7  & Coca-Cola (3) by Andy Warhol & $0.91113 \pm 1.157 \times 10^{-4}$ \\
8  & Green Coca Cola Bottles by Andy Warhol & $0.90918 \pm 3.165 \times 10^{-4}$ \\
9  & Galaxy by Jackson Pollock & $0.90283 \pm 2.724 \times 10^{-4}$ \\
10 & The Japanese Bridge (The Bridge in Monet's Garden) by Claude Monet & $0.90234 \pm 2.754 \times 10^{-3}$ \\
11 & Water Lilies by Claude Monet & $0.89990 \pm 4.820 \times 10^{-5}$ \\
12 & Untitled (From Marilyn Monroe) in pink by Andy Warhol & $0.89648 \pm 4.367 \times 10^{-4}$ \\
13 & Mickey by Andy Warhol & $0.89648 \pm 8.627 \times 10^{-5}$ \\
14 & Untitled (From Marilyn Monroe) in blue by Andy Warhol & $0.89648 \pm 2.367 \times 10^{-3}$\\
15 & Marilyn Blue by Andy Warhol & $0.89502 \pm 2.533 \times 10^{-5}$ \\
16 & Pathway in Monet's Garden at Giverny by Claude Monet & $0.89404 \pm 1.586 \times 10^{-4}$ \\
17 & Portrait with Pink and Blue Face by Henri Matisse & $0.89355 \pm 3.602 \times 10^{-4}$ \\
18 & A Corner of the Garden at Montgeron by Claude Monet & $0.89355 \pm 3.363 \times 10^{-4}$ \\
19 & Apple Trees on the Chantemesle Hill by Claude Monet & $0.89307 \pm 1.959 \times 10^{-4}$ \\
20 & Portrait of Simonetta Vespucci (Portrait of a Young Woman) by Sandro Botticelli & $0.89209 \pm 2.892 \times 10^{-4}$ \\
\bottomrule
\end{tabular}
\end{table*}

\begin{table*}[t]
\centering
\small
\caption{Artworks in the FitSet for \textbf{SANA-1.5} ranked for proximity: Aesthetics (LPIPS) fitness score (mean $\pm$ variance, with variance computed over $K = 5$ runs of the Fitness test).}
\label{tab:SANA_LPIPS_fitness}
\begin{tabular}{r p{0.5\textwidth} p{0.3\textwidth}}
\toprule
Rank & Artwork Name & Fitness Score (mean $\pm$ variance) \\ \midrule
1  & Old Man in Prayer by Rembrandt & $0.47799 \pm 5.387 \times 10^{-3}$ \\
2  & Portrait of a Bearded Man in Black Beret by Rembrandt & $0.54239 \pm 3.573 \times 10^{-3}$ \\
3  & Saskia Wearing A Veil by Rembrandt & $0.54508 \pm 1.665 \times 10^{-3}$ \\
4  & Portrait of a Man Wearing a Black Hat by Rembrandt & $0.58443 \pm 1.585 \times 10^{-3}$ \\
5  & Crushed Campbell's Soup Can (Beef Noodle) by Andy Warhol & $0.59945 \pm 8.945 \times 10^{-3}$ \\
6  & Lighting Study of an Elderly Woman in a White Cap by Rembrandt & $0.60581 \pm 2.839 \times 10^{-4}$ \\
7  & Echo (Number 25) by Jackson Pollock & $0.60956 \pm 1.439 \times 10^{-4}$ \\
8  & Portrait of a Man in the Hat Decorated with Pearls by Rembrandt & $0.61198 \pm 1.784 \times 10^{-3}$ \\
9  & Sistine Chapel Ceiling by Michelangelo & $0.61229 \pm 1.895 \times 10^{-3}$ \\
10  & Portrait of a seated man rising from his chair by Rembrandt & $0.61783 \pm 4.755 \times 10^{-4}$ \\
11 & A lute player by Caravaggio & $0.62036 \pm 1.014 \times 10^{-3}$ \\
12 & Portrait of a Man by Rembrandt & $0.62205 \pm 4.717 \times 10^{-3}$ \\
13 & Tree with Ivy in the Asylum Garden by Vincent van Gogh & $0.62627 \pm 1.903 \times 10^{-4}$ \\
14 & Saint Jerome in Meditation by Caravaggio & $0.62755 \pm 1.953 \times 10^{-4}$ \\
15 & Black \& White (Number 20) by Jackson Pollock & $0.63005 \pm 7.120 \times 10^{-5}$ \\
16 & Number 32 by Jackson Pollock & $0.63268 \pm 3.174 \times 10^{-4}$ \\
17 & Portrait of a Woman Wearing a Gold Chain by Rembrandt & $0.63688 \pm 9.493 \times 10^{-4}$ \\
18 & The Garden of Doctor Gachet at Auvers-sur-Oise by Vincent van Gogh & $0.63957 \pm 1.504 \times 10^{-5}$ \\
19 & Daubigny's Garden by Vincent van Gogh & $0.64068 \pm 4.459 \times 10^{-5}$ \\
20 & Cathedral by Jackson Pollock & $0.64179 \pm 2.218 \times 10^{-4}$ \\
\bottomrule
\end{tabular}
\end{table*}

\begin{table*}[t]
\centering
\small
\caption{Artworks in the FitSet for \textbf{SANA-1.5} ranked for proximity: Fidelity (CSD) fitness score (mean $\pm$ variance, with variance computed over $K = 5$ runs of the Fitness test).}
\label{tab:SANA_CSD_fitness}
\begin{tabular}{r p{0.5\textwidth} p{0.3\textwidth}}
\toprule
Rank & Artwork Name & Fitness Score (mean $\pm$ variance) \\ \midrule
1  & After Marilyn Pink by Andy Warhol & $0.83232 \pm 3.064 \times 10^{-4}$ \\
2  & Marilyn Monroe by Andy Warhol & $0.82431 \pm 1.323 \times 10^{-4}$ \\
3  & Tree Trunks in the Asylum Garden by Vincent van Gogh & $0.81055 \pm 3.615 \times 10^{-3}$ \\
4  & Marilyn Red by Andy Warhol & $0.80763 \pm 1.645 \times 10^{-4}$ \\
5  & Mademoiselle Gachet in her garden at Auvers-sur-Oise by Vincent van Gogh & $0.79405 \pm 4.237 \times 10^{-4}$ \\
6  & Three Marilyns by Andy Warhol & $0.76940 \pm 3.382 \times 10^{-4}$ \\
7  & Beatles by Andy Warhol & $0.76661 \pm 5.145 \times 10^{-4}$ \\
8  & Daubigny's Garden by Vincent van Gogh & $0.76505 \pm 4.839 \times 10^{-4}$ \\
9  & Tree Trunks in the Grass by Vincent van Gogh & $0.76380 \pm 8.086 \times 10^{-3}$ \\
10 & Haystacks in Provence by Vincent van Gogh & $0.75971 \pm 4.214 \times 10^{-3}$ \\
11 & Sunny Lawn in a Public Park by Vincent van Gogh & $0.75823 \pm 7.780 \times 10^{-3}$ \\
12 & The Alpilles with Olive Trees in the Foreground by Vincent van Gogh & $0.75520 \pm 1.545 \times 10^{-3}$ \\
13 & Marilyn Blue by Andy Warhol & $0.75383 \pm 8.501 \times 10^{-4}$ \\
14 & The Garden of Doctor Gachet at Auvers-sur-Oise by Vincent van Gogh & $0.75161 \pm 6.120 \times 10^{-3}$ \\
15 & Trees in the garden of the Hospital Saint-Paul by Vincent van Gogh & $0.74900 \pm 5.012 \times 10^{-3}$ \\
16 & Coca-Cola (3) by Andy Warhol & $0.74500 \pm 1.914 \times 10^{-3}$ \\
17 & Convergence (Number 10) by Jackson Pollock & $0.74393 \pm 3.110 \times 10^{-4}$ \\
18 & Orange Prince by Andy Warhol & $0.73172 \pm 3.410 \times 10^{-4}$ \\
19 & Cross by Andy Warhol & $0.73017 \pm 3.109 \times 10^{-3}$ \\
20 & Untitled (From Marilyn Monroe) by Andy Warhol & $0.71003 \pm 9.527 \times 10^{-5}$ \\
\bottomrule
\end{tabular}
\end{table*}

\begin{table*}[t]
\centering
\footnotesize
\setlength{\tabcolsep}{6pt}
\caption{Unique identifiers \texttt{<id>} used to personalize each model on the artworks of the two artists. Each artwork is trained with the prompt \texttt{painting in the style of <id>}. Since each artist has a separate LoRA adapter, the same identifiers are reused for both artists.}
\label{tab:identifiers}
\begin{tabular}{@{}l c c@{}}
\toprule
\textbf{Artwork} & \textbf{\begin{tabular}[c]{@{}c@{}}SD v1.5,\\SDXL\end{tabular}} & \textbf{SANA-1.5} \\
\midrule
\multicolumn{3}{@{}l}{\textbf{Trent Anthony Francis}} \\
\begin{tabular}[t]{@{}l@{}}\textit{Essays on Nature and Growing}\\\textit{(Path IV)}\end{tabular} & \texttt{sks} & \texttt{ljvk} \\
\textit{Meadow May}                             & \texttt{ska} & \texttt{fpzs} \\
\textit{Star Strangled Banner}                  & \texttt{skb} & \texttt{uywg} \\
\textit{The Moon and Sixpence}                  & \texttt{skc} & \texttt{mxrh} \\
\begin{tabular}[t]{@{}l@{}}\textit{The Star Strangled Banner}\\\textit{(Deviations)}\end{tabular} & \texttt{skd} & \texttt{tdqe} \\
\midrule
\multicolumn{3}{@{}l}{\textbf{Gwenn Seemel}} \\
\textit{American Bison}                         & \texttt{sks} & \texttt{ljvk} \\
\textit{Ladybugs on Daisies}                    & \texttt{ska} & \texttt{fpzs} \\
\textit{Sally with Wolves}                      & \texttt{skb} & \texttt{uywg} \\
\textit{Siblings}                               & \texttt{skc} & \texttt{mxrh} \\
\textit{Springtime in Lambertville}             & \texttt{skd} & \texttt{tdqe} \\
\bottomrule
\end{tabular}
\end{table*}

\begin{table*}[t]
  \centering
  \small
  \caption{Configuration of the case study in detail.}
  \label{tab:setup}
  \begin{tabular}{@{}lp{0.66\textwidth}@{}}
    \toprule
    \textbf{Component} & \textbf{Setting} \\
    \midrule
    \multicolumn{2}{@{}l}{\textit{Data and models}} \\
    \quad Canonical collection & 8{,}737 WikiArt artworks by 20 artists \\
    \quad Artists & Trent Anthony Francis; Gwenn Seemel \\
    \quad Non-canonical artworks $\Non$ & 5 per artist \\
    \quad Models $\M$ & SD v1.5, SDXL, SANA-1.5 \\
    \quad Proximity metrics & CLIP, CSD (higher is closer); LPIPS (lower is closer) \\
    \midrule
    \multicolumn{2}{@{}l}{\textit{Canonical Influence Profiling (base model $\M$)}} \\
    \quad Entry Trials & $K = 5$; \texttt{FitSet} of the top 20 artworks per model and metric \\
    \quad Motif Duels & Round-robin over the \texttt{FitSet}; $R = 5$; $K = 1$ \\
    \quad Influence Ledger & $\delta = 0$; sensitivity analysis over $\delta$ \\
    \quad Canonical artworks $\Can$ & Top 5 artworks of each ledger \\
    \midrule
    \multicolumn{2}{@{}l}{\textit{Personalization Leakage Assessment (personalized model $\Mp$)}} \\
    \quad Method & DreamBooth LoRA, one adapter per artist and model; rank 64, $\alpha = 32$ \\
    \quad Identifiers & One unique identifier per artwork (Table~\ref{tab:identifiers}) \\
    \quad Training prompt & \texttt{painting in the style of <id>} \\
    \quad Training & 10{,}000 steps; batch size 1; learning rate $1 \times 10^{-4}$; random seeds \\
    \quad Resolution & $512 \times 512$ (SD v1.5); $1024 \times 1024$ (SDXL, SANA-1.5) \\
    \quad Personalized model $\Mp$ & One checkpoint per model (saved every 500 steps): highest average fitness on $\Non$ across the three metrics \\
    \quad Entry Trials & $K = 3$; fitness prompt \texttt{painting in the style of <id>} \\
    \quad Motif Duels & Challengers $\Can$, defenders $\Non$; defender template \texttt{in the style painting by <id>}; 25 matches; $R = 5$; $K = 1$; random seeds \\
    \quad Margin $\delta$ & 0, 0.05, 0.10, 0.15, 0.20 \\
    \quad Compute & One NVIDIA A100 GPU (80\,GB) \\
    \bottomrule
  \end{tabular}
\end{table*}

\clearpage

\begin{figure*}[p]
  \centering
  \input{Cards/Trent/LC_SD_Trent_clip}
  \vspace{10pt}
  \input{Cards/Gwenn/LC_SD_Gwenn_clip}
\end{figure*}
\begin{figure*}[p]
  \centering
  \input{Cards/Trent/LC_SD_Trent_csd}
  \vspace{10pt}
  \input{Cards/Gwenn/LC_SD_Gwenn_csd}
\end{figure*}
\begin{figure*}[p]
  \centering
  \input{Cards/Trent/LC_SD_Trent_lpips}
  \vspace{10pt}
  \input{Cards/Gwenn/LC_SD_Gwenn_lpips}
\end{figure*}
\begin{figure*}[p]
  \centering
  \input{Cards/Trent/LC_SDXL_Trent_clip}
  \vspace{10pt}
  \input{Cards/Gwenn/LC_SDXL_Gwenn_clip}
\end{figure*}
\begin{figure*}[p]
  \centering
  \input{Cards/Trent/LC_SDXL_Trent_lpips}
  \vspace{10pt}
  \input{Cards/Gwenn/LC_SDXL_Gwenn_lpips}
\end{figure*}
\begin{figure*}[p]
  \centering
  \input{Cards/Trent/LC_SANA_Trent_clip}
  \vspace{10pt}
  \input{Cards/Gwenn/LC_SANA_Gwenn_clip}
\end{figure*}
\begin{figure*}[p]
  \centering
  \input{Cards/Trent/LC_SANA_Trent_csd}
  \vspace{10pt}
  \input{Cards/Gwenn/LC_SANA_Gwenn_csd}
\end{figure*}
\begin{figure*}[p]
  \centering
  \input{Cards/Trent/LC_SANA_Trent_lpips}
  \vspace{10pt}
  \input{Cards/Gwenn/LC_SANA_Gwenn_lpips}
\end{figure*}
\begin{figure*}[t]
  \centering
  \IfFileExists{LeakageCards_Gwenn/SDXL_Gwenn_csd_LeakageCard_v4_Type2_Dumbbell.pdf}{%
    \includegraphics[width=\textwidth]{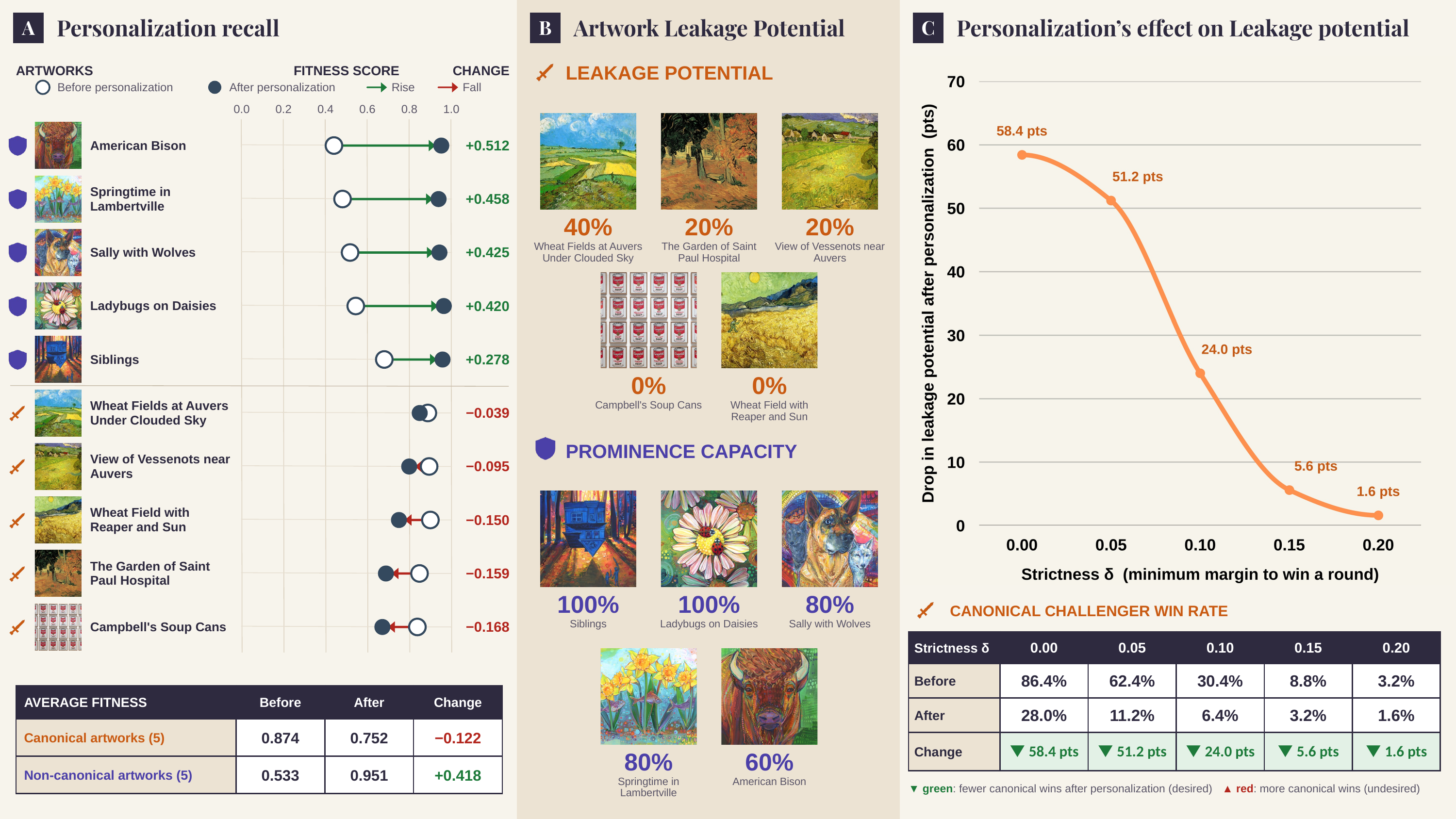}}{%
    \fbox{\parbox[c][6cm][c]{0.95\textwidth}{\centering
      \textit{Placeholder: SDXL leakage card (Gwenn Seemel, CSD)}}}}
  \caption{Leakage card for SDXL personalized using non-canonical artworks by Gwenn Seemel, evaluated for fidelity using CSD.}
  \label{fig:LC_SDXL_Gwenn_CSD}
\end{figure*}


\clearpage
\begin{table}[t]
\centering
\scriptsize
\setlength{\tabcolsep}{2.5pt}
\renewcommand{\arraystretch}{1.0}
\caption{\textbf{Controlled validation of ArtArena under SD v1.5 for Semantics.}
Top: Changes in imitation rank and fitness following strengthening of the bottom-ranked artwork and weakening of the top-ranked artwork from the Influence Ledger.
Bottom: Defender Loss Rate (DLR) before and after each intervention, along with the corresponding changes across comparison margins $\delta$.}
\label{tab:val_sd_clip_imit}
\begin{tabularx}{\columnwidth}{@{}>{\raggedright\arraybackslash}X c c c c c c@{}}
\toprule
\multirow{2}{*}{\textbf{Target artwork}} & \multicolumn{3}{c}{\textbf{Imitation rank}} & \multicolumn{3}{c}{\textbf{Fitness score}} \\
\cmidrule(lr){2-4}\cmidrule(l){5-7}
 & Base & New & Shift & Base & New & Change \\
\midrule
Raphael, \textit{The Fall on the Road to Calvary}\newline(Fine-tuning, bottom-1) & 16 & 6 & \valgood{+10} & 0.9228 & 0.9358 & \valgood{+0.0130} \\
\addlinespace[2pt]
Leonardo da Vinci, \textit{Madonna Litta (Madonna and the Child)}\newline(Unlearning, top-1) & 9 & 20 & \valgood{$-$11} & 0.9267 & 0.5173 & \valgood{$-$0.4094} \\
\bottomrule
\end{tabularx}

\vspace{0.3em}
\label{tab:val_sd_clip_def}
\setlength{\tabcolsep}{4pt}
\begin{tabularx}{\columnwidth}{@{}>{\raggedright\arraybackslash}X c c c c c@{}}
\toprule
\multirow{2}{*}{\textbf{Target artwork / Model}} & \multicolumn{5}{c}{\textbf{Defender loss rate (\%) at margin $\delta$}} \\
\cmidrule(l){2-6}
 & 0.00 & 0.05 & 0.10 & 0.15 & 0.20 \\
\midrule
\multicolumn{6}{@{}c}{Raphael, \textit{The Fall on the Road to Calvary} (bottom-1)} \\ \hline
\quad Base & 31.6 & 20.0 & 12.6 & 7.4 & 3.2 \\
\quad Fine-tuned & 18.9 & 10.5 & 4.2 & 1.1 & 0.0 \\
\quad Change & \valgood{$-$12.6} & \valgood{$-$9.5} & \valgood{$-$8.4} & \valgood{$-$6.3} & \valgood{$-$3.2} \\
\midrule
\multicolumn{6}{@{}c}{Leonardo da Vinci, \textit{Madonna Litta (Madonna and the Child)} (top-1)} \\ \hline
\quad Base & 23.2 & 17.9 & 9.5 & 7.4 & 2.1 \\
\quad Unlearned & 81.1 & 54.7 & 27.4 & 11.6 & 6.3 \\
\quad Change & \valgood{+57.9} & \valgood{+36.8} & \valgood{+17.9} & \valgood{+4.2} & \valgood{+4.2} \\
\bottomrule
\end{tabularx}
\end{table}

\begin{table}[t]
\centering
\scriptsize
\setlength{\tabcolsep}{2.5pt}
\renewcommand{\arraystretch}{1.0}
\caption{\textbf{Controlled validation of ArtArena under SD v1.5 for Aesthetics.}
Top: Changes in imitation rank and fitness following strengthening of the bottom-ranked artwork and weakening of the top-ranked artwork from the Influence Ledger.
Bottom: Defender Loss Rate (DLR) before and after each intervention, along with the corresponding changes across comparison margins $\delta$. Scores are LPIPS distances, so lower values are closer.}
\label{tab:val_sd_lpips_imit}
\begin{tabularx}{\columnwidth}{@{}>{\raggedright\arraybackslash}X c c c c c c@{}}
\toprule
\multirow{2}{*}{\textbf{Target artwork}} & \multicolumn{3}{c}{\textbf{Imitation rank}} & \multicolumn{3}{c}{\textbf{Fitness score}} \\
\cmidrule(lr){2-4}\cmidrule(l){5-7}
 & Base & New & Shift & Base & New & Change \\
\midrule
Andy Warhol, \textit{Rorschach}\newline(Fine-tuning, bottom-1) & 17 & 1 & \valgood{+16} & 0.5939 & 0.5177 & \valgood{$-$0.0762} \\
\addlinespace[2pt]
Jackson Pollock, \textit{Circumcision January}\newline(Unlearning, top-1) & 1 & 9 & \valgood{$-$8} & 0.5419 & 0.6155 & \valgood{+0.0735} \\
\bottomrule
\end{tabularx}

\vspace{0.3em}
\label{tab:val_sd_lpips_def}
\setlength{\tabcolsep}{4pt}
\begin{tabularx}{\columnwidth}{@{}>{\raggedright\arraybackslash}X c c c c c@{}}
\toprule
\multirow{2}{*}{\textbf{Target artwork / Model}} & \multicolumn{5}{c}{\textbf{Defender loss rate (\%) at margin $\delta$}} \\
\cmidrule(l){2-6}
 & 0.00 & 0.05 & 0.10 & 0.15 & 0.20 \\
\midrule
\multicolumn{6}{@{}c}{Andy Warhol, \textit{Rorschach} (bottom-1)} \\ \hline
\quad Base & 89.5 & 63.2 & 27.4 & 9.5 & 1.1 \\
\quad Fine-tuned & 15.8 & 3.2 & 0.0 & 0.0 & 0.0 \\
\quad Change & \valgood{$-$73.7} & \valgood{$-$60.0} & \valgood{$-$27.4} & \valgood{$-$9.5} & \valgood{$-$1.1} \\
\midrule
\multicolumn{6}{@{}c}{Jackson Pollock, \textit{Circumcision January} (top-1)} \\ \hline
\quad Base & 14.7 & 1.1 & 1.1 & 0.0 & 0.0 \\
\quad Unlearned & 20.0 & 2.1 & 0.0 & 0.0 & 0.0 \\
\quad Change & \valgood{+5.3} & \valgood{+1.1} & \valbad{$-$1.1} & 0.0 & 0.0 \\
\bottomrule
\end{tabularx}
\end{table}

\begin{table}[t]
\centering
\scriptsize
\setlength{\tabcolsep}{2.5pt}
\renewcommand{\arraystretch}{1.0}
\caption{\textbf{Controlled validation of ArtArena under SD v1.5 for Fidelity.}
Top: Changes in imitation rank and fitness following strengthening of the bottom-ranked artwork and weakening of the top-ranked artwork from the Influence Ledger.
Bottom: Defender Loss Rate (DLR) before and after each intervention, along with the corresponding changes across comparison margins $\delta$.}
\label{tab:val_sd_csd_imit}
\begin{tabularx}{\columnwidth}{@{}>{\raggedright\arraybackslash}X c c c c c c@{}}
\toprule
\multirow{2}{*}{\textbf{Target artwork}} & \multicolumn{3}{c}{\textbf{Imitation rank}} & \multicolumn{3}{c}{\textbf{Fitness score}} \\
\cmidrule(lr){2-4}\cmidrule(l){5-7}
 & Base & New & Shift & Base & New & Change \\
\midrule
Vincent van Gogh, \textit{Flowering Garden}\newline(Fine-tuning, bottom-1) & 20 & 9 & \valgood{+11} & 0.8297 & 0.8874 & \valgood{+0.0577} \\
\addlinespace[2pt]
Vincent van Gogh, \textit{Olive Grove}\newline(Unlearning, top-1) & 1 & 20 & \valgood{$-$19} & 0.8978 & 0.6010 & \valgood{$-$0.2969} \\
\bottomrule
\end{tabularx}

\vspace{0.3em}
\label{tab:val_sd_csd_def}
\setlength{\tabcolsep}{4pt}
\begin{tabularx}{\columnwidth}{@{}>{\raggedright\arraybackslash}X c c c c c@{}}
\toprule
\multirow{2}{*}{\textbf{Target artwork / Model}} & \multicolumn{5}{c}{\textbf{Defender loss rate (\%) at margin $\delta$}} \\
\cmidrule(l){2-6}
 & 0.00 & 0.05 & 0.10 & 0.15 & 0.20 \\
\midrule
\multicolumn{6}{@{}c}{Vincent van Gogh, \textit{Flowering Garden} (bottom-1)} \\ \hline
\quad Base & 60.0 & 53.7 & 36.8 & 16.8 & 5.3 \\
\quad Fine-tuned & 0.0 & 0.0 & 0.0 & 0.0 & 0.0 \\
\quad Change & \valgood{$-$60.0} & \valgood{$-$53.7} & \valgood{$-$36.8} & \valgood{$-$16.8} & \valgood{$-$5.3} \\
\midrule
\multicolumn{6}{@{}c}{Vincent van Gogh, \textit{Olive Grove} (top-1)} \\ \hline
\quad Base & 16.8 & 1.1 & 0.0 & 0.0 & 0.0 \\
\quad Unlearned & 1.1 & 0.0 & 0.0 & 0.0 & 0.0 \\
\quad Change & \valbad{$-$15.8} & \valbad{$-$1.1} & 0.0 & 0.0 & 0.0 \\
\bottomrule
\end{tabularx}
\end{table}

\begin{table}[t]
\centering
\scriptsize
\setlength{\tabcolsep}{2.5pt}
\renewcommand{\arraystretch}{1.0}
\caption{\textbf{Controlled validation of ArtArena under SDXL for Semantics.}
Top: Changes in imitation rank and fitness following strengthening of the bottom-ranked artwork and weakening of the top-ranked artwork from the Influence Ledger.
Bottom: Defender Loss Rate (DLR) before and after each intervention, along with the corresponding changes across comparison margins $\delta$.}
\label{tab:val_sdxl_clip_imit}
\begin{tabularx}{\columnwidth}{@{}>{\raggedright\arraybackslash}X c c c c c c@{}}
\toprule
\multirow{2}{*}{\textbf{Target artwork}} & \multicolumn{3}{c}{\textbf{Imitation rank}} & \multicolumn{3}{c}{\textbf{Fitness score}} \\
\cmidrule(lr){2-4}\cmidrule(l){5-7}
 & Base & New & Shift & Base & New & Change \\
\midrule
Caravaggio, \textit{Incredulity of Saint Thomas}\newline(Fine-tuning, bottom-1) & 17 & 1 & \valgood{+16} & 0.9170 & 0.9546 & \valgood{+0.0376} \\
\addlinespace[2pt]
Vincent van Gogh, \textit{Vincent's Bedroom in Arles}\newline(Unlearning, top-1) & 16 & 16 & 0 & 0.9175 & 0.7397 & \valgood{$-$0.1777} \\
\bottomrule
\end{tabularx}

\vspace{0.3em}
\label{tab:val_sdxl_clip_def}
\setlength{\tabcolsep}{4pt}
\begin{tabularx}{\columnwidth}{@{}>{\raggedright\arraybackslash}X c c c c c@{}}
\toprule
\multirow{2}{*}{\textbf{Target artwork / Model}} & \multicolumn{5}{c}{\textbf{Defender loss rate (\%) at margin $\delta$}} \\
\cmidrule(l){2-6}
 & 0.00 & 0.05 & 0.10 & 0.15 & 0.20 \\
\midrule
\multicolumn{6}{@{}c}{Caravaggio, \textit{Incredulity of Saint Thomas} (bottom-1)} \\ \hline
\quad Base & 52.6 & 43.2 & 36.8 & 28.4 & 20.0 \\
\quad Fine-tuned & 11.6 & 10.5 & 6.3 & 3.2 & 3.2 \\
\quad Change & \valgood{$-$41.1} & \valgood{$-$32.6} & \valgood{$-$30.5} & \valgood{$-$25.3} & \valgood{$-$16.8} \\
\midrule
\multicolumn{6}{@{}c}{Vincent van Gogh, \textit{Vincent's Bedroom in Arles} (top-1)} \\ \hline
\quad Base & 33.7 & 30.5 & 24.2 & 12.6 & 0.0 \\
\quad Unlearned & 37.9 & 33.7 & 23.2 & 6.3 & 0.0 \\
\quad Change & \valgood{+4.2} & \valgood{+3.2} & \valbad{$-$1.1} & \valbad{$-$6.3} & 0.0 \\
\bottomrule
\end{tabularx}
\end{table}

\begin{table}[t]
\centering
\scriptsize
\setlength{\tabcolsep}{2.5pt}
\renewcommand{\arraystretch}{1.0}
\caption{\textbf{Controlled validation of ArtArena under SDXL for Aesthetics.}
Top: Changes in imitation rank and fitness following strengthening of the bottom-ranked artwork and weakening of the top-ranked artwork from the Influence Ledger.
Bottom: Defender Loss Rate (DLR) before and after each intervention, along with the corresponding changes across comparison margins $\delta$. Scores are LPIPS distances, so lower values are closer.}
\label{tab:val_sdxl_lpips_imit}
\begin{tabularx}{\columnwidth}{@{}>{\raggedright\arraybackslash}X c c c c c c@{}}
\toprule
\multirow{2}{*}{\textbf{Target artwork}} & \multicolumn{3}{c}{\textbf{Imitation rank}} & \multicolumn{3}{c}{\textbf{Fitness score}} \\
\cmidrule(lr){2-4}\cmidrule(l){5-7}
 & Base & New & Shift & Base & New & Change \\
\midrule
Michelangelo, \textit{Design for Julius II tomb (first version)}\newline(Fine-tuning, bottom-1) & 19 & 2 & \valgood{+17} & 0.6597 & 0.4770 & \valgood{$-$0.1827} \\
\addlinespace[2pt]
Georgia O'Keeffe, \textit{Sky Above Clouds IV}\newline(Unlearning, top-1) & 3 & 6 & \valgood{$-$3} & 0.6200 & 0.6891 & \valgood{+0.0690} \\
\bottomrule
\end{tabularx}

\vspace{0.3em}
\label{tab:val_sdxl_lpips_def}
\setlength{\tabcolsep}{4pt}
\begin{tabularx}{\columnwidth}{@{}>{\raggedright\arraybackslash}X c c c c c@{}}
\toprule
\multirow{2}{*}{\textbf{Target artwork / Model}} & \multicolumn{5}{c}{\textbf{Defender loss rate (\%) at margin $\delta$}} \\
\cmidrule(l){2-6}
 & 0.00 & 0.05 & 0.10 & 0.15 & 0.20 \\
\midrule
\multicolumn{6}{@{}c}{Michelangelo, \textit{Design for Julius II tomb (first version)} (bottom-1)} \\ \hline
\quad Base & 85.3 & 56.8 & 28.4 & 12.6 & 6.3 \\
\quad Fine-tuned & 0.0 & 0.0 & 0.0 & 0.0 & 0.0 \\
\quad Change & \valgood{$-$85.3} & \valgood{$-$56.8} & \valgood{$-$28.4} & \valgood{$-$12.6} & \valgood{$-$6.3} \\
\midrule
\multicolumn{6}{@{}c}{Georgia O'Keeffe, \textit{Sky Above Clouds IV} (top-1)} \\ \hline
\quad Base & 5.3 & 2.1 & 0.0 & 0.0 & 0.0 \\
\quad Unlearned & 10.5 & 3.2 & 0.0 & 0.0 & 0.0 \\
\quad Change & \valgood{+5.3} & \valgood{+1.1} & 0.0 & 0.0 & 0.0 \\
\bottomrule
\end{tabularx}
\end{table}

\begin{table}[t]
\centering
\scriptsize
\setlength{\tabcolsep}{2.5pt}
\renewcommand{\arraystretch}{1.0}
\caption{\textbf{Controlled validation of ArtArena under SANA-1.5 for Semantics.}
Top: Changes in imitation rank and fitness following strengthening of the bottom-ranked artwork and weakening of the top-ranked artwork from the Influence Ledger.
Bottom: Defender Loss Rate (DLR) before and after each intervention, along with the corresponding changes across comparison margins $\delta$.}
\label{tab:val_sana_clip_imit}
\begin{tabularx}{\columnwidth}{@{}>{\raggedright\arraybackslash}X c c c c c c@{}}
\toprule
\multirow{2}{*}{\textbf{Target artwork}} & \multicolumn{3}{c}{\textbf{Imitation rank}} & \multicolumn{3}{c}{\textbf{Fitness score}} \\
\cmidrule(lr){2-4}\cmidrule(l){5-7}
 & Base & New & Shift & Base & New & Change \\
\midrule
Andy Warhol, \textit{Mickey}\newline(Fine-tuning, bottom-1) & 13 & 6 & \valgood{+7} & 0.8964 & 0.9065 & \valgood{+0.0101} \\
\addlinespace[2pt]
Andy Warhol, \textit{Green Coca-Cola Bottles}\newline(Unlearning, top-1) & 8 & 19 & \valgood{$-$11} & 0.9092 & 0.7605 & \valgood{$-$0.1487} \\
\bottomrule
\end{tabularx}

\vspace{0.3em}
\label{tab:val_sana_clip_def}
\setlength{\tabcolsep}{4pt}
\begin{tabularx}{\columnwidth}{@{}>{\raggedright\arraybackslash}X c c c c c@{}}
\toprule
\multirow{2}{*}{\textbf{Target artwork / Model}} & \multicolumn{5}{c}{\textbf{Defender loss rate (\%) at margin $\delta$}} \\
\cmidrule(l){2-6}
 & 0.00 & 0.05 & 0.10 & 0.15 & 0.20 \\
\midrule
\multicolumn{6}{@{}c}{Andy Warhol, \textit{Mickey} (bottom-1)} \\ \hline
\quad Base & 90.5 & 76.8 & 52.6 & 31.6 & 19.0 \\
\quad Fine-tuned & 51.6 & 36.8 & 21.1 & 13.7 & 12.6 \\
\quad Change & \valgood{$-$39.0} & \valgood{$-$40.0} & \valgood{$-$31.6} & \valgood{$-$17.9} & \valgood{$-$6.3} \\
\midrule
\multicolumn{6}{@{}c}{Andy Warhol, \textit{Green Coca-Cola Bottles} (top-1)} \\ \hline
\quad Base & 30.5 & 19.0 & 6.3 & 4.2 & 2.1 \\
\quad Unlearned & 33.7 & 21.2 & 8.5 & 5.2 & 3.2 \\
\quad Change & \valgood{+3.2} & \valgood{+2.2} & \valgood{+2.1} & \valgood{+1.0} & \valgood{+1.1} \\
\bottomrule
\end{tabularx}
\end{table}

\begin{table}[t]
\centering
\scriptsize
\setlength{\tabcolsep}{2.5pt}
\renewcommand{\arraystretch}{1.0}
\caption{\textbf{Controlled validation of ArtArena under SANA-1.5 for Aesthetics.}
Top: Changes in imitation rank and fitness following strengthening of the bottom-ranked artwork and weakening of the top-ranked artwork from the Influence Ledger.
Bottom: Defender Loss Rate (DLR) before and after each intervention, along with the corresponding changes across comparison margins $\delta$. Scores are LPIPS distances, so lower values are closer.}
\label{tab:val_sana_lpips_imit}
\begin{tabularx}{\columnwidth}{@{}>{\raggedright\arraybackslash}X c c c c c c@{}}
\toprule
\multirow{2}{*}{\textbf{Target artwork}} & \multicolumn{3}{c}{\textbf{Imitation rank}} & \multicolumn{3}{c}{\textbf{Fitness score}} \\
\cmidrule(lr){2-4}\cmidrule(l){5-7}
 & Base & New & Shift & Base & New & Change \\
\midrule
Rembrandt, \textit{Portrait of a Seated Man Rising from His Chair}\newline(Fine-tuning, bottom-1) & 10 & 1 & \valgood{+9} & 0.6178 & 0.4719 & \valgood{$-$0.1459} \\
\addlinespace[2pt]
Jackson Pollock, \textit{Echo (Number 25)}\newline(Unlearning, top-1) & 7 & 19 & \valgood{$-$12} & 0.6096 & 0.9076 & \valgood{+0.2980} \\
\bottomrule
\end{tabularx}

\vspace{0.3em}
\label{tab:val_sana_lpips_def}
\setlength{\tabcolsep}{4pt}
\begin{tabularx}{\columnwidth}{@{}>{\raggedright\arraybackslash}X c c c c c@{}}
\toprule
\multirow{2}{*}{\textbf{Target artwork / Model}} & \multicolumn{5}{c}{\textbf{Defender loss rate (\%) at margin $\delta$}} \\
\cmidrule(l){2-6}
 & 0.00 & 0.05 & 0.10 & 0.15 & 0.20 \\
\midrule
\multicolumn{6}{@{}c}{Rembrandt, \textit{Portrait of a Seated Man Rising from His Chair} (bottom-1)} \\ \hline
\quad Base & 91.6 & 81.1 & 52.6 & 27.4 & 12.6 \\
\quad Fine-tuned & 89.5 & 77.2 & 49.7 & 32.6 & 8.6 \\
\quad Change & \valgood{$-$2.1} & \valgood{$-$3.9} & \valgood{$-$2.9} & \valbad{+5.3} & \valgood{$-$4.1} \\
\midrule
\multicolumn{6}{@{}c}{Jackson Pollock, \textit{Echo (Number 25)} (top-1)} \\ \hline
\quad Base & 14.7 & 9.5 & 6.3 & 5.3 & 5.3 \\
\quad Unlearned & 22.1 & 12.5 & 7.3 & 5.3 & 3.1 \\
\quad Change & \valgood{+7.4} & \valgood{+3.0} & \valgood{+1.0} & 0.0 & \valbad{$-$2.2} \\
\bottomrule
\end{tabularx}
\end{table}

\begin{table}[t]
\centering
\scriptsize
\setlength{\tabcolsep}{2.5pt}
\renewcommand{\arraystretch}{1.0}
\caption{\textbf{Controlled validation of ArtArena under SANA-1.5 for Fidelity.}
Top: Changes in imitation rank and fitness following strengthening of the bottom-ranked artwork and weakening of the top-ranked artwork from the Influence Ledger.
Bottom: Defender Loss Rate (DLR) before and after each intervention, along with the corresponding changes across comparison margins $\delta$.}
\label{tab:val_sana_csd_imit}
\begin{tabularx}{\columnwidth}{@{}>{\raggedright\arraybackslash}X c c c c c c@{}}
\toprule
\multirow{2}{*}{\textbf{Target artwork}} & \multicolumn{3}{c}{\textbf{Imitation rank}} & \multicolumn{3}{c}{\textbf{Fitness score}} \\
\cmidrule(lr){2-4}\cmidrule(l){5-7}
 & Base & New & Shift & Base & New & Change \\
\midrule
Vincent van Gogh, \textit{Mademoiselle Gachet in her Garden at Auvers-sur-Oise}\newline(Fine-tuning, bottom-1) & 5 & 1 & \valgood{+4} & 0.7941 & 0.9541 & \valgood{+0.1601} \\
\addlinespace[2pt]
Jackson Pollock, \textit{Convergence (Number 10)}\newline(Unlearning, top-1) & 17 & 20 & \valgood{$-$3} & 0.7439 & 0.4433 & \valgood{$-$0.3006} \\
\bottomrule
\end{tabularx}

\vspace{0.3em}
\label{tab:val_sana_csd_def}
\setlength{\tabcolsep}{4pt}
\begin{tabularx}{\columnwidth}{@{}>{\raggedright\arraybackslash}X c c c c c@{}}
\toprule
\multirow{2}{*}{\textbf{Target artwork / Model}} & \multicolumn{5}{c}{\textbf{Defender loss rate (\%) at margin $\delta$}} \\
\cmidrule(l){2-6}
 & 0.00 & 0.05 & 0.10 & 0.15 & 0.20 \\
\midrule
\multicolumn{6}{@{}c}{\begin{tabular}{@{}c@{}}Vincent van Gogh, \textit{Mademoiselle Gachet in her Garden}\\\textit{at Auvers-sur-Oise} (bottom-1)\end{tabular}} \\ \hline
\quad Base & 55.8 & 33.7 & 25.3 & 16.8 & 14.7 \\
\quad Fine-tuned & 53.7 & 24.2 & 17.9 & 13.7 & 5.3 \\
\quad Change & \valgood{$-$2.1} & \valgood{$-$9.5} & \valgood{$-$7.4} & \valgood{$-$3.2} & \valgood{$-$9.5} \\
\midrule
\multicolumn{6}{@{}c}{Jackson Pollock, \textit{Convergence (Number 10)} (top-1)} \\ \hline
\quad Base & 28.4 & 17.9 & 12.6 & 5.3 & 1.1 \\
\quad Unlearned & 39.0 & 23.6 & 6.3 & 0.0 & 0.0 \\
\quad Change & \valgood{+10.5} & \valgood{+5.7} & \valbad{$-$6.3} & \valbad{$-$5.3} & \valbad{$-$1.1} \\
\bottomrule
\end{tabularx}
\end{table}

\end{document}